%% file: main.tex
\documentclass{article}
\usepackage{microtype}
\usepackage{graphicx}
\usepackage{booktabs} 
\usepackage{hyperref}
\usepackage{tabularx}

\usepackage{caption}
\usepackage{subcaption}
\usepackage[accepted]{icml2024}
\usepackage{multirow}
\usepackage{makecell}

\usepackage{amsmath}

\usepackage{amssymb}
\usepackage{mathtools}
\usepackage{amsfonts}
\usepackage{amsthm}
\usepackage[capitalize,noabbrev]{cleveref}
\theoremstyle{plain}

\theoremstyle{definition}

\theoremstyle{remark}

\usepackage[textsize=tiny]{todonotes}

\input{math_commands.tex}

\newcommand{\dataset}{\mathcal{D}}

\newcommand{\tablestyle}[2]{\setlength{\tabcolsep}{#1}\renewcommand{\arraystretch}{#2}\centering\small}

\icmltitlerunning{Position: Reinforcement Learning in Dynamic Treatment Regimes Needs Critical Reexamination}

\begin{document}

\twocolumn[
\icmltitle{Position: Reinforcement Learning in Dynamic Treatment Regimes Needs Critical Reassessment}


\icmlsetsymbol{equal}{*}

\begin{icmlauthorlist}
\icmlauthor{Zhiyao Luo}{engsci}
\icmlauthor{Yangchen Pan}{engsci}
\icmlauthor{Peter Watkinson}{ccg}
\icmlauthor{Tingting Zhu}{engsci}
\end{icmlauthorlist}

\icmlaffiliation{engsci}{Department of Engineering Science, University of Oxford, Parks Road, Oxford OX1 3PJ, United Kingdom}
\icmlaffiliation{ccg}{Nuffield Department of Population Health (NDPH), University of Oxford, Richard Doll Building, Old Road Campus, Headington, Oxford OX3 7LF, United Kingdom }
\icmlcorrespondingauthor{Zhiyao Luo}{zhiyao.luo@eng.ox.ac.uk}
\icmlkeywords{Offline Reinforcement Learning, AI for Healthcare, Dynamic Treatment Regime}

\vskip 0.3in
]



\printAffiliationsAndNotice{}  

\begin{abstract}

In the rapidly changing healthcare landscape, the implementation of offline reinforcement learning (RL) in dynamic treatment regimes (DTRs) presents a mix of unprecedented opportunities and challenges. This position paper offers a critical examination of the current status of offline RL in the context of DTRs. We argue for a reassessment of applying RL in DTRs, citing concerns such as inconsistent and potentially inconclusive evaluation metrics, the absence of naive and supervised learning baselines, and the diverse choice of RL formulation in existing research. Through a case study with more than 17,000 evaluation experiments using a publicly available Sepsis dataset, we demonstrate that the performance of RL algorithms can significantly vary with changes in evaluation metrics and Markov Decision Process (MDP) formulations. Surprisingly, it is observed that in some instances, RL algorithms can be surpassed by random baselines subjected to policy evaluation methods and reward design. This calls for more careful policy evaluation and algorithm development in future DTR works. Additionally, we discussed potential enhancements toward more reliable development of RL-based dynamic treatment regimes and invited further discussion within the community. Code is available at \href{https://github.com/GilesLuo/ReassessDTR}{https://github.com/GilesLuo/ReassessDTR}.
\end{abstract}

\input{main_sections/1introduction}

\input{main_sections/2background}

\input{main_sections/3formulation}

\input{main_sections/4ope}

\input{main_sections/5reward_designs}

\input{main_sections/6baselines}

\input{main_sections/7result}

\input{main_sections/9suggestions}

\bibliography{main_bib}
\bibliographystyle{icml2024}

\newpage
\appendix
\onecolumn
\input{appendix_sections/1data_preprocessing}
\input{appendix_sections/2OPE_details}
\input{appendix_sections/3baseline_details}
\input{appendix_sections/4training_details}
\input{appendix_sections/5discussion}
\input{appendix_sections/model_calibration}
\input{appendix_sections/6full_result}
\input{appendix_sections/ratio_plot}

\end{document}

%% file: math_commands.tex
\usepackage{amsmath,amsfonts,bm}

\def\eqref#1{equation~\ref{#1}}

\def\1{\bm{1}}

\DeclareMathAlphabet{\mathsfit}{\encodingdefault}{\sfdefault}{m}{sl}
\SetMathAlphabet{\mathsfit}{bold}{\encodingdefault}{\sfdefault}{bx}{n}

\def\sD{{\mathbb{D}}}

\def\sR{{\mathbb{R}}}

\def\cA{{\mathcal{A}}}

\def\cD{{\mathcal{D}}}

\def\cS{{\mathcal{S}}}

\newcommand{\KL}{D_{\mathrm{KL}}}

\newcommand{\EE}{\mathbb{E}}

%% file: main_sections/1introduction.tex
\section{Introduction}

The advent of machine learning in the medical field has opened new avenues for treatment optimization \cite{chernozhukov2018double,myszczynska2020applications}. Among various machine learning techniques, reinforcement learning (RL) has steadily gained recognition as a transformative tool in healthcare \cite{coronato2020reinforcement, yu2021reinforcement}, particularly within dynamic treatment regimes (DTRs)\cite{chakraborty2014dynamic}. The core strength of RL-DTRs is their ability to learn from patient responses to treatments and adapt treatment plans accordingly, leading to better patient outcomes. This adaptability is especially valuable in healthcare, where patient conditions and treatment responses are often diverse and can change over time. 

Recent advancements in RL applied to DTRs, especially in offline RL\cite{agarwal2020optimistic}, have demonstrated promising potential to guide future treatment decisions without direct environment interaction. For instance, \citeauthor{wu2023value} introduced a weighted dueling double deep Q-network with embedded human expertise (WD3QNE), addressing domain knowledge embedding. \citeauthor{luckett2019estimating} explored V-learning to estimate optimal DTRs in mobile health, addressing the challenges of indefinite time horizons and high-frequency decision-making. These studies highlight the diverse applications of offline RL in healthcare, from improving personalized treatment in chronic diseases to optimizing real-time interventions in mobile platforms. 

The application of RL in DTRs is not without its criticisms \cite{jeter2019does} \cite{gottesman2018evaluating}, prompting a thorough reassessment of its utility and necessity in this field. One major issue is the lack of standardized evaluation metrics.  For example, \citeauthor{komorowski2018artificial} use weighted importance sampling; \citeauthor{raghu2017deep, raghu2018model, peng2018improving} use different variants of doubly robust off-policy evaluation methods \cite{jiang2016doubly}. Another approach involves using direct methods \cite{mannor2007bias}, yet this, too, lacks uniformity across studies. The diversity of evaluation techniques underscores the need for a consensus on benchmark metrics to facilitate meaningful comparisons in this field. Furthermore, the diverse formulations of Markov decision processes (MDPs) and the lack of established baselines make it challenging to gauge the treatment improvement offered by RL approaches over traditional methods. Regarding the state space, \citeauthor{komorowski2018artificial} converted continuous variables into discrete clusters, in contrast to \citeauthor{liu2017deep, wang2018supervised}, who used continuous variables directly. In action space design, although discrete actions are commonly employed, some studies, such as \cite{wang2022learning, huang2022reinforcement} have explored continuous action spaces. For reward, \citeauthor{komorowski2018artificial} chose simplistic 90-day mortality for patient outcomes, whereas \citeauthor{raghu2017deep,raghu2018model, peng2018improving} integrated clinically validated risk scores into the reward function, but with varying weights on reward components. This diversity highlights systematic reevaluation to enhance RL's reliability and effectiveness in developing treatment plans.

The position of this paper is that \textbf{while RL holds significant promise for optimizing DTRs in healthcare, there is a critical need for a comprehensive reassessment of its application.} Previous studies \cite{gottesman2018evaluating,tang2021model} focused mainly on practical healthcare concerns with respect to policy evaluation. However, we call for a more thorough understanding of methodological inconsistencies, including variations in policy evaluation metrics, the absence of standardized baselines for comparison, and the diverse formulations of MDPs. The paper advocates for the establishment of uniform standards and methodologies to ensure that RL's application is both scientifically rigorous and practically applicable to healthcare.

Our paper presents a critical perspective on the effectiveness and necessity of using RL in DTRs. Initially, we review fundamental concepts in offline RL and its application in DTRs. We then review the literature in this domain, highlighting the various policy evaluation methods used in different studies and pointing out that the diversity often leads to significant variability in performance among algorithms. Furthermore, we compare basic baselines (e.g., random policies and supervised learning) that a notable number of existing studies omitted and found that RL algorithms can underperform compared to these simple baselines, which raises a cause for concern. Moving forward, we standardize our evaluation method to focus on the impact of different reward designs, demonstrating that varying rewards can lead to relatively disparate performance. Through our critical analysis, we offer several considerations to improve the reliability of model development and evaluations in this field.

%% file: main_sections/2background.tex
\section{Background}

This section provides the basic notation and conceptual background necessary for applying offline RL to DTRs, along with a common offline RL formulation for addressing problems in DTR.

\subsection{RL and Offline RL}
\label{sec:offline-rl}

RL is building upon Markov decision process (MDP) denoted as $M=\{\cS, \cA, P, r, \gamma\}$  \citep{puterman2014markov}, 
where $\cS$ is the state space, $\cA$ is the action space, 
 $\gamma\in[0, 1)$ is the discount factor, 
$r:\cS\times\cA\rightarrow \sR$ 
and 
$P:\cS\times\cA\rightarrow \cS)$ 
are the reward and transition functions, respectively. 
The \emph{value function} is the expectation of future discounted total reward obtained by following a policy $\pi:\cS\rightarrow \Delta(\cA)$, 
$v^\pi(s) = \EE^\pi[ \sum_{t=0}^\infty \gamma^t r(s_t, a_t) | s_0 = s]$ 
where $\EE^\pi$ means the expectation under the policy $\pi$ and the transition probability. The corresponding \emph{action-value} function is $q^\pi(s,a) = r(s,a) + \gamma \EE_{s'\sim P(\cdot|s,a)} [v^\pi(s')]$. 
The goal is to find an \emph{optimal policy} $\pi^*$ that maximizes the values $\forall s\in\cS$. 

In an \emph{offline RL} setting, our focus is on learning an optimal policy for decision-making based on a pre-gathered dataset, denoted as $\dataset = \{s_i, a_i, r_i, s_i'\}^{n-1}_{i=0}$. This dataset is assumed to be the result of actions taken according to a specific behavior policy $\pi_\dataset$. One primary challenge in offline RL is that $\pi_\dataset$ may not thoroughly explore all possible actions, leading to potential overestimation of those out-of-distribution actions. Acting greedily with respect to such actions could be problematic~\cite {fujimoto2019offpolicy}. 

To address the challenge, a widely adopted method involves restricting the learned policy $\pi$ to be close to a baseline policy $\pi_\dataset$, by incorporating a KL-divergence term into the optimization objective: $\max_{\pi} \EE_{s\sim \rho}[ \sum_{a} \pi(a | s) q(s,a) - \tau \KL(\pi(\cdot | s) || \pi_\dataset(\cdot | s))]$ with $\tau > 0$. This formulation ensures that the optimized policy $\pi$ is only supported where $\pi_\dataset$ is non-zero, effectively setting $\pi(a|s) = 0$ wherever $\pi_\dataset(a|s)=0$. This principle underlies numerous offline RL strategies~\citep{wu2019offlinerl,peng2020advantage,nair2021awac,brandfonbrener2021offline,fujimoto2021minimalist}. \citeauthor{fujimoto2021minimalist} introduce a behaviour cloning regularization term, $(\pi(s) -a)^2$. An alternative approach is in-sample policy optimization, which aims to prevent the selection of actions outside $\pi_\dataset$'s distribution. A more recent work~\citep{xiao2023inac} proposes an in-sample softmax to directly sample from support. We will later provide more details about the algorithms used in our experiments. 

\subsection{Dynamic Treatment Regime}\label{sec:dtr}

A Dynamic Treatment Regime (DTR) represents a sequence of decision rules that guide treatment adaptations over time based on an individual patient's evolving conditions and responses. An action (i.e., treatment) should take into account the patient's current state, including medical history and previous treatment responses, to recommend the next best action. Offline RL is particularly suitable for DTR because it allows the use of extensive historical healthcare data to learn optimal treatment strategies without the need for real-time experimentation in patients. 

In this work, we use the intravenous vasopressor fluid (IV) and vasopressor dosage regime task \cite{komorowski2018artificial} for sepsis treatment in the intensive care unit (ICU) as a proof-of-position. The dataset is derived from the Medical Information Mart for Intensive Care III (MIMIC-III) database \cite{johnson2016mimic}.  The selection of patients follows the sepsis-3 criteria \cite{singer2016third}, focusing on the early stage of sepsis management up to 24 hours prior to and 48 hours after the estimated onset of sepsis.

%% file: main_sections/3formulation.tex
\section{Problem Formulation in Offline RL for DTR}
\label{sec:3problem formulation}

A DTR problem is typically modeled as a finite-time MDP as introduced in Section \ref{sec:offline-rl}:

\textbf{States }($s \in \cS$): time-varying clinical variables that represent the medical condition of a patient. Particularly, this paper considers \textit{continuous states} to avoid introducing variance from clustering-based state discretization algorithms.

\textbf{Actions} ($a \in \cA$): actionable treatment decisions. This work primarily focuses on \textit{discrete actions}, as they are not only more extensively studied in literature but also have a more established theoretical foundation of policy evaluation.

\textbf{ Rewards }($r$): the outcome of an action taken in a particular state, reflecting improvements or deterioration in the patient's condition. Various factors can be taken into account in the reward design, including the outcome of treatment, the clinical risk score, and the abnormality of vital signs. 

In our sepsis cohort, the curation and preprocessing of raw data are mostly reproduced from \citeauthor{komorowski2018artificial}'s work. Data are binned into 4-hour intervals, with 46 observational variables at each step. The dataset categorizes IV fluid and vasopressor dosages into five classes (See Appendix \ref{sec:app-data description}), resulting in a $5 \times5$ discrete action space.  Additionally, we employed a series of important modifications. First, we change the treatment outcome from 90-day to in-hospital mortality under clinical guidance to strengthen the correlation between actions and outcomes. Secondly, we exclude patients who have inconsistent time-series data\footnote{The original patient cohort included patients who lost track of information in the middle of treatment trajectories.  This means that they omit the irregular sampling frequency and consider them as consecutive time steps, which is not a common practice in RL. Therefore, we remove patients who have missing information within 4 hours during admissions.}. Furthermore, we removed the 'input 4-hourly' feature from the observation space, as the feature is part of the action. 

%% file: main_sections/4ope.tex
\section{Diversity and Inconsistency of Policy Evaluation Methods in RL-DTR}


\subsection{Challenges of Policy Evaluation in RL}
Evaluating offline RL algorithms for DTRs is challenging for several reasons: \textbf{a) }The dataset is fixed and observational, meaning RL cannot be evaluated by interacting with the environment. \textbf{ b)} Medical decision-making is complex, as the effects of treatments may not be immediately apparent and can be influenced by many confounding factors. \textbf{c) }Patients' responses to treatments are inherently uncertain and variable, making it difficult to assess the true effectiveness of proposed treatment policies. These challenges make it harder to evaluate offline RL algorithms in the context of DTRs, compared to traditional RL settings.

Recent literature addresses the evaluation challenges in DTR by a range of policy evaluation techniques. Notable among these are Inverse Probability Weighting (IPW) \cite{liu2017deep}, Weighted Importance Sampling (WIS) \cite{kidambi2020mbofflinerl, nambiar2023deep}, the Direct Method (DM)\cite{huang2022reinforcement, kondrup2023towards}, and Doubly Robust (DR)\cite{raghu2017deep, wu2023value, wang2018supervised} estimators. These methods tried to tackle the confounding variables and create a counterfactual estimation based on historical data. 

\subsection{Existing Evaluation Methods in RL-DTR}
An offline policy evaluation (OPE) aims to estimate a policy’s value \( \hat{V}^{\pi} \) using a behavior policy  $\pi_{\cD}$. The Direct Method (DM) directly approximates the value function by constructing a model to predict the expected reward for each state-action pair under the target policy. The DM estimator for the policy value is formulated as:
\setlength{\abovedisplayskip}{3pt}
\setlength{\belowdisplayskip}{3pt}
\begin{small}
\begin{equation}
    \hat{V}^{\pi}_{DM} = \EE^{\pi} \hat{Q}^{\pi_{\cD}}(s, a)
\end{equation}
\end{small}
where  $\hat{Q}^{\pi_{\cD}}(s, a)$ is the estimated action-value function under the behavior policy. Despite its straightforward approach, DM relies heavily on the model's accuracy for estimating $\hat{Q}^{\pi_{\cD}}$, making it susceptible to model misspecification and bias due to the imbalanced nature of medical data. Importance Sampling (IS)\cite{tokdar2010importance} adjusts returns from the behavior policy to approximate those under the target policy. WIS mitigates IS's high variance by normalizing each weight by the sum of all weights before applying them to the returns. The WIS estimator is expressed as:
\begin{small}
    \begin{equation}
    \hat{V}^{\pi}_{WIS} = \frac{\sum_{i=1}^{N}[\rho^i_{1:T_i} G^i]}{\sum_{i=1}^{N} \rho^i_{1:T_i}}
\end{equation}
\end{small}

where $ \rho^i_{1:T_i} = \prod_{t=1}^{T_i}{\frac{\pi(a^i_t|s^i_t)}{\pi_{\cD}(a^i_t|s^i_t)}} $, and $G^i=\sum_{t=1}^{T_i}\gamma^{t-1}r_t^i$.  Modifications such as bootstrapping ($\hat{V}^{\pi}_{WIS_b}$), ratio truncation  ($\hat{V}^{\pi}_{WIS_t}$), or a combination of both ($\hat{V}^{\pi}_{WIS_{bt}}$) can further decrease variance. A comprehensive introduction is available in Appendix \ref{sec:app-ope_details}. 
The Doubly Robust (DR) estimator\cite{jiang2016doubly} combines the IS with DM, iteratively calculated per-trajectory estimation as:
\[ {\hat{V}^{T_i + 1-t}_{DR}} = \hat{V}^{\pi}(s_t^i) + \eta_t(r_t^i + \gamma \hat{V}^{H-t}_{DR} - \hat{Q}^{\pi_e}(s_t^i, a_t^i)) \]
Starting from \( t=H \) with $V^0_{DR} := 0$, this calculation proceeds in reverse order to derive \( V_{DR}^{T_i} \) for patient $i$. The overall estimate is then $\hat{V}_{DR}^{\pi} := \sum_{i=1}^{N} \hat{V}_{DR}^{T_i}$. Refer to Appendix \ref{sec:app-ope_details} for our implementation of the DR estimator.

%% file: main_sections/5reward_designs.tex
\section{Reward Design Choices}

This section examines reward design choices in RL for DTR literature. We aim to underscore the importance of reward structures in RL's effectiveness and comparability in DTR.

\subsection{Outcome-Based Reward}

The initial reward design, \cite{komorowski2018artificial}, employs a straightforward reward setting: \( r = 0 \) for non-terminal steps, \( r = +100 \) for patient survival or \( r = -100 \) for death at the final step. This binary approach oversimplifies the complexity of medical scenarios and omits critical factors such as the risk of deterioration, abnormality of the critical signs, and the progression rate of the disease, all of which significantly affect patient mortality. From an RL perspective, this approach might lead to more challenging issues such as credit assignment, sampling inefficiency and learning variance due to sparse reward. 

\subsection{Risk-based Reward}

Incorporating intermediate reward is shown to be beneficial in goal-reaching RL \cite{zhai2022computational}. In DTR applications, intermediate rewards are generally represented by clinical risk scores. Here we introduce two typical risk score-based rewards in the RL-DTR literature.

\subsubsection{ICU Risk-Based Reward}
The Sequential Organ Failure Assessment (SOFA) \cite{kajdacsy2005use, jones2009sequential} score, a commonly used critical care metric to quantify the severity of a patient's organ function or rate of failure, has been used as a reward design in the literature (\citeauthor{raghu2017deep, wang2022learning}).  In addition to the SOFA score, lactate levels are included in the reward calculation, as they are biomarkers\cite{nguyen2004early} for tissue hypoxia and metabolic dysfunction.  Represented SOFA score as $\kappa \in [0,1,2,...,24]$ and  the value of lactate (mmol/L ) as $v$, the reward is intricately defined as:
\setlength{\abovedisplayskip}{3pt}
\setlength{\belowdisplayskip}{3pt}
\begin{small}
\begin{equation}
\begin{split}
    r_t^i &= c_0 (\textbf{1}_{\kappa_t^i = \kappa_{t+1}^i} \cdot \textbf{1}_{\kappa_{t+1}^i > 0}) 
     + c_1 (\kappa_{t+1}^i - \kappa_{t}^i) \\ &\quad + c_2 \tanh(v_{t+1}^i - v_t^i) + \textbf{1}_{t=T_i} r_{\text{outcome}}
\end{split}
\end{equation}
\end{small}
where \( c_0, c_1, c_2 \) is -0.025, -0.125, and -2, respectively. $r_{outcome}$ is 15 for an alive patient; otherwise -15. This formula integrates both stability and changes in organ function (using the SOFA score) and metabolic alterations (using lactate levels).

\subsubsection{Early Warning Risk-Based Reward}

We also present a reward function based on the National Early Warning Score 2 (NEWS2) \cite{inada2018news}  system to keep in line with the latest medical applications \footnote{Recent clinical studies \cite{sivayoham2021observational, mellhammar2019news2} indicates the possible advantage of using NEWS2 against SOFA score in managing sepsis.}.  We normalize NEWS2 to a range of [0,1], representing the probability of mortality. $r_{\text{outcome}}$ is set to -1 in the event of death and to 0 otherwise. The reward is then
\begin{equation}
    r_{t}^i = -r_{\text{NEWS2}} +\textbf{1}_{t=T_i} r_{\text{outcome}}
\end{equation}
This normalization creates a consistent focus on mortality and eliminates the need for tuning weights between different reward components. 


%% file: main_sections/6baselines.tex
\section{Baselines Comparisons}

This section addresses the selection of baselines. Upon reviewing the literature, we observed several inconsistencies: \textbf{1)} the use of different baseline sets across studies, with some lacking state-of-the-art (SOTA) offline RL algorithms; \textbf{2)} the absence of naive baselines, such as random policy, for essential sanity checks; \textbf{3)} the omission of supervised learning baselines. Consequently, this section outlines a set of baselines that we consider appropriate for comparison.

\textbf{Supervised learning baselines:} Using Supervised Learning (SL) algorithms as a benchmark is crucial to determining RL's benefits in DTR. Comparing RL with simpler SL algorithms helps us understand if the complexities of RL lead to better results. Although offline RL may not always perfectly match clinicians' actual decisions, its performance should be close to real-world outcomes (as indicated by clinician decisions in test data). We employ a Long-Short-Term Memory (LSTM) \cite{hochreiter1997long} network to minimize cross-entropy loss, a standard loss function for classification tasks (see the Appendix \ref{sec:app-implementation of bc}). 

\textbf{Naive baselines:} We include random $\pi_r$, zero-drug $\pi_\text{min}$, max-drug $\pi_\text{max}$, and alternating policy $\pi_\text{alt}$ (elaborated in the section \ref{sec:app-naive_baselines}). While often neglected in RL-DTR research, naive baseline comparison is vital for two reasons: \textbf{1)} to assess if naive baselines inadvertently outperform clinicians, indicating potential flaws in evaluation metrics; \textbf{2)} to establish a lower performance bond, providing a worst-case scenario benchmark for algorithmic assessment.

\textbf{Deep Q-Network (DQN)}: DQN\cite{mnih2015human} integrates neural networks to approximate the Q-function with a target network and Experience Replay Buffer.  Although DQN was originally designed for online RL, one can easily adapt it to the offline case where the replay buffer is fixed. The objective function for DQN can be expressed as:
\begin{equation}
\arg\min_{\theta} \mathbb{E}_\cD\left[ \left( r + \gamma \max_{a'} Q_{\theta'}(s', a') - Q_\theta(s, a) \right)^2 \right]
\end{equation}
where \(\theta\) and \(\theta'\) are the parameters of the current and target Q-networks, respectively.

\textbf{Conservative Q-Learning (CQL):} The primary objective of CQL\cite{kumar2020conservative} is to construct a conservative estimate of the Q function, intentionally underestimating the Q values. This is achieved by incorporating an additional regularization term into the standard Bellman update equation. The loss function $L_{CQL}(\theta)$ for CQL is given by:
\begin{small}
\begin{equation*}
L_{\text{DQN}}(\theta) - \lambda \mathbb{E}_{\cD} \bigg[ \log \sum_{a} \exp(Q_\theta(s, a))
- \mathbb{E}_{a \sim \pi(\cdot|s)}[Q_\theta(s, a)] \bigg]
\end{equation*}
\end{small}
Here, \(\lambda\) is a regularization parameter, and \(L_{DQN}(\theta)\) is DQN's loss function. 

\textbf{Batch Constrained Q-Learning (BCQ)}: BCQ\cite{fujimoto2019offpolicy} ensures the action is in-distribution by introducing a Varational Auto-encoder \cite{sohn2015learning} structure. The objective function for BCQ can be formulated as:

\begin{small}
\begin{equation}
\arg\min_{\theta} \mathbb{E}_{\cD}\! \left[ \left( r\! +\! \gamma \max_{a' \in \mathcal{A}_{\phi}(s')} Q_{\theta'}(s', a') \!-\! Q_\theta(s, a) \right)^2 \right]
\end{equation}
\end{small}
where \(\mathcal{A}_{\phi}(s')\) denotes the set of actions similar to the training batch, generated by a VAE with parameters \(\phi\).

\textbf{Implicit Q-Learning (IQL)}: IQL focuses on directly optimizing the Q-function without explicitly defining a policy\cite{kostrikov2022offline}. This approach aims to improve learning efficiency and stability. The update equation for IQL can be written as:
\begin{small}
\begin{equation}
\arg\min_{\theta} \mathbb{E}_{\cD} \left[ \left( r + \gamma \mathbb{E}_{a' \sim \pi_{\theta}}[Q_{\theta'}(s', a')] - Q_\theta(s, a) \right)^2 \right]
\end{equation}
\end{small}
Here, \(\pi_{\theta}\) denotes the implicit policy derived from the Q-function parameterized by \(\theta\).

As the domain of offline RL rapidly evolves, newer algorithms\cite{fujimoto2021minimalist,xiao2023inac} are emerging as more advanced alternatives but are not reviewed here.

%% file: main_sections/7result.tex
\input{rl_tables/all/MIMIC3SepsisNEWS2Env-all_train-all-table}
\input{rl_tables/sub/MIMIC3SepsisNEWS2Env-all_train-sub-test_rate_--0.4_high_std-table}

\section{Experiments}

This section presents empirical findings using the sepsis dataset, illustrating the significant impact of varying policy evaluation algorithms, metrics, and reward design. The data set is divided into training, validation, and testing sets, comprising 70\%, 15\%, and 15\% of the data, respectively. This partitioning adheres to the patient stratification method detailed in Appendix \ref{sec:app-data description}. For hyperparameter optimization, grid search is performed in a unified search space. Please see \cref{sec:app-hyperparameter search} for any missing details. In addition to the main result below, Appendix \ref{sec:app-full result} provides the test set performance using Outcome and SOFA reward in Table \ref{table: Outcome-all} and \ref{table: SOFA-all}, with additional 35 experiments on subsets.

\subsection{Overall Comparison Results}
\label{sec: result-overall comparison}
We compare naive baselines, supervised learning, and RL (10 policies in total) on 9 metrics with 13 patient groups and 3 different reward designs, resulting in 17,550 evaluation experiments. In addition to the performance comparison in the overall test set, we stratified the test set according to clinical outcome and risks, producing 12 subsets. Due to space limitations, only selected figures and tables are shown here, with complete results in Appendix \ref{sec:app-full result}.

To compare the performance across different reward designs and policy evaluation metrics in a straightforward way, we used a measure called "Number of Wins" (No. Wins). This measure counts how many times one algorithm outperforms all others. A win is counted when the algorithm is the best on a metric for a particular patient group with a specific reward setting.
From our experiments, it is evident that the best algorithm varies across settings; moreover, although some algorithms achieve wins under policy evaluation methods such as WIS and DR, they may still behave unreasonably, i.e., deviating significantly from a doctor's policy, as indicated by metrics like RMSE or F1. We summarize the key observations below.

\begin{figure}[h]
    \centering
    \begin{subfigure}[b]{0.45\textwidth}
        \centering
        \includegraphics[width=0.99\linewidth]{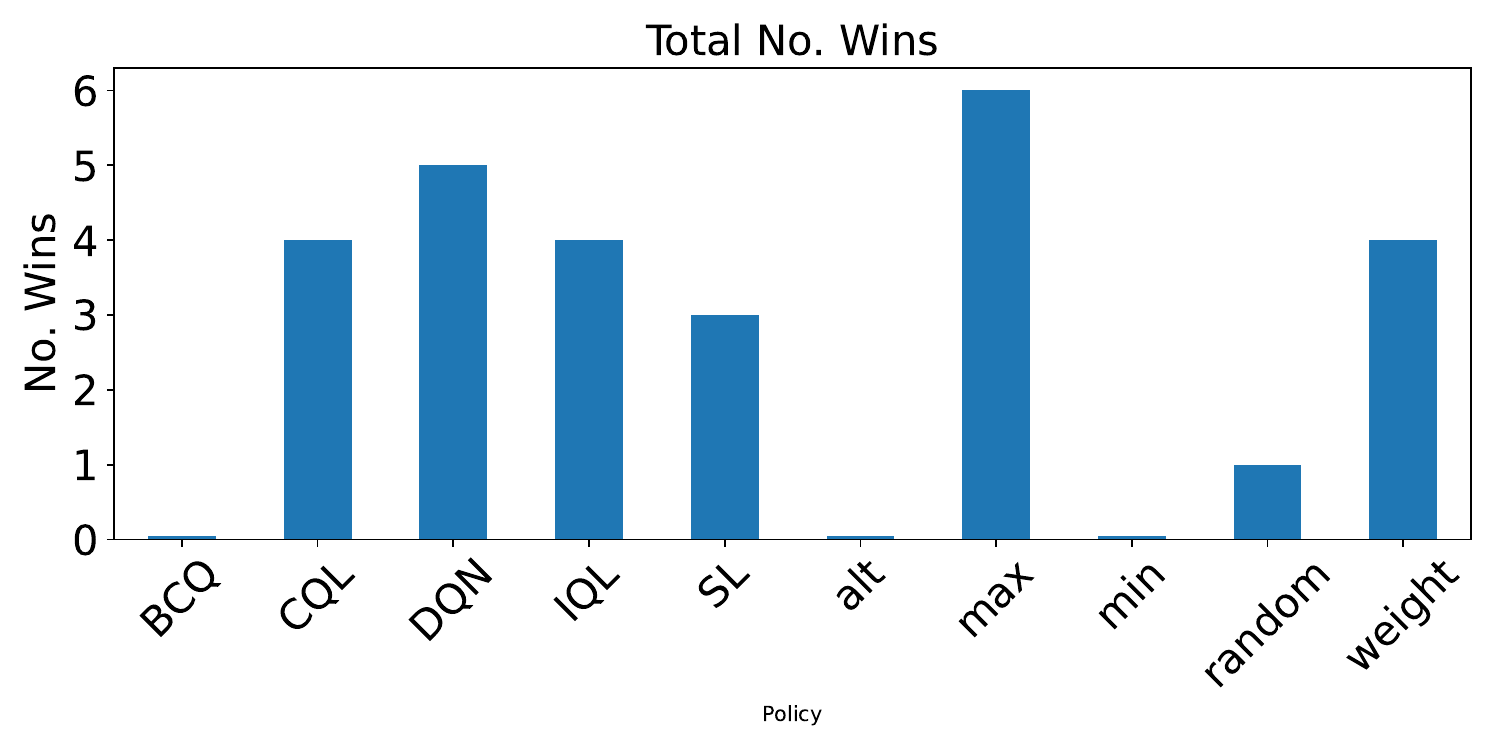}
        \vspace{-0.7cm}
        \caption{Overall No. wins.}
        \label{fig:wins-a}
    \end{subfigure}%
    \hfill
    \begin{subfigure}[b]{0.45\textwidth}
        \centering
        \includegraphics[width=\linewidth]{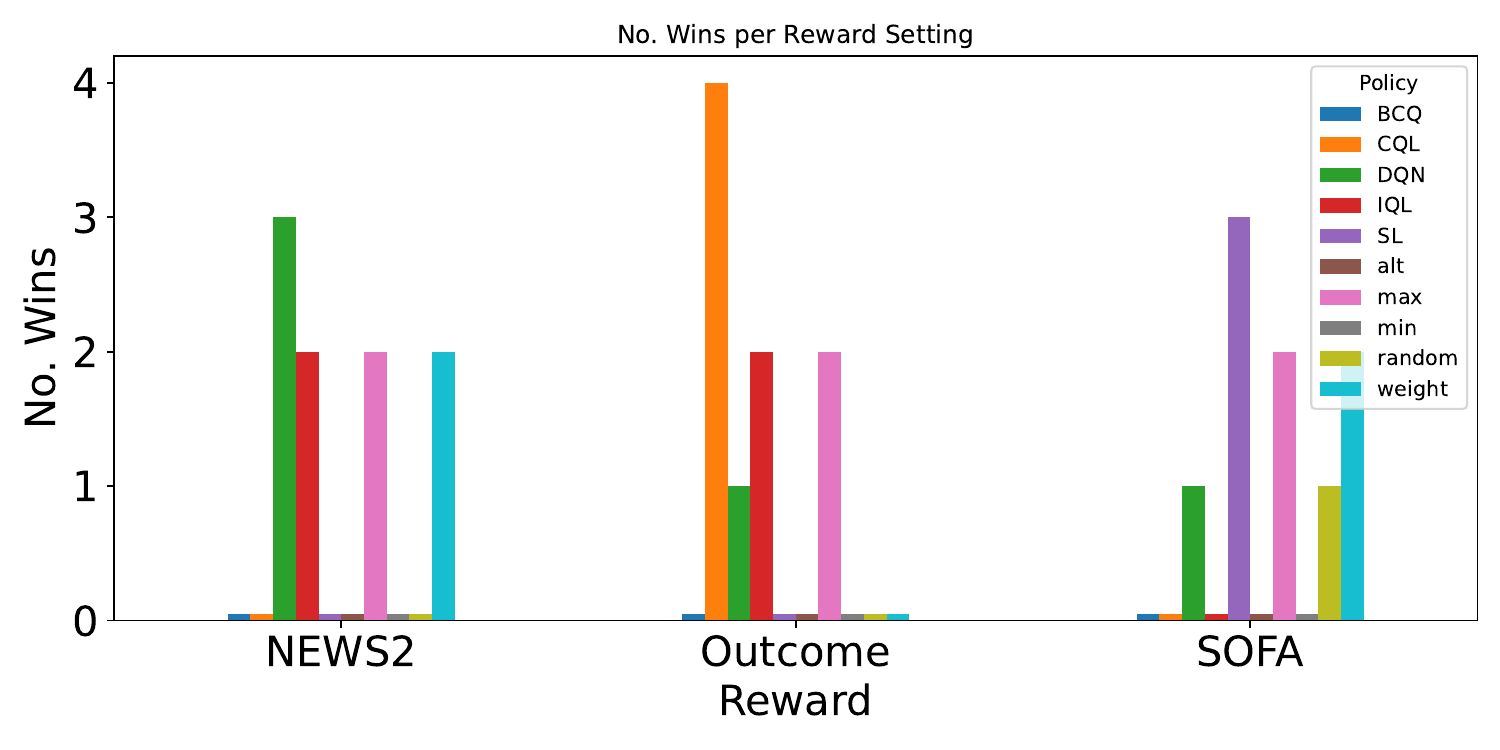}
 
              \caption{No. wins under the three respective reward settings.}
               \label{fig:wins-b}
    \end{subfigure}
    \caption{\small \textbf{Number of wins for each policy in the (overall) test set.} Wins are calculated based on the mean performance of 5 random seeds. Alt, min, max, random, and weight policies are naive baselines. This denotation applies to all the following figures. }
    \label{fig:wins1}
    \vspace{-0.3cm}
\end{figure}

\textbf{Effectiveness of Naive Baselines:} Surprisingly, the naive baselines have wins over RL, SL, and even the doctor returns in the overall test set, as shown in Figure \ref{fig:wins-a} and Figure \ref{fig:wins-b}. Intuitively, naive baselines should not win even once. Table \ref{table: NEWS2-all} reveals that the weight policy performed better than all RL, SL, and even the doctor return in the overall test set on WIS and $\text{WIS}_{t}$. However, this trend is inconsistent with the result in Table \ref{table: Outcome-all}, where RL algorithms generally surpass naive baselines. These findings again support our recommendation for including naive baselines as an easy check to the reliability of OPE methods.


\textbf{RL Performances Across Rewards:} We observed significant differences in how RL algorithms perform with different rewards. For instance, DQN won the most on the NEWS2 reward, while CQL won the most on the outcome reward (see Figure \ref{fig:wins-b}). Yet, on the SOFA reward, all RL policies won less than SL, making it difficult to identify a consistently performing superior algorithm across various reward settings.

\textbf{RL vs SL:} When RL models outperform SL on OPE metrics, it is reasonable to anticipate that RL should also demonstrate comparable performance to SL on supervised learning metrics. However, our findings present a more complex scenario. There are instances where RL's superiority in OPE metrics does not translate to better performance like RMSE or F1 score (see Table \ref{table: NEWS2-all}'s DQN column), while we also observe the opposite cases (see Table \ref{table: NEWS2-all}'s CQL column, WIS, $\text{WIS}_b$ and $\text{WIS}_t$ rows). This discrepancy raises critical questions about the effectiveness and reliability of OPE methods in evaluating RL models.

\begin{figure}[ht]
    \begin{subfigure}[b]{0.45\textwidth}
        \centering
        \includegraphics[width=0.99\linewidth]{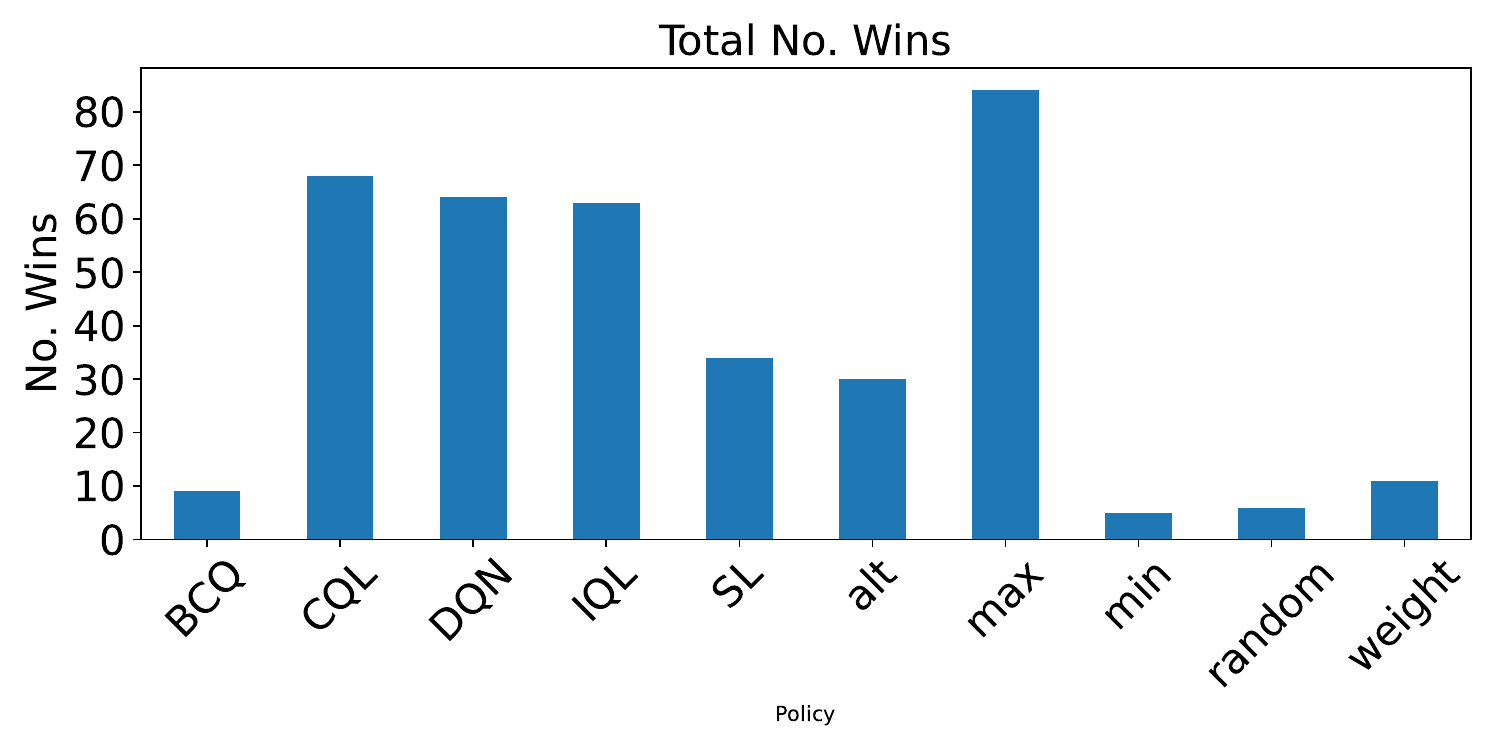}
              \label{fig:wins-c}
               \vspace{-0.7cm}
              \caption{Sum of No. wins across patient subgroups.}
    \end{subfigure}%
    \hfill
    \begin{subfigure}[b]{0.45\textwidth}
        \centering
        \includegraphics[width=0.99\linewidth]{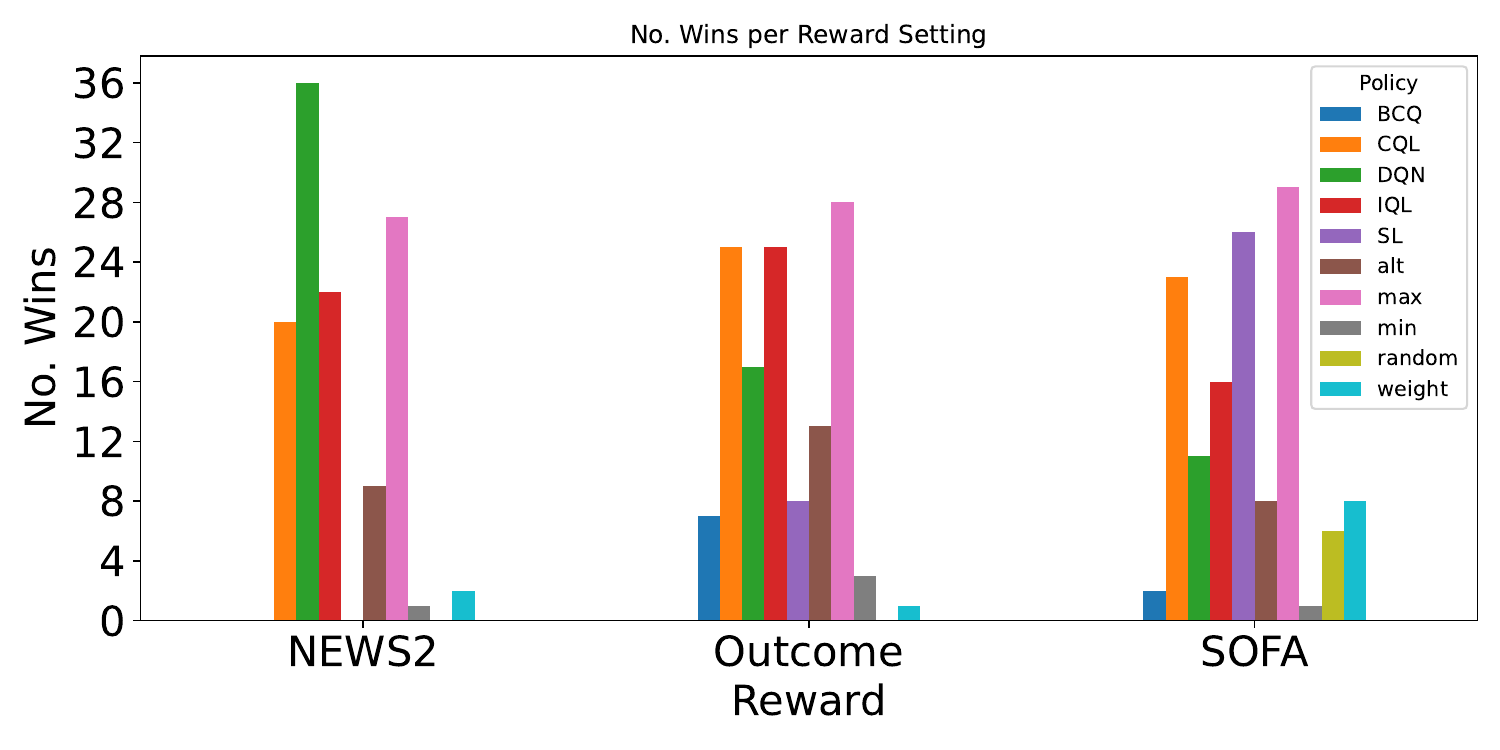}
              \label{fig:wins-d}
              \vspace{-0.3cm}
              \caption{Sum of No. wins on stratification groups under the three respective reward settings.}
    \end{subfigure}
     \vspace{-0.3cm}
    \caption{\small 
    \textbf{A summed number of wins across patient subgroups stratified by mortality risk rate of change.} This figure presents the cumulative performance of each algorithm, measured by the No. win across 12 stratified subsets derived from the test set. Wins are calculated for each algorithm within each subset across all metrics and subsequently aggregated to reflect overall performance. This approach allows for an assessment of the average algorithmic efficacy in various subgroups of patients, stratified by changes in mortality risk.} 
    \label{fig:win2}
    \vspace{-0.3cm}
\end{figure}

\textbf{Comparison of DR and IS:} Our analysis indicated that DR either overestimates or underestimates IS (see Table \ref{table: NEWS2-all} , and Table \ref{table: Outcome-all}, \ref{table: SOFA-all} in Appendix,  and compare the DR row with other importance sampling rows.), subject to the reward setting. Instead of being doubly robust, our experiments suggest that the estimators in DR tend to dual ‘unrobust’ due to behavioral and value approximation error (further discussion in Section \ref{sec: discussion-bc and value}.). This finding calls for a reassessment of the robustness claims of DR methods in healthcare contexts.

\textbf{RL outperforms SL on particular patient cohorts:} We also found RL outperforms SL on both OPE metrics and supervised learning metrics for specific patient subgroups. A notable illustration can be seen in Table \ref{table: NEWS2 reward-subgroup-test_rate_--0.4_high_std}, particularly when comparing the performance of SL with CQL. In this comparison, CQL outperforms SL across nearly all metrics and achieves similar results to SL in the DR metric. This finding highlights RL's capacity to develop more effective treatment strategies than SL for certain patient groups, demonstrating its potential for individualized treatment solutions.

\subsection{Understanding the Variance of Policy Evaluation}

This section aims to offer an in-depth study of the causes of variance in policy evaluation (or at least a portion of these causes). Intuitively, the policy evaluation can be unreliable when the behavioral policy or target value function does not approximate well with the real one. We illustrate this point of view by visualizing the inference probability/value of the behavioral ($\hat{\pi}_{\cD}$)/value ($\hat{Q}^{\pi}$) estimators against their sample losses.

\begin{figure*}[t]
\begin{subfigure}[b]{0.24\textwidth}
    \centering
    \includegraphics[width=\linewidth]{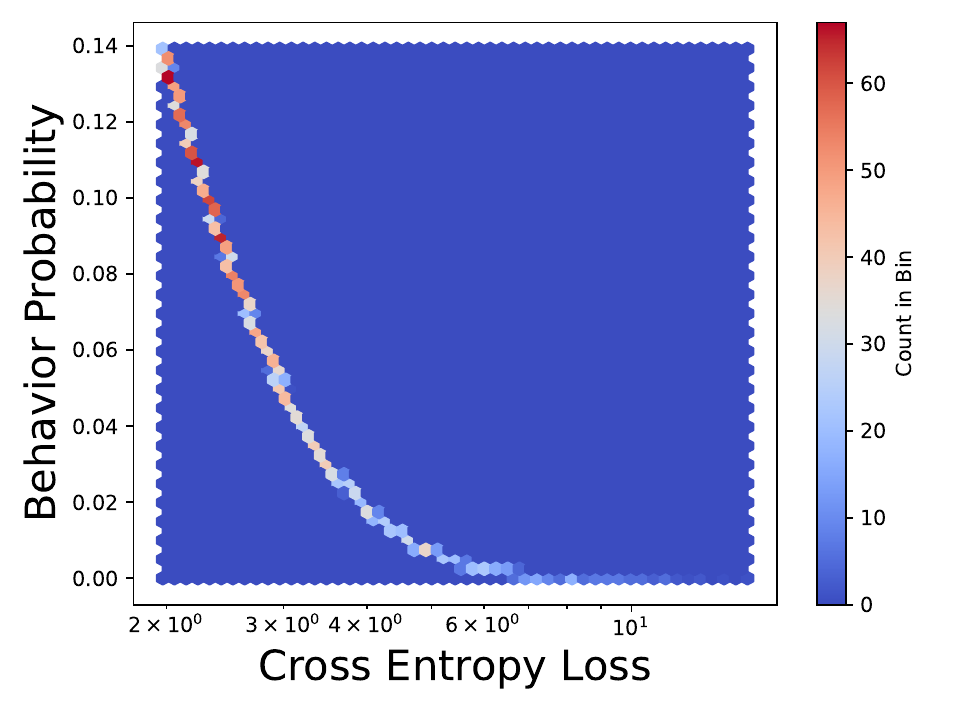}
    \caption{$\hat{\pi}_{\cD}$}
    \label{fig:bc_vs_dm-policy}
\end{subfigure}
 \hfill
\begin{subfigure}[b]{0.24\textwidth}
    \centering
    \includegraphics[width=\linewidth]{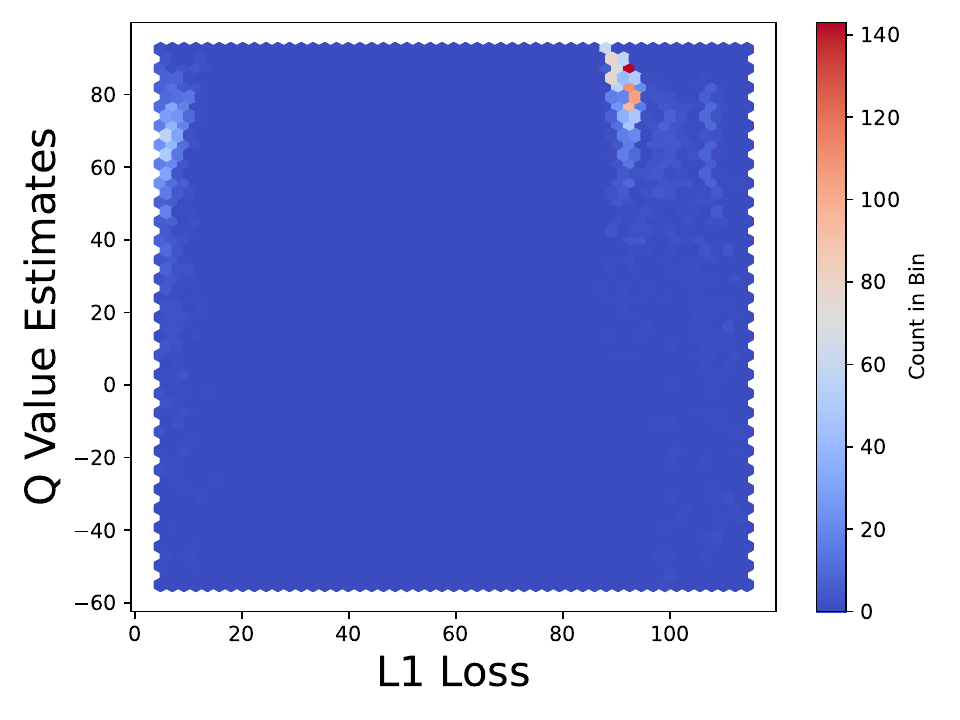}
    \caption{$\hat{Q}^{\pi}$ on the outcome reward}
    \label{fig:bc_vs_dm-outcome value}
\end{subfigure}
\hfill
\begin{subfigure}[b]{0.24\textwidth}
    \centering
    \includegraphics[width=\linewidth]{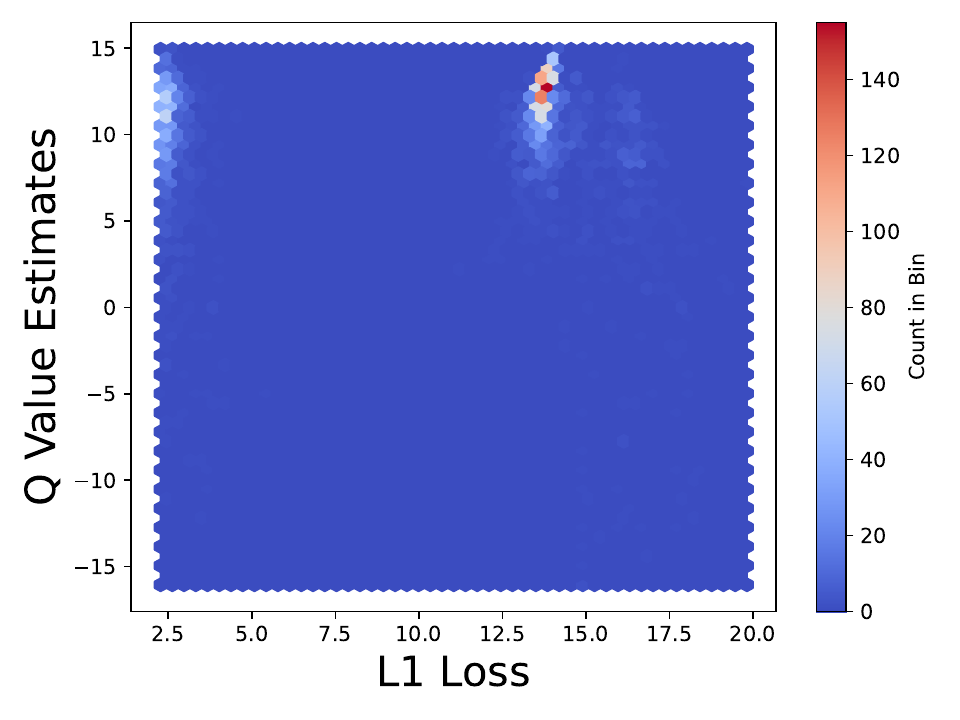}
    \caption{$\hat{Q}^{\pi}$ on the SOFA reward}
        \label{fig:bc_vs_dm-SOFA value}
\end{subfigure}
\hfill
\begin{subfigure}[b]{0.24\textwidth}
    \centering
    \includegraphics[width=\linewidth]{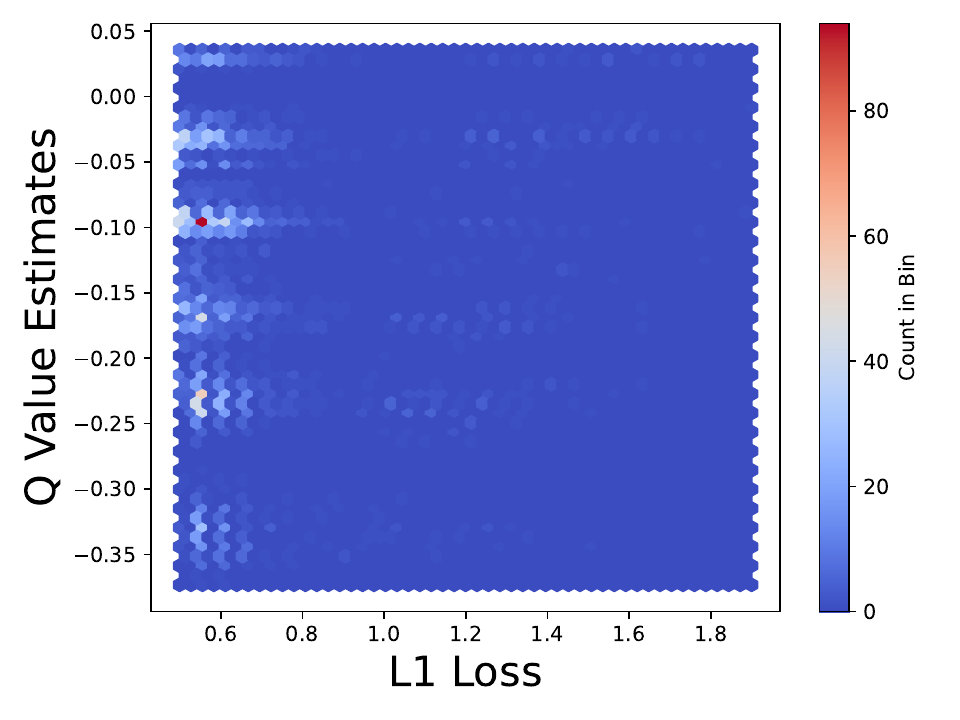}
    \caption{$\hat{Q}^{\pi}$ on the NEWS2 reward}
        \label{fig:bc_vs_dm-NEWS2 value}
\end{subfigure}
\centering
\caption{\small \textbf{Behavioral and value estimator versus their losses on the testing set.} The count in each bin is indicated by a colour bar, transitioning from blue to red as the number increases. \textbf{(a)} depicts the behavioral loss (samples with a cross-entropy loss $>$ 90th percentile ) versus the inference probability. \textbf{(b)}, \textbf{(c)}, and \textbf{(d)} show the direct method estimator loss (samples with L1 loss $>$90th percentile) on the outcome, SOFA, and NEWS2 reward, respectively.
\label{fig: bc_vs_dm}
}
 \vspace{-0.1cm}
\end{figure*}

\begin{figure*}
\begin{subfigure}[b]{0.3\textwidth}
    \centering
    \includegraphics[width=\linewidth]{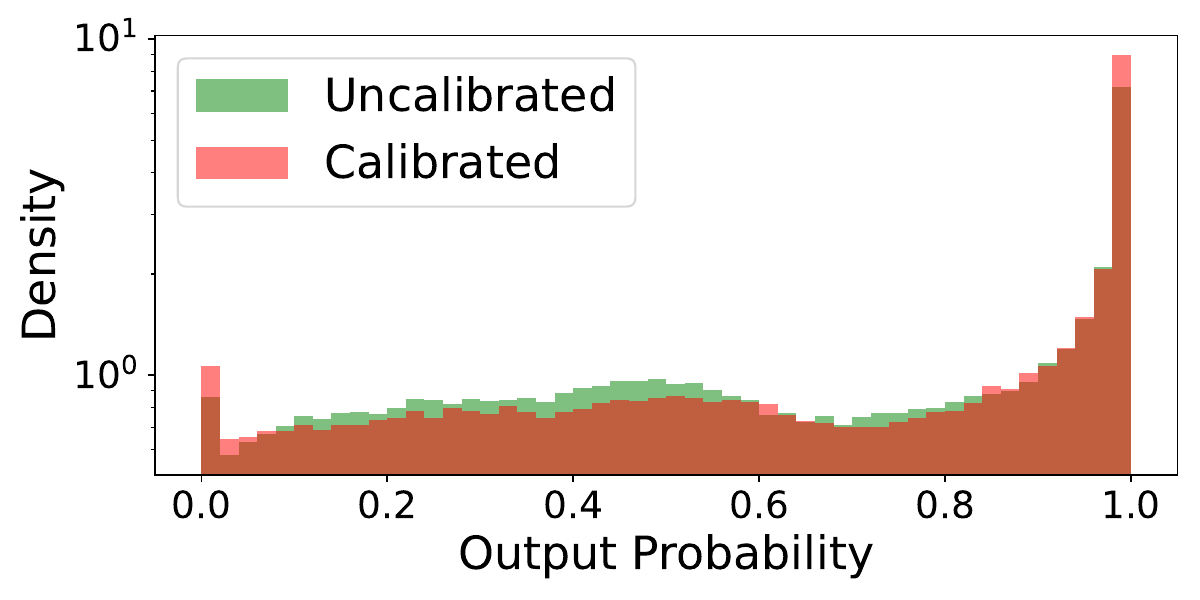}
    \caption{Training set}
\end{subfigure}
\hfill
\begin{subfigure}[b]{0.3\textwidth}
    \centering
    \includegraphics[width=\linewidth]{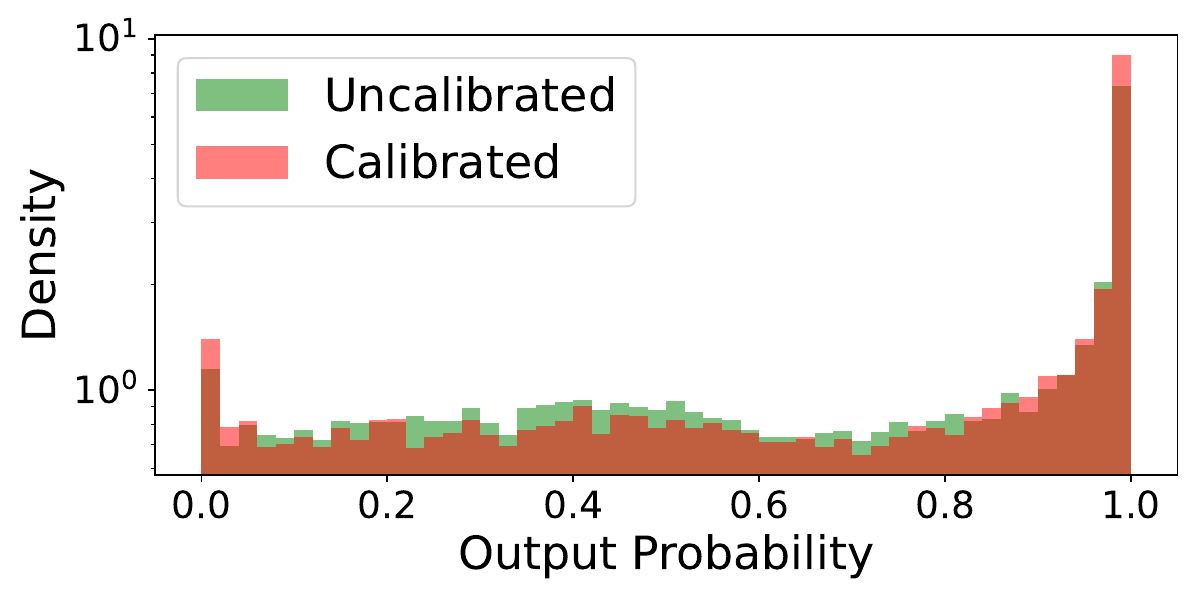}
    \caption{Validation set}
\end{subfigure}
\hfill
\begin{subfigure}[b]{0.3\textwidth}
    \centering
    \includegraphics[width=\linewidth]{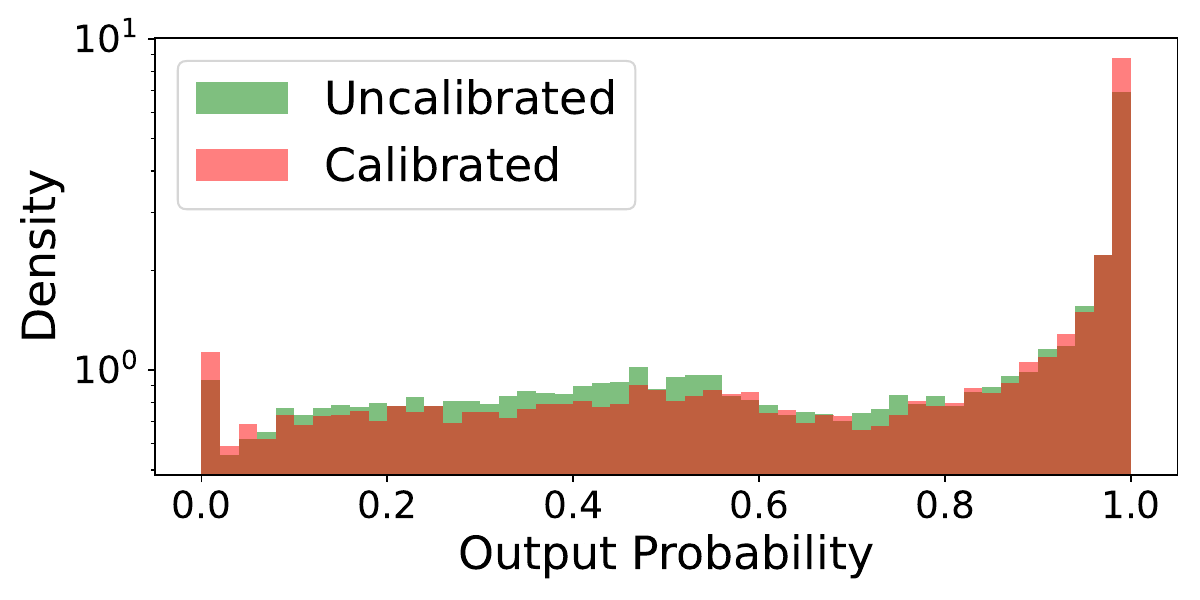}
    \caption{Test set}
\end{subfigure}
\caption{\textbf{Comparison of output probability between calibrated and uncalibrated $\hat{\pi_{\mathcal{D}}}$.} The plot shows a histogram of output probability and the number of counts in the dataset with a logarithm scale on the y-axis on training, validation and test set, respectively. It is observed that the frequencies of extreme probabilities (i.e., probabilities near 0 and 1) are higher after calibration.}
\label{fig:calibration prob}
\end{figure*}

\label{sec: discussion-bc and value}

\subsubsection{Variance from $\hat{\pi}_{\cD}$ }
\label{sec: variance from pi}

A common assumption in OPE research is $\hat{\pi}_{\cD} \simeq \pi_{\cD}$. We challenged this assumption within the DTR setting by analyzing testing samples with significant loss, as shown in Figure \ref{fig: bc_vs_dm}. The variance in $\hat{\pi}_{\cD}$ can be understood through two critical observations:

\textbf{Trajectories with small behavior probabilities determine importance sampling:} Importance sampling gives more weight to the trajectory with large ratios, and large ratios are more likely to be produced by trajectories with small behavior probabilities on the denominator. Therefore, it is critical for a  $\hat{\pi}_{\cD}$ to produce small probabilities correctly since such large errors can severely distort OPE estimates when paired with small probabilities.

\textbf{Negative correlation between errors and inference probabilities:}  Smaller probabilities, which contribute more to the importance sampling, are often associated with higher loss, indicating greater deviation (see Figure \ref{fig:bc_vs_dm-policy}).

\subsubsection{Variance from $\hat{Q}^{\pi}$} 
\textbf{Quality of value approximation:} The quality of learning $\hat{Q}^{\pi}$ depends on reward settings, which lead to inconsistent DR policy estimation results (as discussed in Section \ref{sec: result-overall comparison}) This insight can be validated by Fig \ref{fig:bc_vs_dm-outcome value}, \ref{fig:bc_vs_dm-SOFA value} and \ref{fig:bc_vs_dm-NEWS2 value} across three reward settings.

\textbf{Large errors center around high Q estimates:} Large errors are concentrated around high Q estimates for both the Outcome reward ( $r_{\text{alive/death}}=\pm 100$) and SOFA reward($r_{\text{alive/death}}=\pm15$) as shown in  Figure \ref{fig: bc_vs_dm} (b) and (c), respectively). This suggests that the value estimator does not adequately capture the termination reward (i.e., whether the patient survives or dies). In contrast, the NEWS2 reward does not exhibit this pattern, as its reward function is smoother, ranging from -1 to 0, with -1 indicating death and 0 otherwise. The new experiments support our position of reevaluating RL-DTR. Specifically, careful design of the reward function is crucial not only for clinical meaningfulness but also for facilitating learning value estimators for OPE.

\subsubsection{Investigating Model Calibration for $\hat{\pi}_{\cD}$}
Model calibration refines a model's predictive probabilities to match the actual outcome likelihoods, thereby ensuring the model's confidence reflects its empirical accuracy. We investigated the influence of model calibration on behavior policy and performance change in OPE. In this study, the temperature scale \cite{guo2017calibration} is used as the model calibrator. Implementation details and hyperparameter choice for temperature scaling are provided in Appendix \ref{sec:app-model calibration}.

\begin{figure*}[h]
\centering
\includegraphics[width=0.9\linewidth]{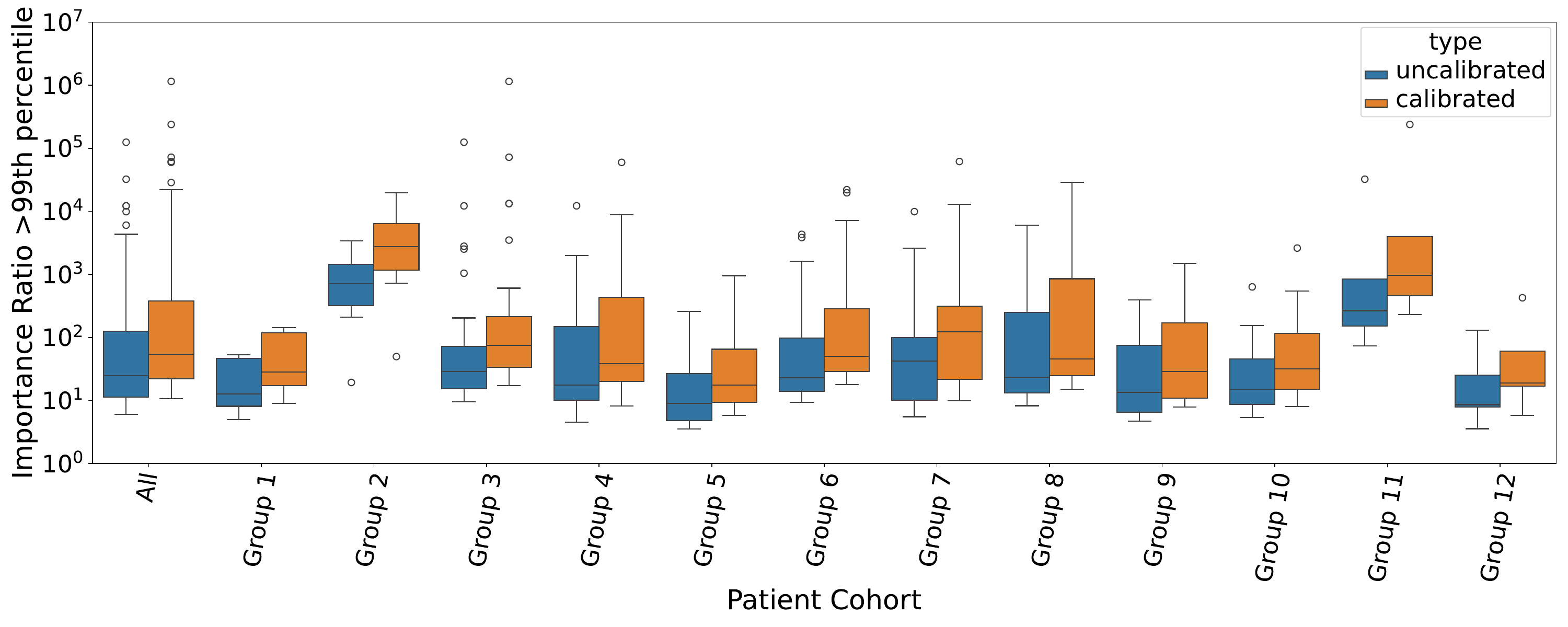}
\caption{\textbf{Importance ratio histogram of random policy $>$ 99th percentile.} The horizontal axis includes different datasets, where 'All' means the test set and the rest are NEWS2 risk-stratified subsets, indexed by the ascending order of NEWS2 change rate. The calibrated model contains more extremely large ratios $>$ 99th percentile. Only ratio outliers (i.e., $>$ 99th percentile) are plotted for visualization convenience. To view the other 14 ratio plots for 5 baseline policies in 3 different reward settings, please see Appendix \ref{sec:app-ratio plot}.}
\label{fig: raw ratio random}
\end{figure*}

Figure \ref{fig:calibration prob} shows the probability distribution before and after calibration. It can be seen that more probabilities are distributed near 0 and 1 after calibration. As discussed in Section \ref{sec: variance from pi}, importance sampling gives small probabilities more weight, and small probabilities can lead to extremely high losses, indicating that calibration may increase OPE variance in this case. To further illustrate this point, we plotted the importance ratio (i.e., $ \rho^i_{1:T_i} = \prod_{t=1}^{T_i}{\frac{\pi(a^i_t|s^i_t)}{\pi_{\cD}(a^i_t|s^i_t)}} $) of the random policy $\pi_r$ in Figure \ref{fig: raw ratio random}. Large ratios became even larger after calibration, which implies that model calibration may increase the variance of OPE and should be applied with caution. An additional 14 figures for baseline ratio comparison are reported in the Appendix \ref{sec:app-ratio plot}.

To further investigate the impact of model calibration on OPE, we ran OPE for all naive baselines using the calibrated behavior policy. The results are presented in Tables \ref{table: calibrated-Outcome-all}, \ref{table: calibrated-SOFA-all} and \ref{table: calibrated-NEWS2-all} in the Appendix \ref{sec:app-model calibration}. Since we do not have access to the ground truth reward estimates for the naive baselines, we use the criterion 'higher than $G_\cD$' as a sanity check: \textbf{If a naive baseline can surpass the performance of clinicians according to the OPE results, it suggests that the OPE method may not be reliable.} This is because we expect clinicians who have extensive domain knowledge to generally outperform naive baselines. The results show that some naive baselines can still surpass clinical experts, regardless of reward design. This result again supports our position of reevaluating DTR and indicates that model calibration may not be universally helpful in DTR. 


%% file: rl_tables/all/MIMIC3SepsisNEWS2Env-all_train-all-table.tex
\begin{table*}[ht]
\centering
\tiny
\begin{tabular}{lcccccccccc}
\toprule
metric & alt & max & min & random & weight & SL & DQN & CQL & IQL & BCQ \\
\midrule
$\text{RMSE}_{\text{IV}}$ & ${763.89}$ & ${861.51}$ & ${645.83}$ & ${671.39}$ & ${645.83}$ & \textcolor{blue}{$557.81 \pm 9.27$} & ${638.51 \pm 8.63}$ & \textcolor{red}{$541.67 \pm 5.74$} & ${578.96 \pm 10.06}$ & ${626.2 \pm 9.56}$ \\
$\text{RMSE}_{\text{vaso}}$ & ${0.67}$ & ${0.89}$ & ${0.32}$ & ${0.5}$ & ${0.59}$ & \textcolor{blue}{$0.31$} & ${0.44 \pm 0.07}$ & \textcolor{red}{$0.3 \pm 0.01$} & \textcolor{blue}{$0.31 \pm 0.01$} & \textcolor{blue}{$0.31$} \\
$\text{WIS}$ & ${-4.58}$ & ${-4.62}$ & ${-4.58}$ & ${-3.84}$ & \textcolor{red}{$-3.78$} & ${-4.22 \pm 0.41}$ & \textcolor{blue}{$-3.79 \pm 0.01$} & ${-4.1 \pm 1.43}$ & ${-5.83}$ & ${-4.58}$ \\
$\text{WIS}_{b}$ & ${-5.43}$ & ${-4.81}$ & ${-5.76}$ & \textcolor{blue}{$-4.4$} & ${-4.73}$ & ${-4.62 \pm 0.17}$ & \textcolor{red}{$-3.88 \pm 0.73$} & ${-4.48 \pm 0.77}$ & ${-5.31 \pm 0.06}$ & ${-5.41 \pm 0.17}$ \\
$\text{WIS}_{t}$ & ${-4.58}$ & ${-4.62}$ & ${-4.58}$ & ${-3.97}$ & \textcolor{red}{$-3.78$} & ${-4.57 \pm 0.62}$ & \textcolor{blue}{$-3.84 \pm 0.11$} & ${-4.1 \pm 1.43}$ & ${-5.83}$ & ${-4.58}$ \\
$\text{WIS}_{bt}$ & ${-5.64}$ & ${-4.69}$ & ${-5.61}$ & ${-4.5}$ & ${-4.5}$ & ${-4.68 \pm 0.3}$ & \textcolor{red}{$-3.87 \pm 0.67$} & \textcolor{blue}{$-4.38 \pm 0.98$} & ${-5.27 \pm 0.05}$ & ${-5.55 \pm 0.19}$ \\
DR & ${-0.54}$ & \textcolor{blue}{$-0.19$} & ${-1.55}$ & ${-0.35}$ & ${-0.3}$ & ${-0.36}$ & \textcolor{red}{$-0.14 \pm 0.04$} & ${-0.71 \pm 0.05}$ & ${-0.51 \pm 0.04}$ & ${-1.54 \pm 0.01}$ \\
P.F1 & ${0.2}$ & ${0.02}$ & ${0.2}$ & ${0.2}$ & ${0.0}$ & ${0.31 \pm 0.01}$ & ${0.06 \pm 0.02}$ & \textcolor{blue}{$0.33 \pm 0.01$} & \textcolor{red}{$0.34 \pm 0.01$} & ${0.23 \pm 0.01}$ \\
S.F1 & ${0.19}$ & ${0.02}$ & ${0.19}$ & ${0.19}$ & ${0.0}$ & ${0.3 \pm 0.01}$ & ${0.06 \pm 0.02}$ & \textcolor{blue}{$0.32 \pm 0.01$} & \textcolor{red}{$0.33 \pm 0.01$} & ${0.22 \pm 0.01}$ \\
\hline
$G_{\cD}$ & \multicolumn{10}{c}{-4.39}\\
\bottomrule

\end{tabular}
\caption{\textbf{Comparison across policies on the test set using NEWS2 reward.} The best and second-best algorithms are highlighted in red and blue, respectively. $\text{RMSE}_{\text{IV}}$ and $\text{RMSE}_{\text{vaso}}$ mean the RMSE loss for the IV fluid treatment and vasopressor treatment. The predicted action class is mapped to continuous values by taking the median of the bin, as detailed in Appendix \ref{sec:app-data description}. P.F1 and S.F1 denote the patient-wise F1 and sample-wise F1. The above denotation applies to all the following tables.}
\label{table: NEWS2-all}
\end{table*}

%% file: rl_tables/sub/MIMIC3SepsisNEWS2Env-all_train-sub-test_rate_--0.4_high_std-table.tex
\begin{table*}[ht]
\centering
\tablestyle{4pt}{1.0}
\tiny
\begin{tabular}{lcccccccccc}
\toprule
metric & alt & max & min & random & weight & SL & DQN & CQL & IQL & BCQ \\
\midrule
$\text{RMSE}_{\text{IV}}$ & ${788.91}$ & ${880.8}$ & ${756.41}$ & ${774.78}$ & ${749.58}$ & \textcolor{blue}{$637.22 \pm 10.04$} & ${734.32 \pm 11.14}$ & \textcolor{red}{$609.62 \pm 11.74$} & ${645.67 \pm 6.1}$ & ${719.77 \pm 12.57}$ \\
$\text{RMSE}_{\text{vaso}}$ & ${0.54}$ & ${0.85}$ & ${0.27}$ & ${0.47}$ & ${0.56}$ & \textcolor{red}{$0.25$} & ${0.4 \pm 0.07}$ & \textcolor{blue}{$0.26 \pm 0.01$} & \textcolor{red}{$0.25 \pm 0.01$} & \textcolor{blue}{$0.26$} \\
$\text{WIS}$ & ${-3.23}$ & ${-3.51}$ & ${-3.44}$ & ${-3.14}$ & ${-3.38}$ & ${-3.01 \pm 0.07}$ & \textcolor{blue}{$-2.85 \pm 0.72$} & \textcolor{red}{$-1.9 \pm 0.36$} & ${-2.98 \pm 0.13}$ & ${-3.44}$ \\
$\text{WIS}_{b}$ & ${-3.33}$ & ${-3.49}$ & ${-3.41}$ & ${-3.23}$ & ${-3.4}$ & ${-2.89 \pm 0.06}$ & ${-2.85 \pm 0.5}$ & \textcolor{red}{$-2.09 \pm 0.19$} & \textcolor{blue}{$-2.84 \pm 0.02$} & ${-3.34 \pm 0.02}$ \\
$\text{WIS}_{t}$ & ${-3.23}$ & ${-3.51}$ & ${-3.44}$ & ${-3.14}$ & ${-3.38}$ & ${-3.01 \pm 0.07}$ & \textcolor{blue}{$-2.85 \pm 0.72$} & \textcolor{red}{$-1.9 \pm 0.36$} & ${-2.98 \pm 0.13}$ & ${-3.44}$ \\
$\text{WIS}_{bt}$ & ${-3.29}$ & ${-3.47}$ & ${-3.42}$ & ${-3.21}$ & ${-3.41}$ & ${-2.93 \pm 0.05}$ & \textcolor{blue}{$-2.81 \pm 0.49$} & \textcolor{red}{$-2.1 \pm 0.24$} & ${-2.86 \pm 0.05}$ & ${-3.34 \pm 0.05}$ \\
DR & ${-0.26}$ & \textcolor{red}{$-0.05$} & ${-1.46}$ & ${-0.39}$ & ${-0.46}$ & ${-0.41 \pm 0.02}$ & \textcolor{blue}{$-0.13 \pm 0.09$} & ${-0.48 \pm 0.13}$ & ${-0.43 \pm 0.02}$ & ${-1.27 \pm 0.03}$ \\
P.F1 & ${0.25}$ & ${0.02}$ & ${0.25}$ & ${0.25}$ & ${0.0}$ & ${0.31 \pm 0.01}$ & ${0.07 \pm 0.02}$ & \textcolor{blue}{$0.33 \pm 0.02$} & \textcolor{red}{$0.34 \pm 0.02$} & ${0.27 \pm 0.01}$ \\
S.F1 & ${0.24}$ & ${0.02}$ & ${0.24}$ & ${0.24}$ & ${0.0}$ & ${0.29 \pm 0.01}$ & ${0.07 \pm 0.02}$ & \textcolor{blue}{$0.31 \pm 0.02$} & \textcolor{red}{$0.33 \pm 0.02$} & ${0.25}$ \\
\hline
$G_{\cD}$ & \multicolumn{10}{c}{-3.41}\\
\bottomrule

\end{tabular}
\caption{\textbf{Comparison across policies on subset ``rate $[-\infty, -0.4]$ high variance" set using the NEWS2 reward.} ``rate $[-\infty, -0.4]$" refers the subgroup where patients are quickly out of risk of death, i.e., a large negative change rate of NEWS2 score. ``high variance" means that patient in this subgroup was observed with high fluctuation in terms of change rate of the risk of death.}
\label{table: NEWS2 reward-subgroup-test_rate_--0.4_high_std}
\end{table*}

%% file: main_sections/9suggestions.tex
\section{Consideration of RL for DTR}

Building on our main findings from the previous sections, here we outline several practical considerations that are crucial when conducting experiments in RL for DTR. 

\textbf{SL as a baseline:} Including SL as a baseline is crucial for evaluating whether RL can outperform a supervised approach using both OPE and supervised learning metrics. While comparing RL to SL is standard practice in the offline RL community, it may not be common knowledge to a broader audience. Therefore, we emphasize the importance of including SL as a baseline when assessing RL performance.

\textbf{Benchmarking against simplicity:}
Naive baselines provide a clear and straightforward benchmark that any advanced model should surpass to justify its complexity. According to our experiments, these simple strategies can surprisingly outperform complex RL models under certain conditions. This highlights the importance of including these simple strategies for sanity checks.

\textbf{Data stratification towards equitable DTR:}
Data stratification reveals the effectiveness of a policy on different patient subgroups and helps to identify treatment learning bias, ensuring equitable healthcare outcomes across patient populations. A comparative analysis of Figure \ref{fig:wins1} and Figure \ref{fig:win2} support this insight: On one hand, algorithms that excel in the overall test set may not necessarily maintain their superiority in stratified patient groups. On the other hand, RL has the potential to derive improved policies from sub-optimal offline data, and its effectiveness may be particularly pronounced in specific patient groups. Stratifying data by patient groups can be a valuable strategy to swiftly pinpoint where RL provides the most benefit. This targeted approach not only facilitates the identification of these advantages but also allows for detailed examination and validation by medical experts. Such a nuanced analysis could lead to more personalized and effective treatment strategies, demonstrating the true potential of RL in healthcare. Our study used episodic stratification based on the rate of change in NEWS2 scores. However, alternative stratification approaches, such as initial state stratification, could also be considered for future research.

\textbf{Alternative OPE methods}:
Recent studies have introduced more advanced methods for quantifying the dispersion or variance of OPE \cite{thomas2015high1, thomas2015high2,gottesman2020interpretable} and avoiding overfitting the importance-weighted return as an RL agent \cite{liu2022offline}. While these methods provide valuable insights into RL treatment decisions and are encouraged to be used in future RL-DTR works, they do not eliminate the impact of variance or reduce the variance of policy evaluation. This highlights the need to develop OPE methods specifically tailored to DTR. Another category of OPE methods, that were proposed for resolving some theoretical challenges such as convergence or stability, can also be adapted for use in DTR environments. This group includes algorithms such as least squared TD \cite{steven1996lstd}, emphatic TD \cite{sutton2016etd}, gradient TD \cite{maei2011gradient}, and accelerated TD \cite{pan2017atd,pan2017sketchatd}. These methods present opportunities for further investigation and development.

\textbf{Behavioral model selection and calibration:}
Previous work by \citeauthor{raghu2018behaviour} suggests that judicious selection and calibration of the behavioral model can help mitigate variance in OPE. However, our experiments indicated that the effectiveness of calibration may not be universal. We suggest exploring a spectrum of models and calibration techniques to identify the most suitable approach for the specific DTR setting.

\section{Discussions}

Our work critically examines the application of offline RL algorithms in DTRs, focusing on three key aspects: diversity in policy evaluation methods, variability in reward definitions, and the absence of informative baselines. We demonstrate that the comparative performance of RL algorithms can vary significantly depending on these three factors through extensive empirical analysis using the medical sepsis dataset. Additionally, we offer practical suggestions to guide future research in this field. Moving forward, it is essential to address these complexities with a more structured and standardized approach to fully realize the benefits of RL in DTRs.

\textbf{Future work and limitations.} Our study has several limitations that should be acknowledged and could be important future work directions. First, while we validated our position using the Sepsis dataset, future research could explore the generalizability of our findings across other relevant datasets. 
Second, our work focuses on linear function approximation and does not explore other representation learning methods. Investigating locality-encouraging representations could be beneficial \cite{engel2004krls,gomes2010stream,matt2017kernel,pan2021fuzzy}, as patients with highly similar conditions are likely to benefit from similar treatments. Future research might also consider exploring neural network-based representation learning methods, such as recurrent neural networks (RNN), or studying causal effects \cite{raghu2018model, zhang2020designing}.

Third, this paper does not consider that multiple optimal treatments might exist for a patient’s condition. Although one might expect this to have a minor effect on policy evaluation, as it is counterintuitive for vastly different doses to be simultaneously optimal for the same patient. However, it could still impact algorithm comparison if an algorithm consistently favors a certain type of mode. For instance, an SL algorithm might fit the mode presented in the training data, whereas an offline RL agent might learn a different mode. To address this issue, specialized algorithms that can capture multiple modes may be employed~\cite{Bishop1994MixtureDN,pan2020implicit} to model a policy. Finally, we performed episodic stratification based on the rate of change in NEWS2 scores, while there are other interesting stratification approaches, such as initial state stratification, to be explored. 



\section*{Acknowledgements}
Zhiyao Luo was supported by the Tang Oxford Scholarship from the China Oxford Scholar Fund (COSF) and the SBFT fellowship from the Sino-British Fellowship Trust (SBFT). Tingting Zhu was supported by the Royal Academy of Engineering under the Research Fellowship scheme.

\section*{Impact Statement}
Applying RL in high-stakes medical decision-making like Sepsis carries significant implications for patient care and treatment outcomes. While RL has the potential to revolutionize personalized medicine by learning from patient responses and adapting treatment plans accordingly, its deployment in clinical settings must be approached with utmost caution.  

This position paper critically examines the application of RL in DTRs. Particularly, our critique highlights significant challenges that must be addressed to prevent potential harm. The inconsistent and potentially inconclusive evaluation metrics present a clear risk: without robust assessment frameworks, the deployment of RL in medical decision-making could lead to suboptimal or even harmful treatment recommendations. Furthermore, our call for incorporating more baselines into the research of RL in DTRs underlines the necessity for benchmarking against simpler baselines. This is not only a matter of scientific rigor but also of ethical responsibility, ensuring that the adoption of more complex models is justified by demonstrable benefits to patient outcomes. 

It is the research community’s collective responsibility to ensure that these technologies are introduced in a manner that is safe, ethical, and ultimately beneficial to patient care. The impact of RL in healthcare could be profound, but it must be guided by a commitment not to harm, ensuring that the leap forward does not come at the expense of patient trust or safety.

%% file: appendix_sections/1data_preprocessing.tex
\section{Appendix Content}
\begin{itemize}
    \item Appendix \ref{sec:app-data description}: Data Description and Stratification.
    \item Appendix \ref{sec:app-ope_details}: Details of off-policy evaluation methods and our implementation.
    \item Appendix \ref{sec:app-baseline details}: Details of naive baselines.
    \item Appendix \ref{sec:app-training details}: Model Implementation and training details.
    \item Appendix \ref{sec:app-discussion}: Supplementary materials for Section \ref{sec: discussion-bc and value}.
    \item Appendix \ref{sec:app-model calibration}: Details of model calibration and the comparison between calibrated and uncalibrated models.
    \item Appendix \ref{sec:app-full result}: Full results, including the testing result on Outcome and SOFA reward settings (2 tables), and the respective performance on stratified patient groups (36 tables).
    \item Appendix \ref{sec:app-ratio plot}: Importance ratio histogram comparison between calibrated and uncalibrated behavior model on 3 reward settings across 5 naive baselines.
    
\end{itemize}

\section{Data Description}
\label{sec:app-data description}
\subsection{Patient Stratification}
Patient stratification is a pivotal process in healthcare\cite{chen2006risk, klein2015likelihood}, primarily due to the heterogeneity of patient responses and outcomes. This practice is critical for ensuring that medical interventions are tailored to the unique characteristics and needs of different patient groups. Despite its importance, many existing studies in the field of RL-DTR overlook this crucial step prior to model development.

In our work, we address this gap by incorporating a simple yet effective patient stratification into our RL framework. We use the NEWS2 score as a basis for stratifying patients, specifically focusing on their risk of mortality. This approach allows us to train and evaluate our RL models on a cohort that is balanced in terms of mortality risk and outcome, thereby reducing data selection bias. 

\begin{figure}
    \centering
    \includegraphics[width=1\linewidth]{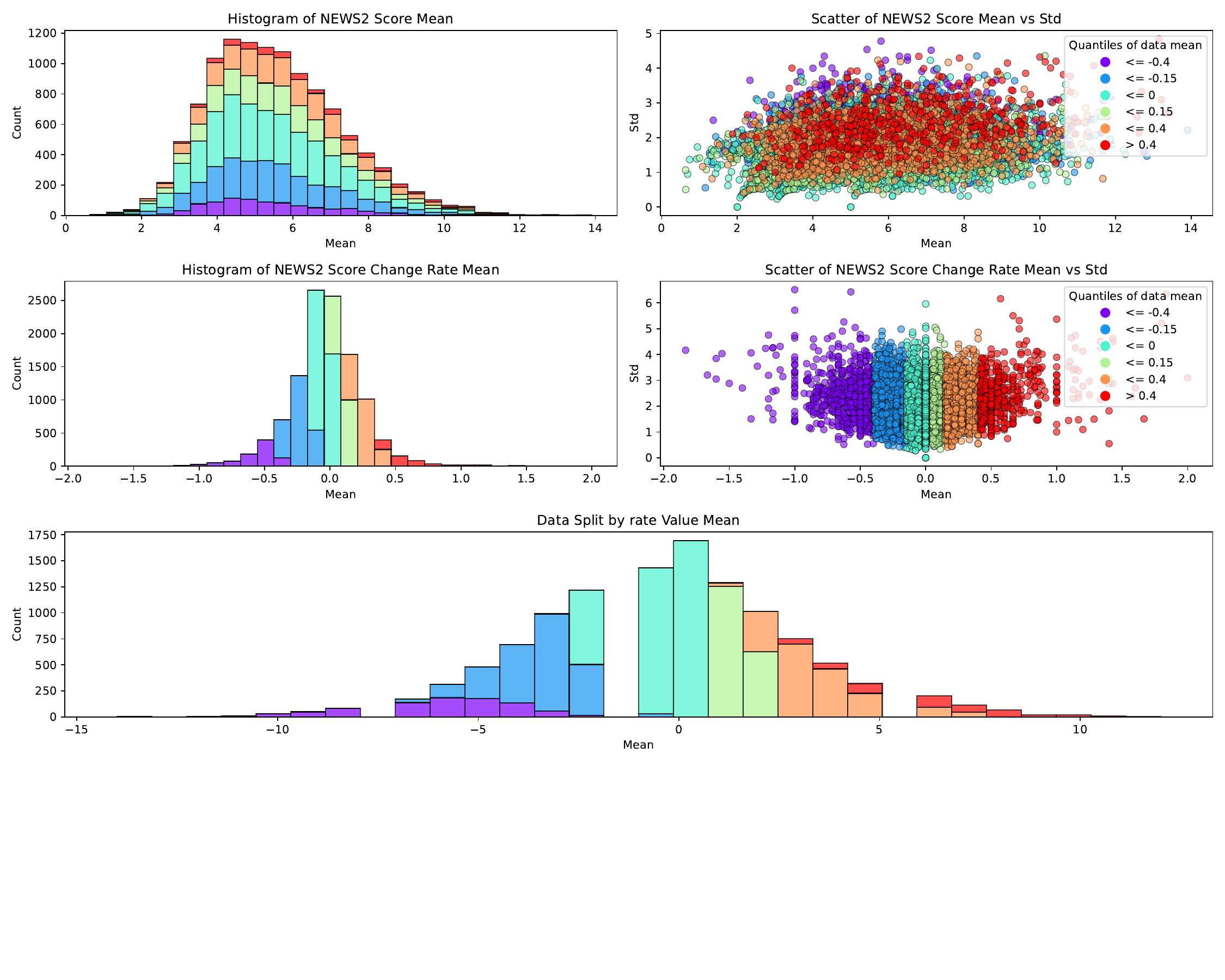}
    \caption{Data Stratification using change rate of NEWS2 score and admission outcome. The plot includes \textbf{ (top left)} the distribution of mean NEWS2 score,\textbf{ (top right)} mean NEWS2 against SD , \textbf{(middle left)} distribution of NEWS2 rate of change , \textbf{ (middle right)} NEWS2 rate of change mean against its SD , and \textbf{(bottom)} the distribution of mean NEWS2 score for each stratified fold. Stratified by the rate of change of NEWS2, it is observed that patient of different deterioration/recovery speed distributed across all risk levels. The plot also shows that the number of patients can significantly differ across risk stratification, supporting the need of RL evaluation on all subsets.}
    \label{fig:data stratification}
 
\end{figure}

We define our stratification process by considering the rate of change in NEWS2 scores during admission, which reflects the speed of patient deterioration or recovery. Additionally, patients are categorized into two groups based on the standard deviation (SD) of their NEWS2 score changes: those with high variance indicating a more fluctuated state and those with low variance indicating a more stable deterioration/recovery, split by the SD median in the subgroup. The NEWS2 scores are segmented into distinct brackets [-0.4, -0.15, 0, 0.15, 0.4], and the variance is classified as either low or high.  A comprehensive data distribution plot is shown in Fig\ref{fig:data stratification}.

For dataset split, we first divided all data into subgroups (2 outcomes, 6 NEWS2 score bins and 2 SD levels) and make sure each group is (near) evenly distributed in the training, validation and test set for fair evaluations.

\subsection{Action Discretization}
The 2 drugs of interest is binned into 5 classes each to formulate a discrete aciton space. We keep the binning method aligned with \citeauthor{komorowski2018artificial}: 

\begin{table}[ht]
    \centering
    \small
    \begin{tabular}{c|cc|cc}
    \toprule
    Action & \multicolumn{2}{c|}{IV Fluids} & \multicolumn{2}{c}{Vasopressor} \\
           & Range        & Median         & Range        & Median         \\
    \midrule
    0      & 0            & 0              & 0            & 0              \\
    1      & (0, 50]      &  40.0              & (0, 0.08]    &                0.044\\
    2      & (50, 180]    &    93.75            & (0.08, 0.22] &                 0.15\\
    3      & (180, 530]   &    315.35            & (0.22, 0.45] &                0.301\\
    4      & (530, $\infty$) &    949.8          & (0.45, $\infty$) &            0.9\\
    \bottomrule
    \end{tabular}
    \caption{Action Discretization Range and Median }
    \label{tab:action binning}
\end{table}

Please note that the median for each bin is different from the material provided in \url{https://static-content.springer.com/esm/art%3A10.1038%2Fs41591-018-0213-5/MediaObjects/41591_2018_213_MOESM1_ESM.pdf} since we employed a series of modifications (See section \ref{sec:3problem formulation}) to improve medical soundness.

%% file: appendix_sections/2OPE_details.tex
\section{Equations and Implementations of Off-policy Evaluation Methods}
\label{sec:app-ope_details}

\subsection{Weighted Importance Sampling with Bootstrapping}

The bootstrapped estimator in the context of RL involves resampling the dataset to generate multiple bootstrap samples. The value estimate for each sample is calculated, with the final bootstrapped estimate being the average of these individual estimates. Mathematically, the bootstrapped version of the Weighted Importance Sampling (WIS) estimator can be expressed as:

\begin{equation}
\hat{V}^{\pi}_{WIS_{b}} = \frac{1}{B} \sum_{b=1}^{B} \left( \frac{\sum_{i=1}^{N_b} \rho_{i}^{(b)} G_{i}^{(b)}}{\sum_{i=1}^{N_b} \rho_{i}^{(b)}} \right)
\end{equation}

Here, \( B \) represents the number of bootstrap samples, \( N_b \) is the number of trajectories in the \( b \)-th bootstrap sample, \( \rho_{i}^{(b)} \) denotes the importance sampling ratio for the \( i \)-th trajectory in the \( b \)-th sample, and \( G_{i}^{(b)} \) is the return for the \( i \)-th trajectory. This can be simplified as:

\begin{equation}
\hat{V}^{\pi}_{WIS_b} = \mathbb{E}_{b \sim B}\left[ \hat{V}^{\pi}_{WIS_{(b)}}\right] 
\end{equation}

For our implementation, we set $B=100$ and $N_b = N$ for all $b \in B$.

\subsection{Truncated Weighted Importance Sampling}

Truncated Weighted Importance Sampling (WIS) involves limiting the influence of trajectories with disproportionately high importance weights. This is achieved by truncating the importance weights at a specified threshold \( \tau \). The truncated WIS estimator is given by:

\begin{equation}
\hat{V}^{\pi}_{WIS_{tr}} = \frac{\sum_{i=1}^{N} \min(\rho_i, \tau) G_i}{\sum_{i=1}^{N} \min(\rho_i, \tau)}
\end{equation}

In our application, we select a truncation threshold of $\tau = 1$.

\subsection{Combined Weighted Importance Sampling with Bootstrapping and Truncation}

This approach integrates the robustness of bootstrapping with the stability of truncation, offering a more reliable estimate. The combined estimator is defined as:

\begin{equation}
\hat{V}^{\pi}_{WIS_{tr,B}} = \mathbb{E}_{b \sim B}\left[ \hat{V}^{\pi}_{{WIS_{tr}}_{(b)}}\right] 
\end{equation}

This revised section succinctly explains the implementation of different off-policy evaluation methods, ensuring the mathematical expressions are consistent with the defined symbols and terms, and providing clarity in the methodology.

To complete the paragraph with the appropriate equations and coherent explanations, here is the revised section:

\subsection{Implementation of Behavioral Cloning Policy}
\label{sec:app-implementation of bc}
We implement a supervised learning approach using a Long Short-Term Memory (LSTM) neural network to approximate the doctor's policy, referred to as the behavioral policy. The neural network is trained on a combined set of training and test data, and model selection is performed based on performance on a separate validation set. The training objective is to minimize the cross-entropy loss between the model's predicted actions and the actions taken by the doctor over time. The loss function can be formulated as:

\begin{equation}
L_{\text{CE}}(\hat{\pi}_{\sD}, \pi_{\sD}) = -\EE_{a\sim\cD} \log \hat{\pi}_{\sD}(a | s; \theta) \cdot a
\end{equation}

Here, \( \theta \) represents the parameters of the LSTM network and \( y_t^i \) is the actual action taken by the doctor at time \( t \) for patient $i$. The model with the highest patient-wise F1 score on the validation set is selected as the best model. We prefer the F1 score over the Area Under the Receiver Operating Characteristic (AUROC) due to the latter's potential to be misleading in imbalanced datasets.

\subsection{Doubly Robust Estimator}
\label{sec:app-DR}
The doubly robust estimator combines elements of both the direct method and importance sampling. For the direct method component, we use a training dataset comprising both the training and test sets, with the validation set used for model selection. The estimator for the direct method is updated using an offline 'SARSA' approach, which can be represented as:

\begin{equation}
L_{\text{DM}} = L_{\text{MSE}}\left( r + \gamma Q(s', a'), Q(s, a) \right)
\end{equation}

 The best model is chosen based on the minimum Temporal Difference (TD) error observed on the validation set. The behavioral policy utilized is consistent with the one described in the previous section.

%% file: appendix_sections/3baseline_details.tex
\section{Baseline Details}
\label{sec:app-baseline details}
\subsection{Naive Baselines}
\label{sec:app-naive_baselines}
The naive baselines apply action based on simple rules without considering the difference of states. To avoid zero in the importance ratio, we set a small value to zeros when $\pi(a_i)=0$. For convenience, we denote this small value as $\epsilon_1$. The probability of the rest action will receive a reduction of $\epsilon_2$  to guarantee that $\sum_{a}\pi(a)=1$. Assume there are $M$ actions in the action space. We denote the first action $a_0$ as zero drug, and the last action $a_{M-1}$as max drug. The equations for each baseline are given in Table \ref{table:naive baseline equations}.
\begin{table}[h]
\centering
\caption{Equations for Naive Baselines}
\label{table:naive baseline equations}
\begin{tabularx}{\textwidth}{|c|X|}
\hline
\textbf{Policy} & \textbf{Equation} \\
\hline
Alt Policy &
\(
\pi_{alt}(a) = \begin{cases} 
0.5 - \epsilon_2  & \text{if } a = a_0 \\
0.5 - \epsilon_2 & \text{if } a = a_{M-1} \\
\epsilon_1 & \text{otherwise}
\end{cases}
\) \\
\hline
Max Policy &
\(
\pi_{max}(a) = \begin{cases} 
1-\epsilon_2 & \text{if } a = a_{M-1} \\
\epsilon_1 & \text{otherwise}
\end{cases}
\) \\
\hline
Min Policy &
\(
\pi_{min}(a) = \begin{cases} 
1 - \epsilon_2 & \text{if } a = a_0 \\
\epsilon_1 & \text{otherwise}
\end{cases}
\) \\
\hline
Random Policy &
\(
\pi_r(a) = \frac{1}{M}
\) \\
\hline
Weight Policy &
\(
\pi_w(a = a_j) = p_j, \quad \text{for } j = 1, 2, ..., M
\) \\
\hline
\end{tabularx}
\end{table}

where $p$ is the occurrence probability vector of all 25 actions present in the training set.

\begin{align}
    p=[&0.00144, 0.07288, 0.12563, 0.13930, 0.10696, \nonumber \\
       &0.00178, 0.03407, 0.04142, 0.06409, 0.08265, \nonumber \\
       &0.00173, 0.02256, 0.02194, 0.04831, 0.04838, \nonumber \\
       &0.00184, 0.03228, 0.02171, 0.02993, 0.02659, \nonumber \\
       &0.00221, 0.02637, 0.01584, 0.01834, 0.01162]
\end{align}

%% file: appendix_sections/4training_details.tex
\section{Training and Hyperparameter Search}
\label{sec:app-training details}
\subsection{Network Structure}
\label{sec:app-network structure}
We employ a simplified model structure consisting of a single linear layer to minimize the influence of network architecture on the convergence of our RL algorithms.  This approach also eliminates the influence of training tricks for deep learning. The input to this model is a flattened 3-frame observation window, which effectively utilizes patient data from the past 12 hours of admission, ensuring that the model has access to a relevant and recent history of patient states.

\subsection{Hyperparameter Search}
\label{sec:app-hyperparameter search}
We conducted a comprehensive search for optimal hyperparameters, common to all RL algorithms used in this study. The key hyperparameters and their respective search ranges are presented in Table \ref{tab:hyperparameters}. 

\begin{table}[ht]
\centering
\begin{tabular}{|l|c|c|}
\hline
 \textbf{Owner} &\textbf{Hyperparameter} & \textbf{Values} \\ \hline
 \multirow{5}{*}{All} &Seed & [6311, 6890, 663, 4242, 8376] \\  &learning rate& [0.01, 0.001, 0.0001, 0.00001] \\  &batch size& [256] \\  &n step& [1] \\  &$\gamma$& [0.99] \\ \hline
 \multirow{2}{*}{BCQ} &unlikely action threshold & [0.3, 0.5] \\ 
 &imitation logits penalty& [0.02, 0.1, 0.5] \\ \hline
 \multirow{4}{*}{IQL} &actor update frequency & [1, 5] \\  &quantile& [0.7, 0.9] \\  &$\beta$& [0.7, 1.0] \\  &$\tau$& [0.001] \\ \hline
 CQL &$\alpha$ & [0.1, 0.5, 1.0] \\ \hline
\end{tabular}
\caption{\textbf{Summary of Hyperparameters and Their Values.} The owner column shows where the hyperparameter was used. $\beta$ is the temperature parameter for policy loss calculation, $\tau$ is the coefficient for soft update of target networks, and $\alpha$ is the weight for the conservative loss.}
\label{tab:hyperparameters}
\end{table}

The seeds for the random number generator were chosen to avoid any intentional bias and ensure reproducibility. They were generated using a standard random number generation process in Python.

%% file: appendix_sections/5discussion.tex
\section{Supplementary of 'Understanding the Variance of Policy Evaluation'}
\label{sec:app-discussion}

\subsection{Behavioral Model and Value Estimation Error}
\begin{figure}[h]
\begin{subfigure}[b]{0.245\textwidth}
    \centering
    \includegraphics[width=\linewidth]{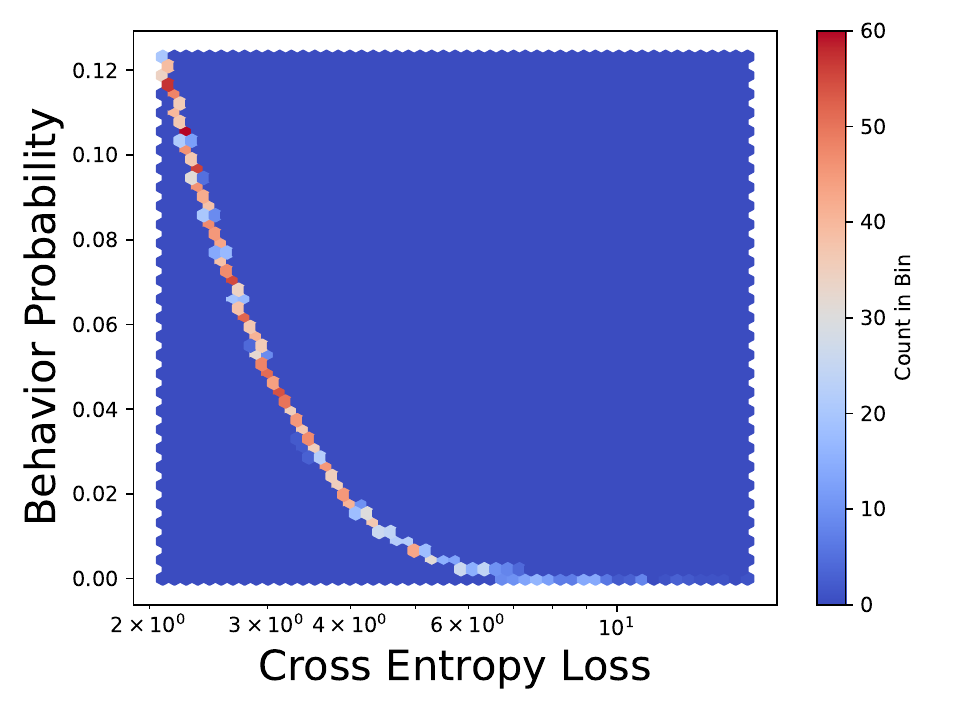}
    \caption{$\hat{\pi}_{\cD}$, train}
\end{subfigure}
\hfill
\begin{subfigure}[b]{0.245\textwidth}
    \centering
    \includegraphics[width=\linewidth]{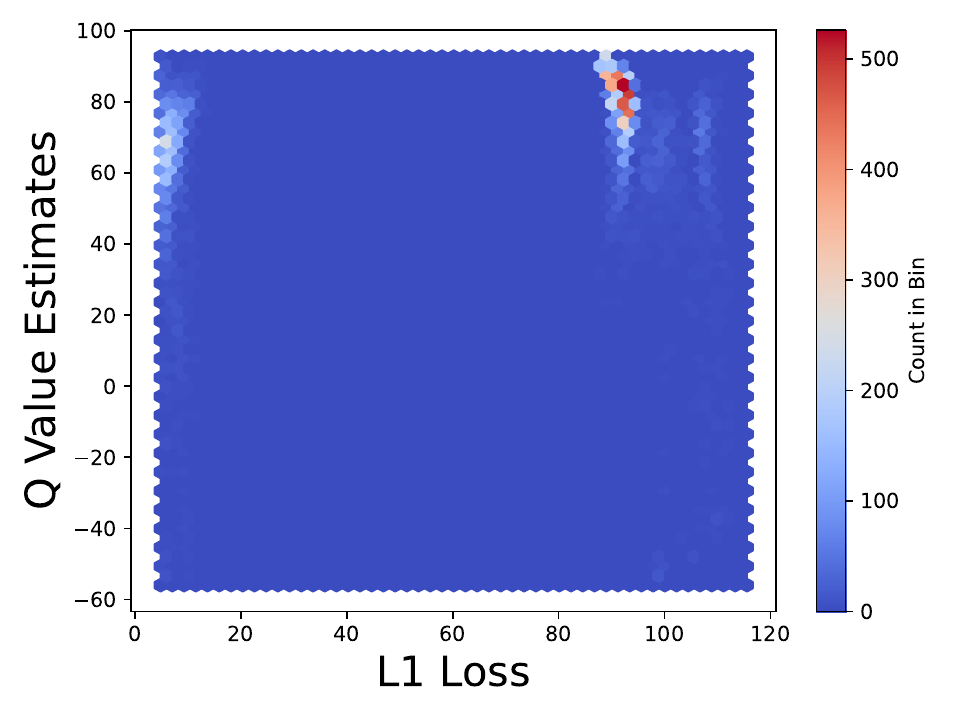}
    \caption{Training set, Outcome reward}
\end{subfigure}
\hfill
\begin{subfigure}[b]{0.245\textwidth}
    \centering
    \includegraphics[width=\linewidth]{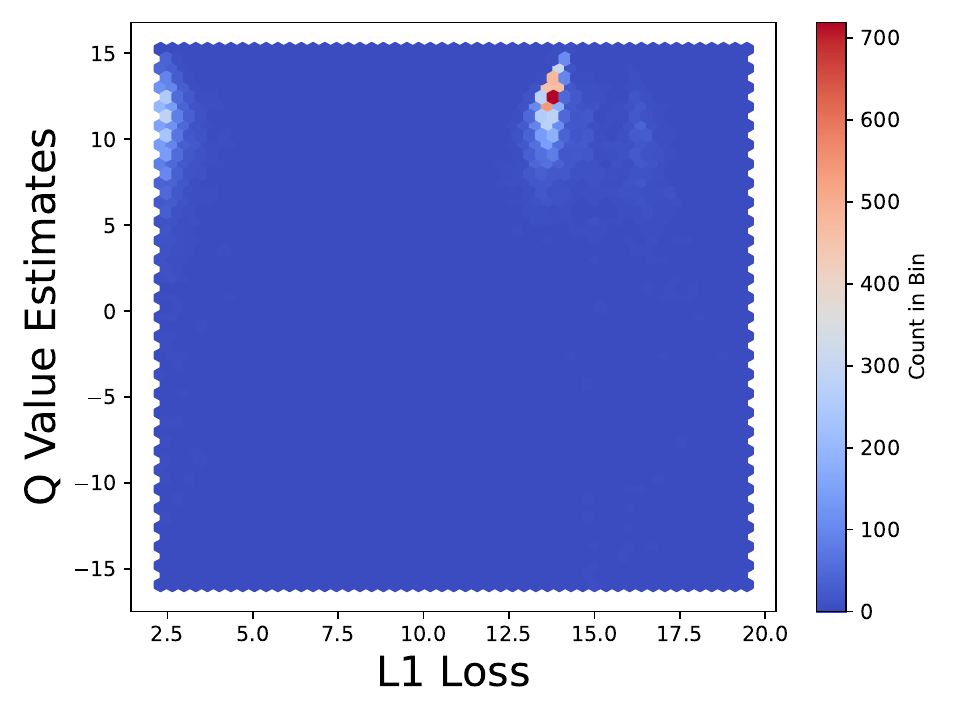}
    \caption{$\hat{Q}^{\pi}$, train, SOFA reward}
\end{subfigure}
\hfill
\begin{subfigure}[b]{0.245\textwidth}
    \centering
    \includegraphics[width=\linewidth]{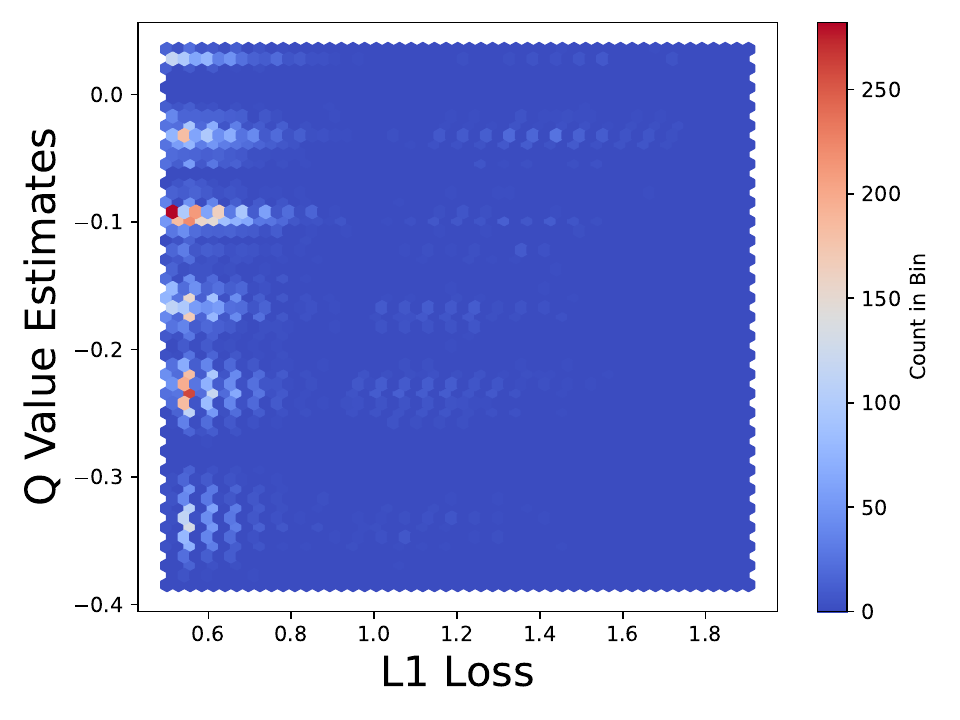}
    \caption{$\hat{Q}^{\pi}$, train, NEWS reward}
\end{subfigure}

\begin{subfigure}[b]{0.245\textwidth}
    \centering
    \includegraphics[width=\linewidth]{plot_bc_value/all_val_bc_loss_vs_value.pdf}
    \caption{$\hat{\pi}_{\cD}$, val}
\end{subfigure}
\hfill
\begin{subfigure}[b]{0.245\textwidth}
    \centering
    \includegraphics[width=\linewidth]{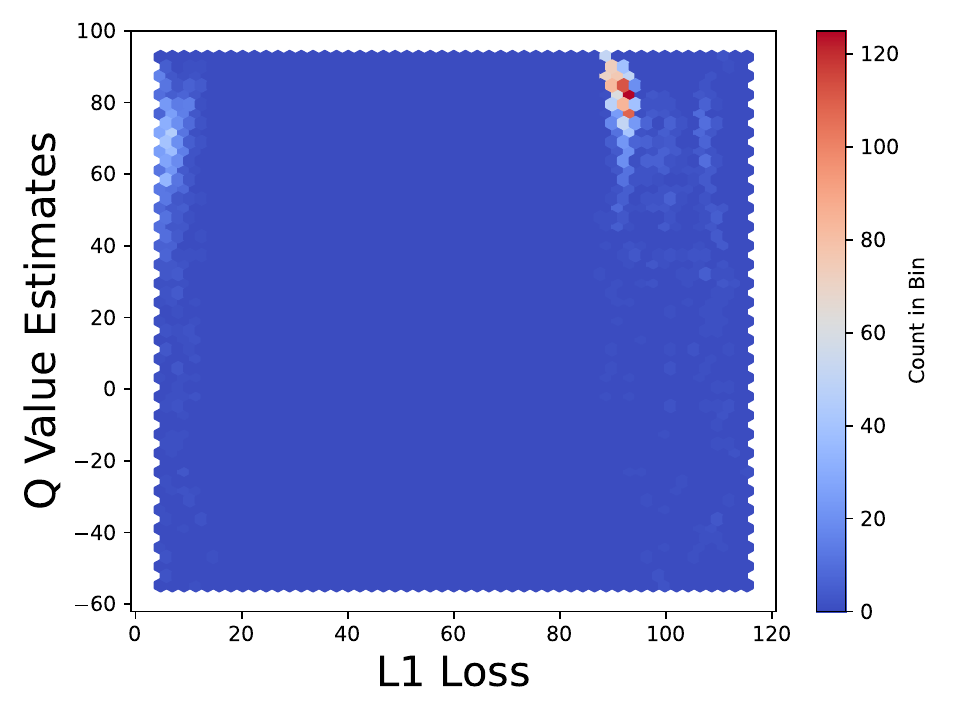}
    \caption{$\hat{Q}^{\pi}$, val, Outcome reward}
\end{subfigure}
\hfill
\begin{subfigure}[b]{0.245\textwidth}
    
    \centering
    \includegraphics[width=\linewidth]{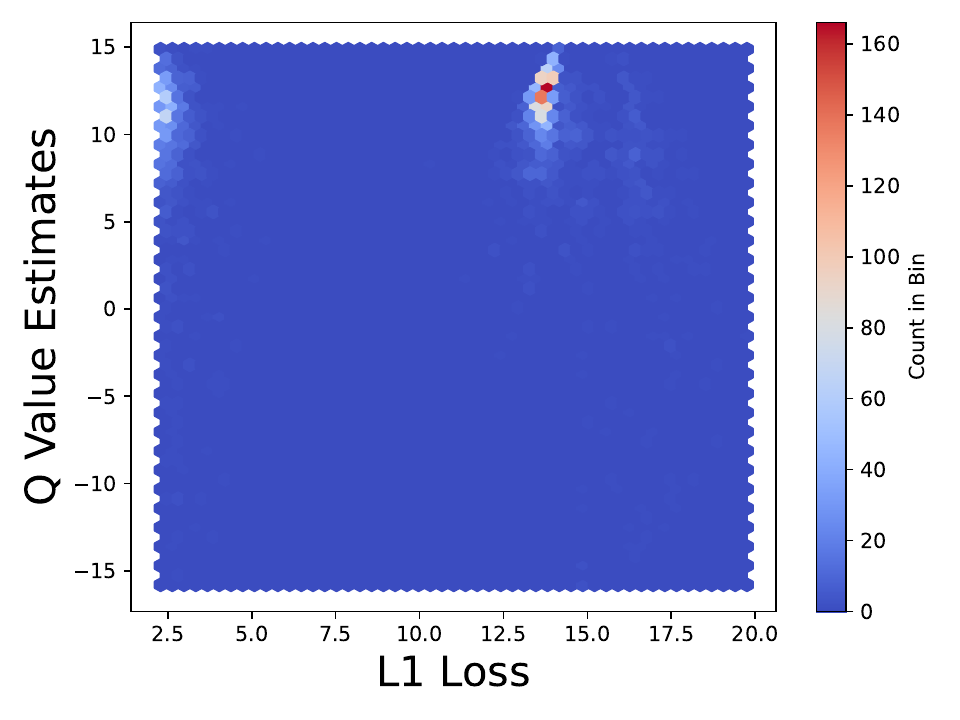}
    \caption{$\hat{Q}^{\pi}$, val, SOFA reward}
\end{subfigure}
\hfill
\begin{subfigure}[b]{0.245\textwidth}
    \centering
    \includegraphics[width=\linewidth]{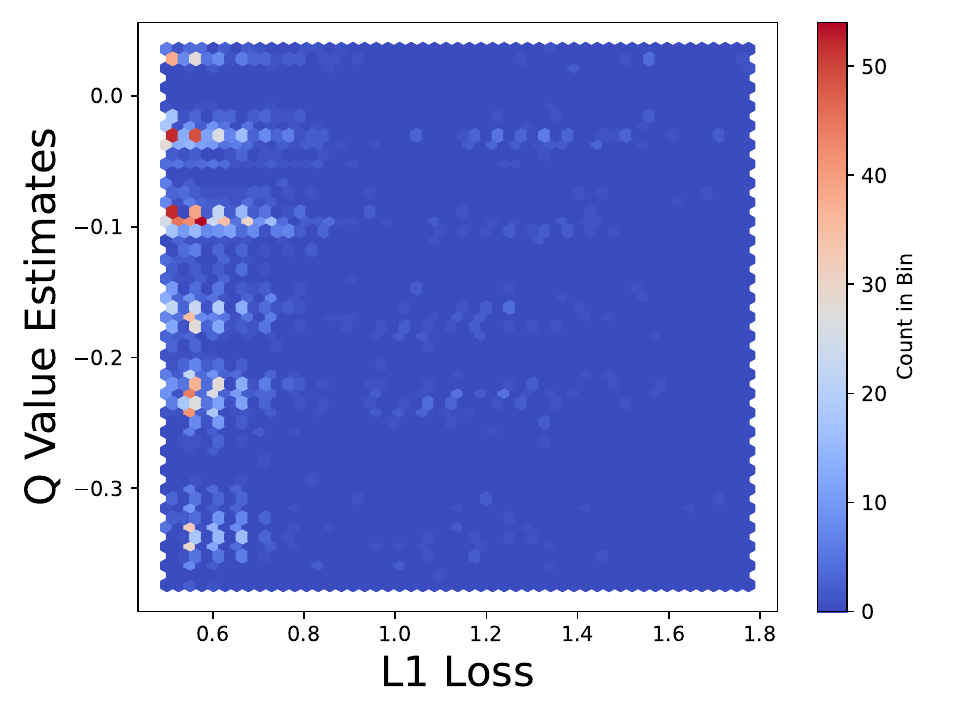}
    \caption{$\hat{Q}^{\pi}$, val, NEWS2 reward}
\end{subfigure}
\caption{\textbf{a)}, \textbf{e)} Cross entropy losses against inference probabilities in the training and validation set for $\hat{\pi}_{\cD}$. \textbf{b)}, \textbf{c)}, \textbf{d)}, \textbf{f)}, \textbf{g)}, \textbf{h)} L1 losses against Q estimates in the training and validation set on the 3 reward settings for $\hat{Q}^{\pi}$.}
\label{fig:app-bc_value_loss_vs_prob}
\end{figure}

Fig \ref{fig:app-bc_value_loss_vs_prob} shows the correlation between behavioral cloning loss w.r.t the inference probability for $\hat{\pi}_{\cD}$, and the correlation between value approximation loss w.r.t the inference state-action value for $\hat{Q}^{\pi}$. It is supplementary to Section \ref{sec: discussion-bc and value}, showing that the correlations exists not only in the testing set but in both the training and validation sets, independent of the reward settings. 

\begin{figure}[h]
\begin{subfigure}[b]{0.32\textwidth}
    \centering
    \includegraphics[width=\linewidth]{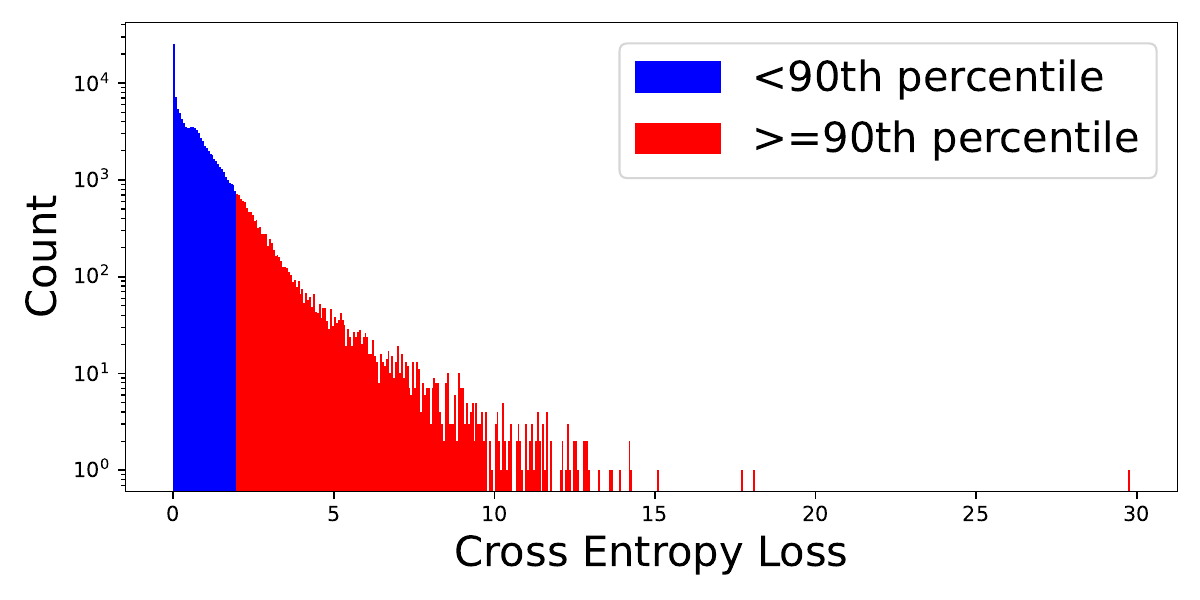}
    \caption{$\hat{\pi}_{\cD}$, train}
\end{subfigure}
\hfill
\begin{subfigure}[b]{0.32\textwidth}
    \centering
    \includegraphics[width=\linewidth]{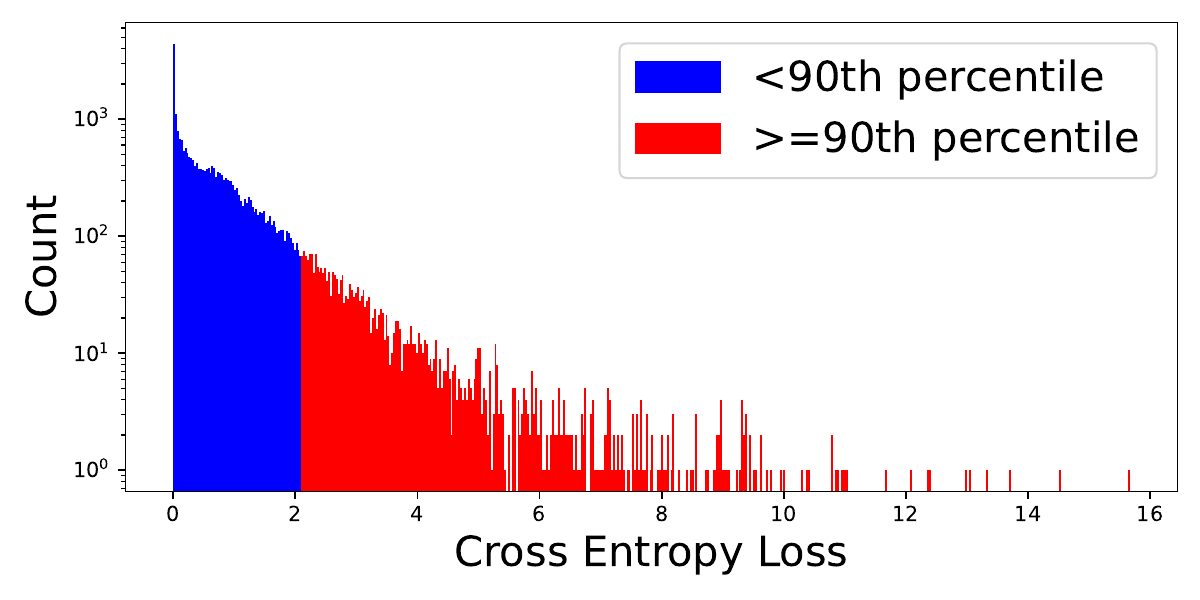}
    \caption{$\hat{\pi}_{\cD}$, val}
\end{subfigure}
\hfill
\begin{subfigure}[b]{0.32\textwidth}
    \centering
    \includegraphics[width=\linewidth]{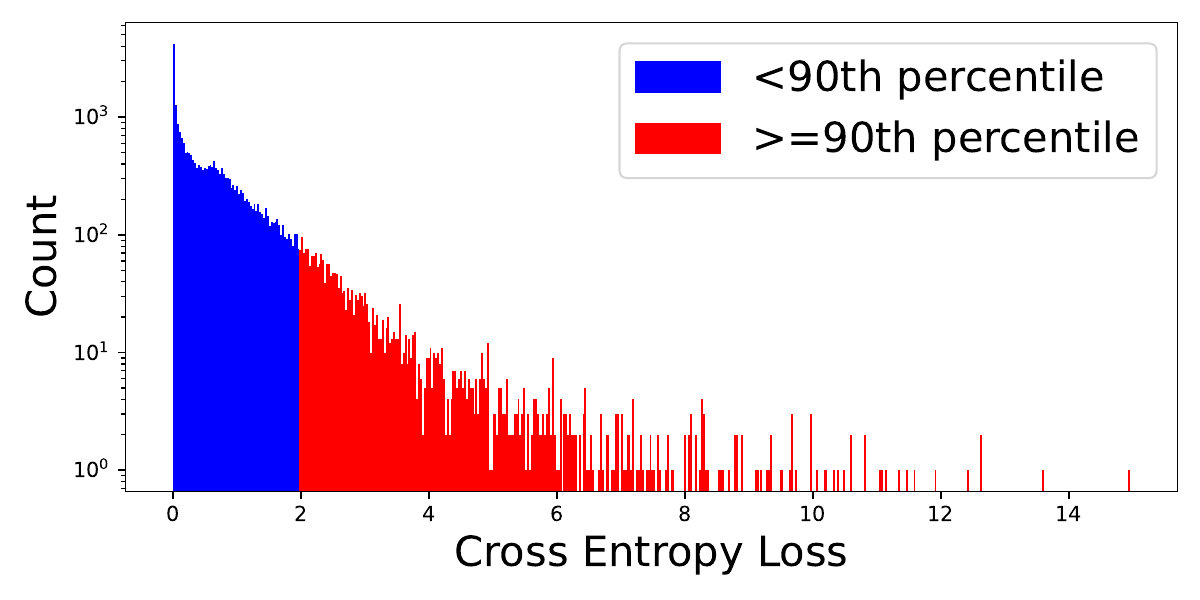}
    \caption{$\hat{\pi}_{\cD}$, test}
\end{subfigure}
\begin{subfigure}[b]{0.32\textwidth}
    \centering
    \includegraphics[width=\linewidth]{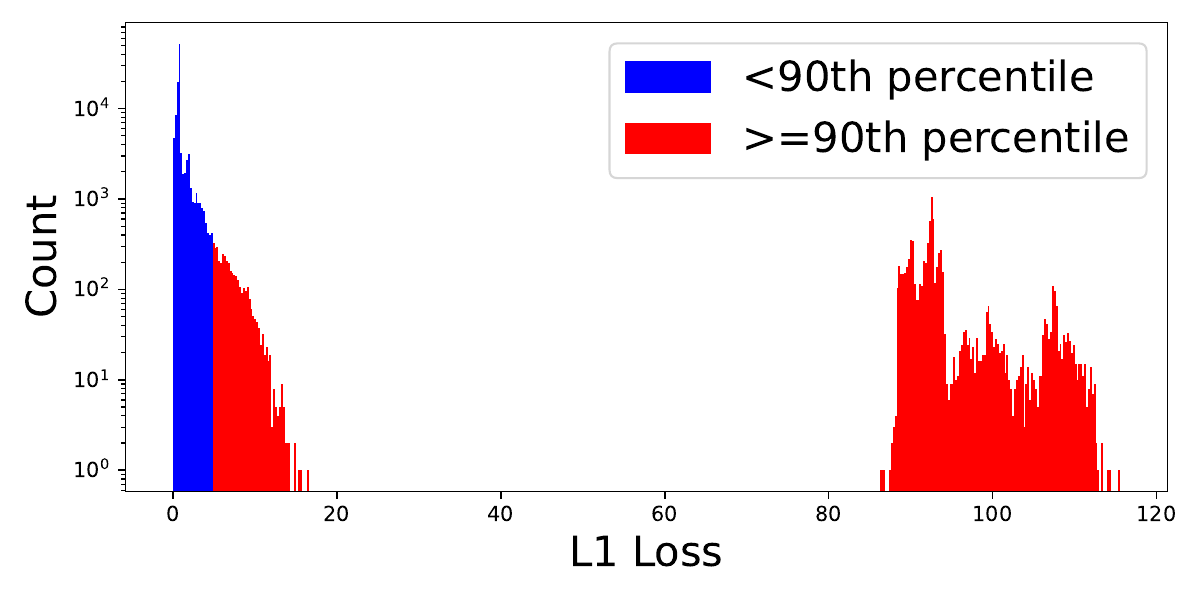}
    \caption{$\hat{Q}^{\pi}$, Outcome reward, train set}
\end{subfigure}
\hfill
\begin{subfigure}[b]{0.32\textwidth}
    \centering
    \includegraphics[width=\linewidth]{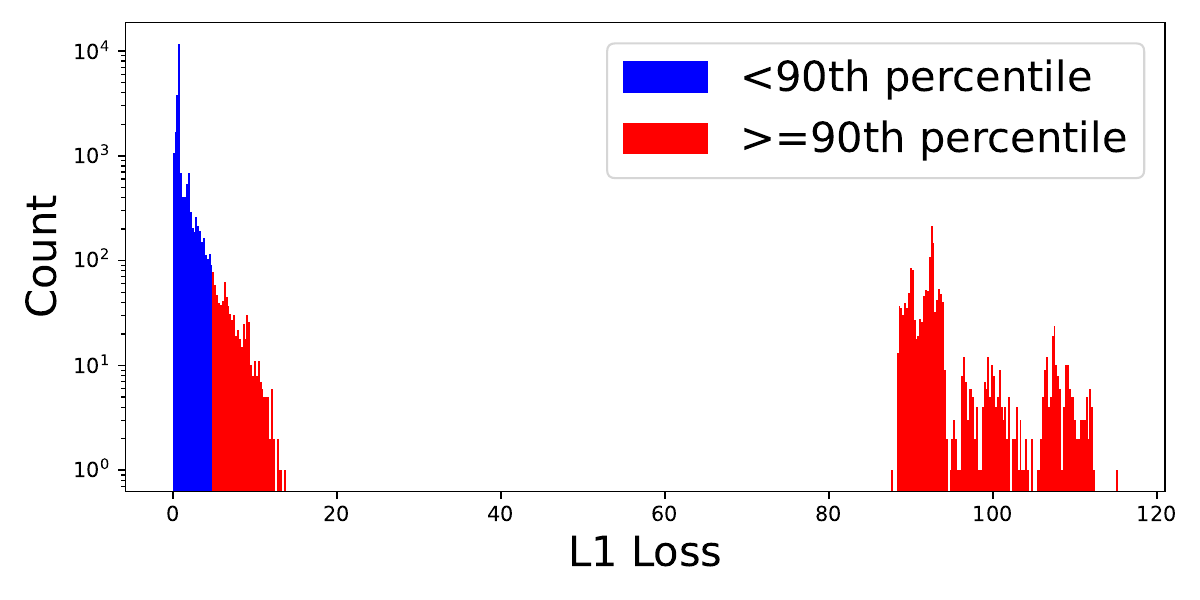}
    \caption{$\hat{Q}^{\pi}$, Outcome reward, val set}
\end{subfigure}
\hfill
\begin{subfigure}[b]{0.32\textwidth}
    \centering
    \includegraphics[width=\linewidth]{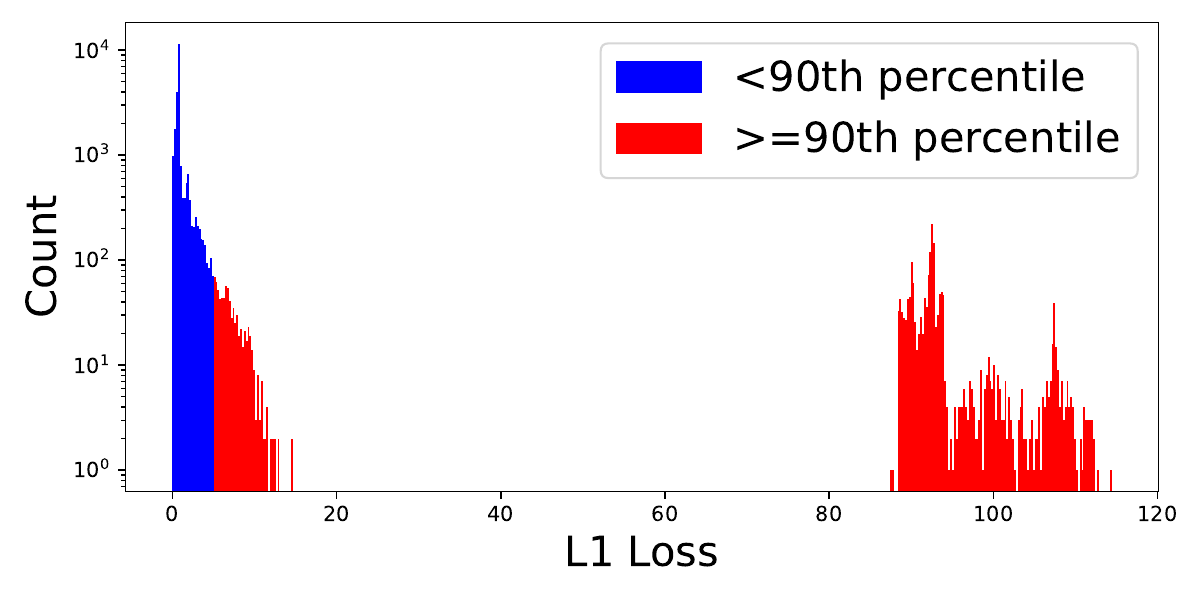}
    \caption{$\hat{Q}^{\pi}$, Outcome reward, test set}
\end{subfigure}
\begin{subfigure}[b]{0.32\textwidth}
    \centering
    \includegraphics[width=\linewidth]{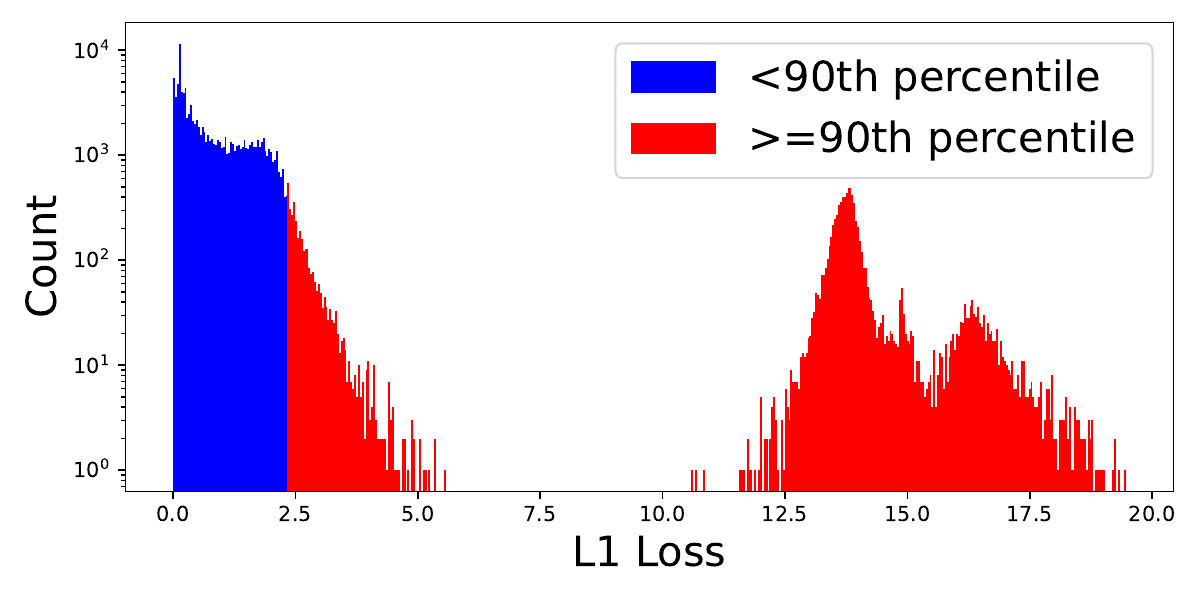}
    \caption{$\hat{Q}^{\pi}$, SOFA reward, train set}
\end{subfigure}
\hfill
\begin{subfigure}[b]{0.32\textwidth}
    \centering
    \includegraphics[width=\linewidth]{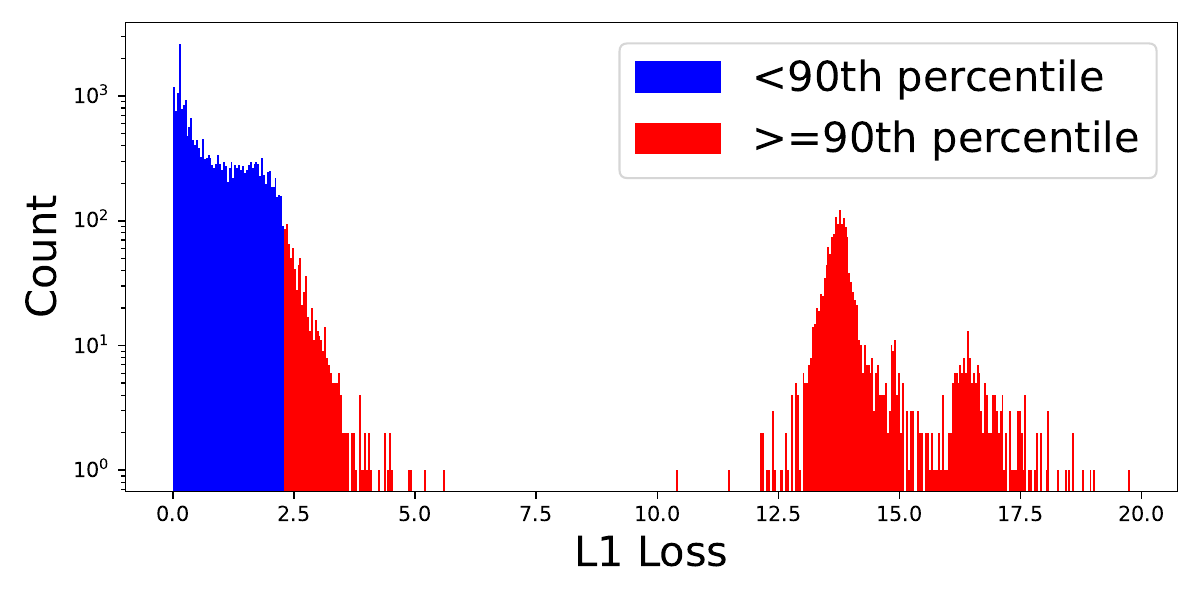}
    \caption{$\hat{Q}^{\pi}$, SOFA reward, val set}
\end{subfigure}
\hfill
\begin{subfigure}[b]{0.32\textwidth}
    \centering
    \includegraphics[width=\linewidth]{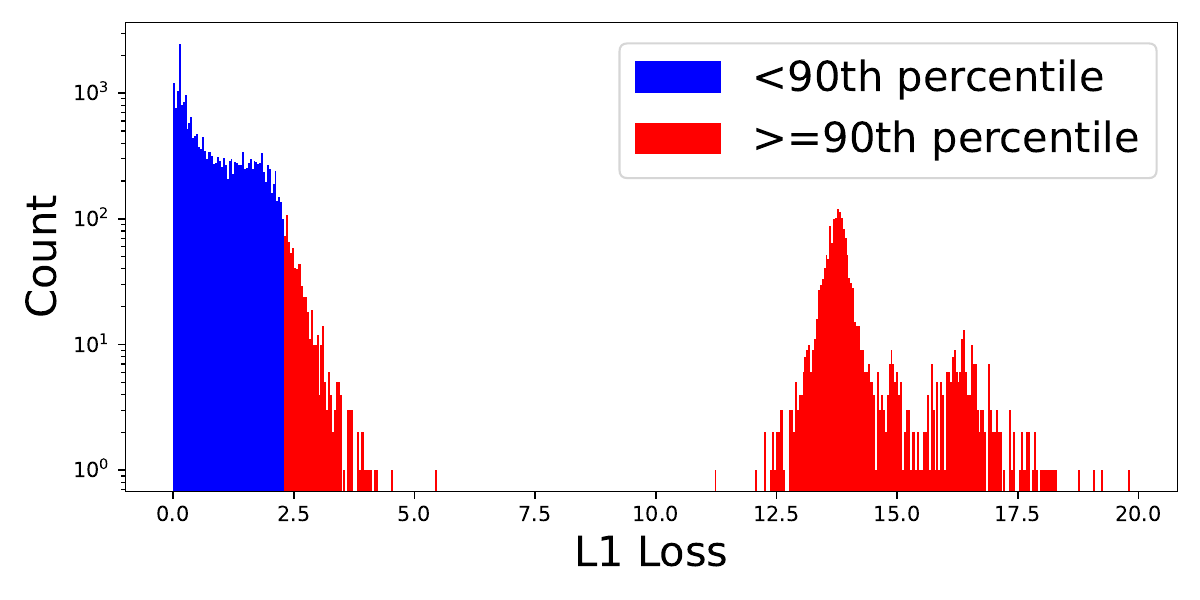}
    \caption{$\hat{Q}^{\pi}$, SOFA reward, test set}
\end{subfigure}
\begin{subfigure}[b]{0.32\textwidth}
    \centering
    \includegraphics[width=\linewidth]{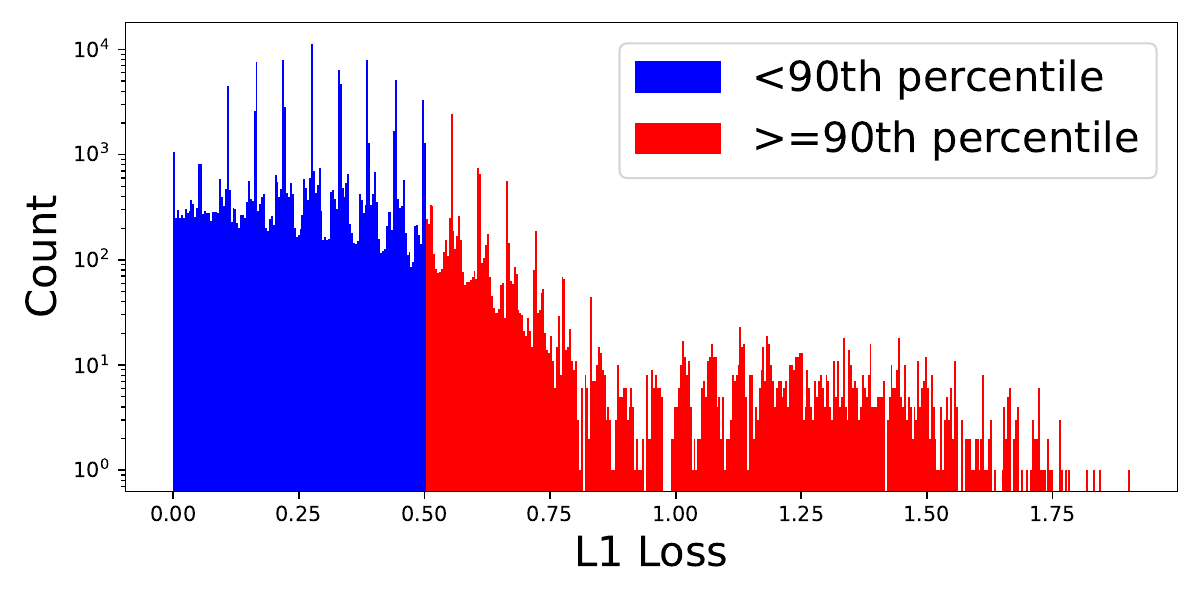}
    \caption{$\hat{Q}^{\pi}$, NEWS2 reward, train set}
\end{subfigure}
\hfill
\begin{subfigure}[b]{0.32\textwidth}
    \centering
    \includegraphics[width=\linewidth]{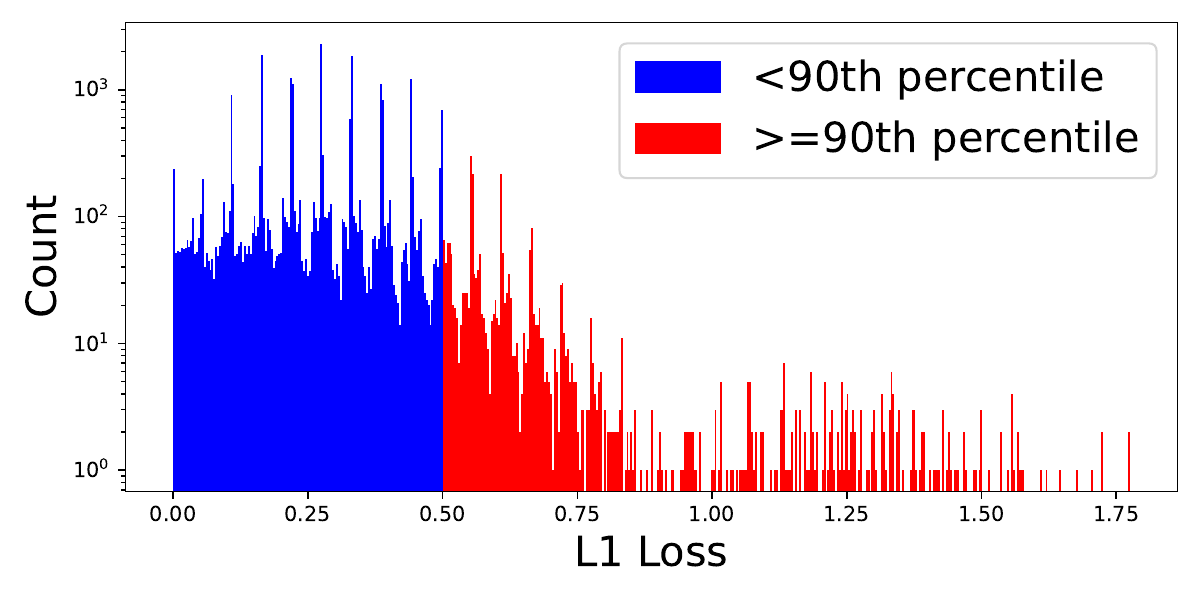}
    \caption{$\hat{Q}^{\pi}$, NEWS2 reward, val set}
\end{subfigure}
\hfill
\begin{subfigure}[b]{0.32\textwidth}
    \centering
    \includegraphics[width=\linewidth]{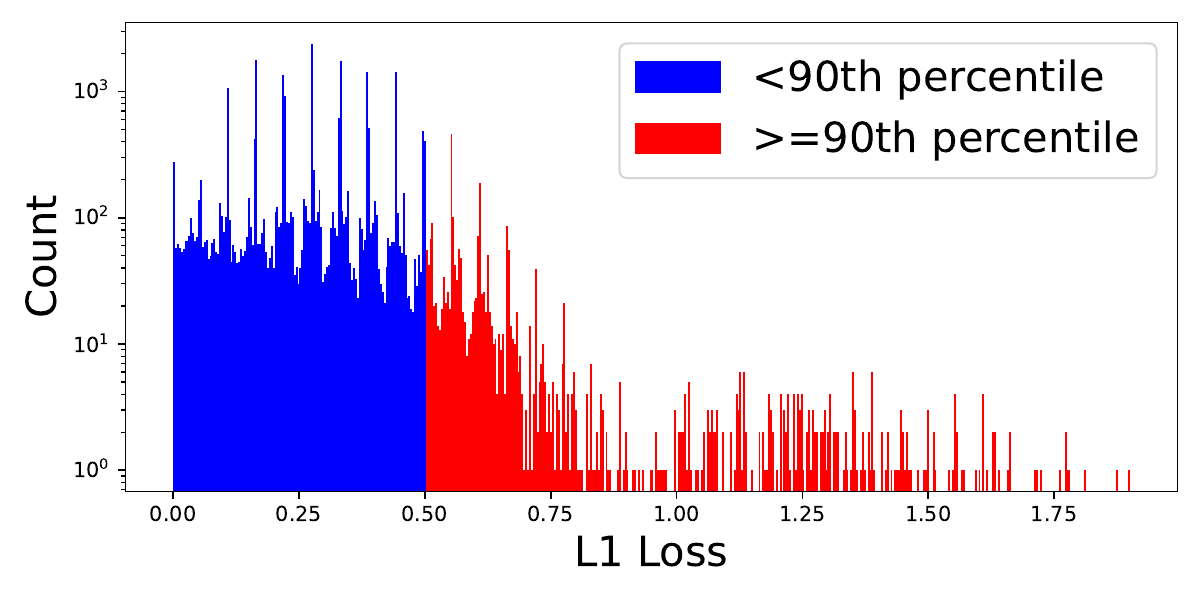}
    \caption{$\hat{Q}^{\pi}$, NEWS2 reward, test set}
\end{subfigure}
\caption{\textbf{a)}, \textbf{b)}, \textbf{c)} $\hat{\pi}_{\cD}$ cross-entropy loss histogram on training, validation and test set. \textbf{d)}, \textbf{e)}, \textbf{f)} $\hat{Q}^{\pi}$ L1 loss histogram on Outcome reward, \textbf{g)}, \textbf{h)}, \textbf{i)}, $\hat{Q}^{\pi}$ L1 loss histogram on SOFA reward,\textbf{j)}, \textbf{k)}, \textbf{l)} $\hat{Q}^{\pi}$ L1 loss histogram on NEWS2  reward.}
\label{fig:bc_app-value_loss_hist}
\end{figure}

We want to explain further that the error of $\hat{Q}^{\pi}$ and $\hat{\pi}_{\cD}$ is not a consequence of ill model training. Our selected $\hat{\pi}_{\cD}$ reaches an F1 score of 0.7 on both OPE training set and validation set\footnote{''OPE training set' here means a combined dataset of training and test set. 'Validation set' still follows its original meaning. We avoid using the original training, validation and test set as RL for decoupling}. Similarly, we select the $\hat{Q}^{\pi}$ that can best minimize the TD error. Your provided text describes loss distribution trends in a model's performance, contrasting behavior models and value functions. To enhance clarity and logical flow, the text can be revised for more precise language, structured presentation of the data, and improved coherence. Here is a refined version:

Figure \ref{fig:bc_app-value_loss_hist} provides a visualization of loss distributions across samples, highlighting the predictive accuracy of both the behavior model, denoted as $\hat{\pi}_{\cD}$, and the value functions, $\hat{Q}^{\pi}$. The behavior model generally predicts well across the majority of samples but exhibits notably higher losses in a minority of cases. This pattern follows a logarithmic trend on a linear exponential (y) scale, a characteristic often observed in imbalanced learning scenarios. For the value functions, the losses distribute differently. Specifically, the loss distribution for the Outcome and SOFA rewards form Gaussian-shaped peaks centered at 15 and 100, respectively. This distribution correlates with their termination rewards of $\pm15$ and $\pm100$. Such peaks suggest that the value functions struggle to accurately predict outcomes when there is a significant discrepancy between intermediate and termination rewards. In contrast, the NEWS2 score does not show this pattern, likely due to its more gradual reward function.

\begin{figure}[h]

\begin{subfigure}[b]{0.245\textwidth}
    \centering
    \includegraphics[width=\linewidth]{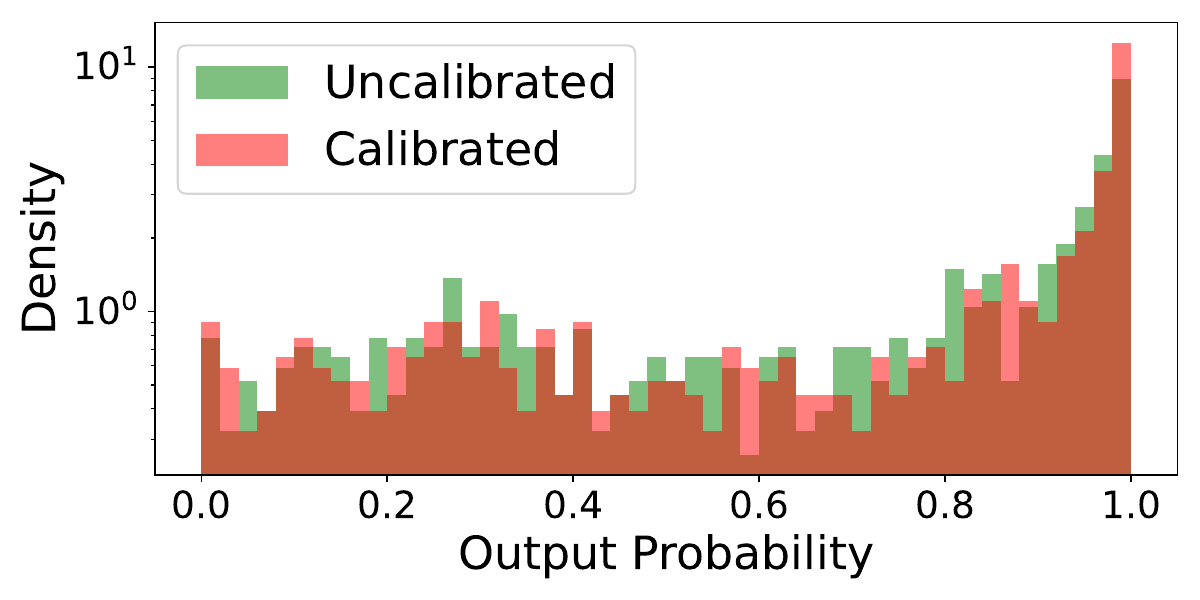}
    \caption{[$-\infty, -0.4$] low var}
\end{subfigure}
\hfill
\begin{subfigure}[b]{0.245\textwidth}
    \centering
    \includegraphics[width=\linewidth]{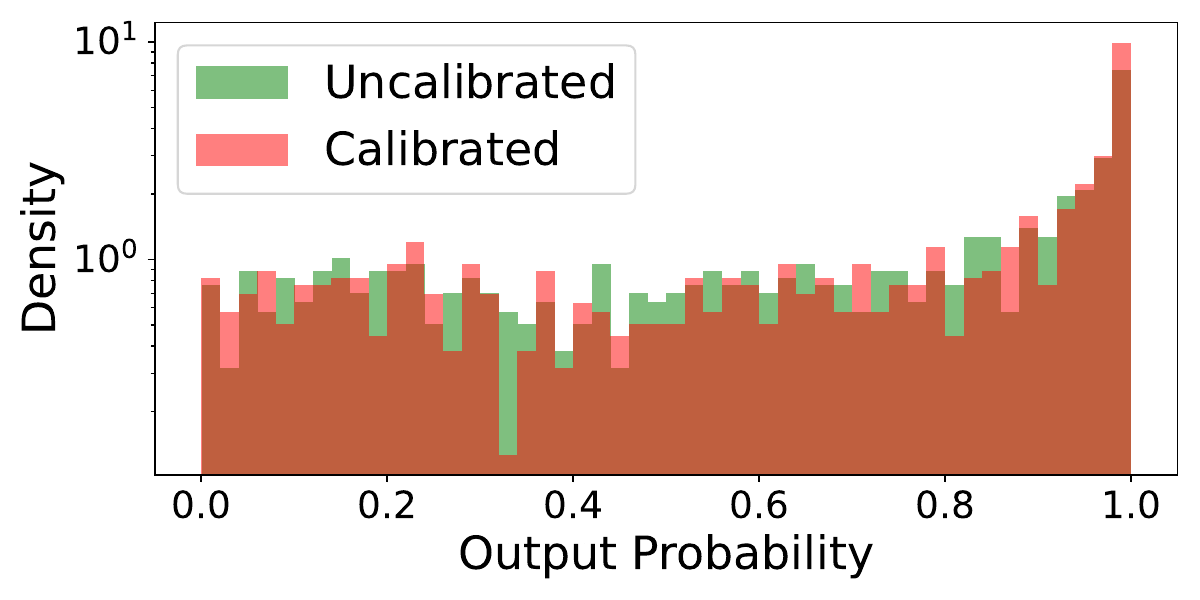}
    \caption{[$-0.4, -0.15$] high var}
\end{subfigure}
\hfill
\begin{subfigure}[b]{0.245\textwidth}
    \centering
    \includegraphics[width=\linewidth]{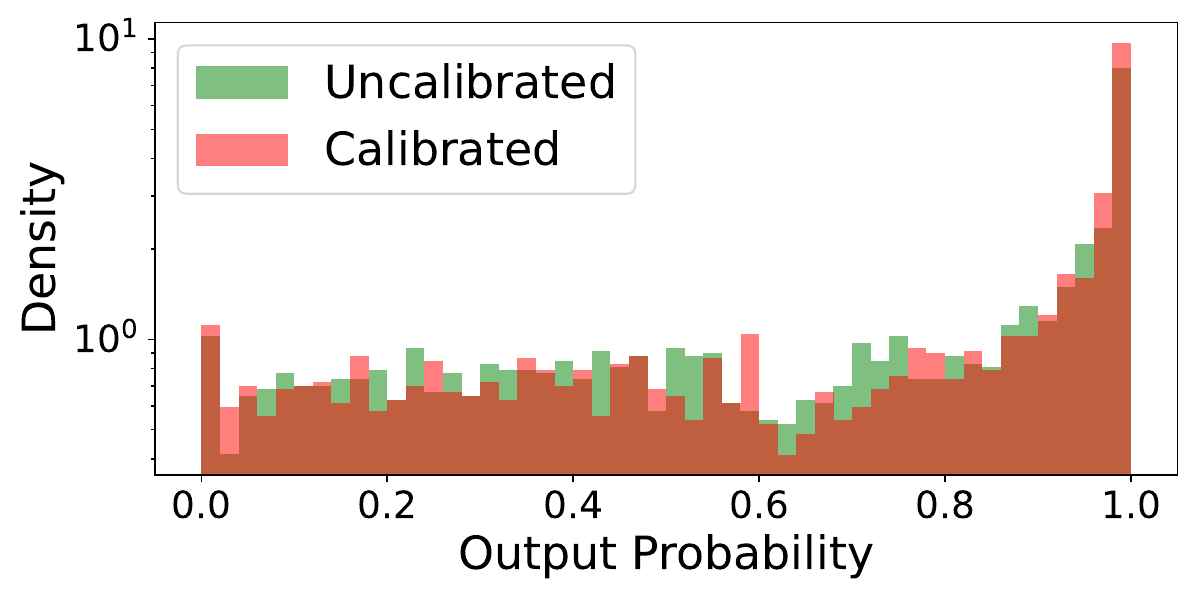}
    \caption{[$-0.4, -0.15$] low var}
\end{subfigure}
\hfill
\begin{subfigure}[b]{0.245\textwidth}
    \centering
    \includegraphics[width=\linewidth]{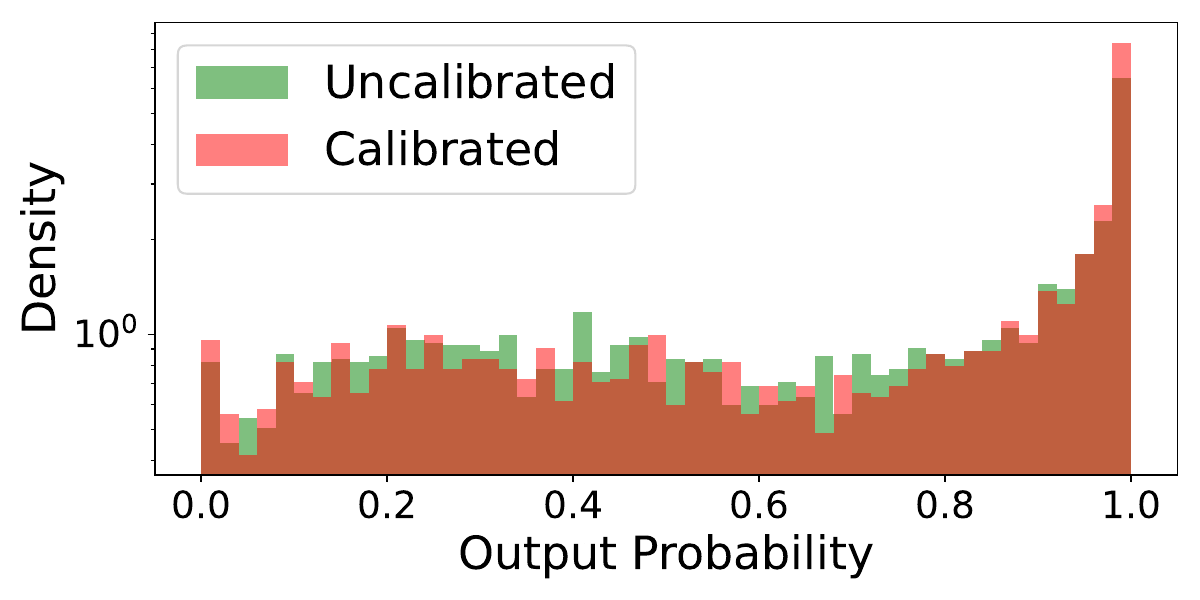}
    \caption{[$-0.4, -0.15$] high var}
\end{subfigure}
\begin{subfigure}[b]{0.245\textwidth}
    \centering
    \includegraphics[width=\linewidth]{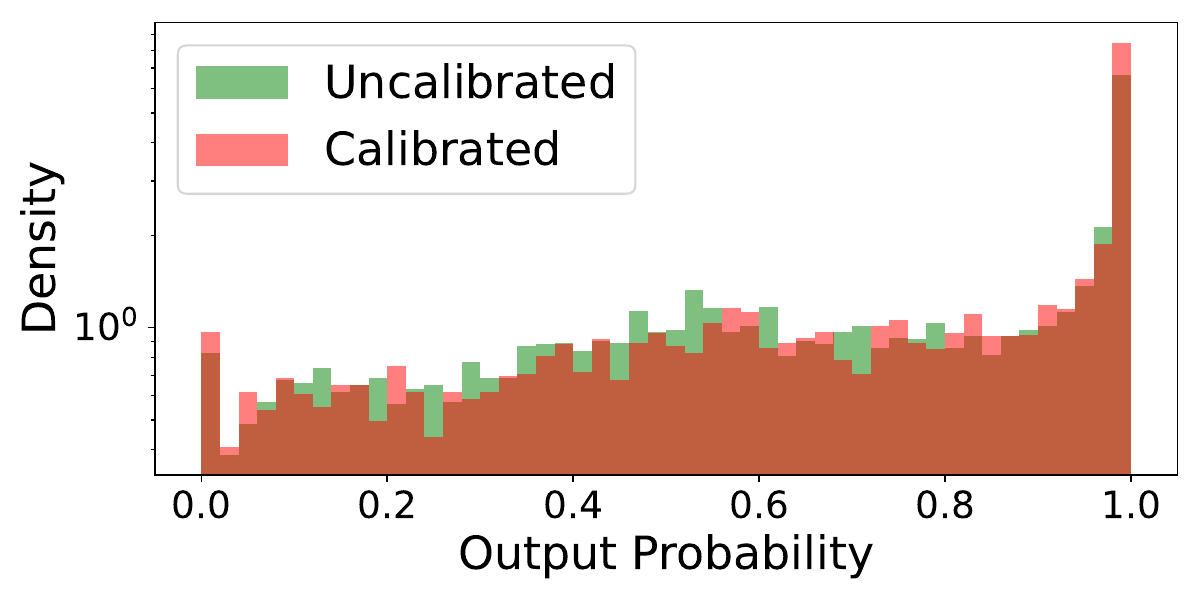}
    \caption{[$-0.15, 0$] low var}
\end{subfigure}
\hfill
\begin{subfigure}[b]{0.245\textwidth}
    \centering
    \includegraphics[width=\linewidth]{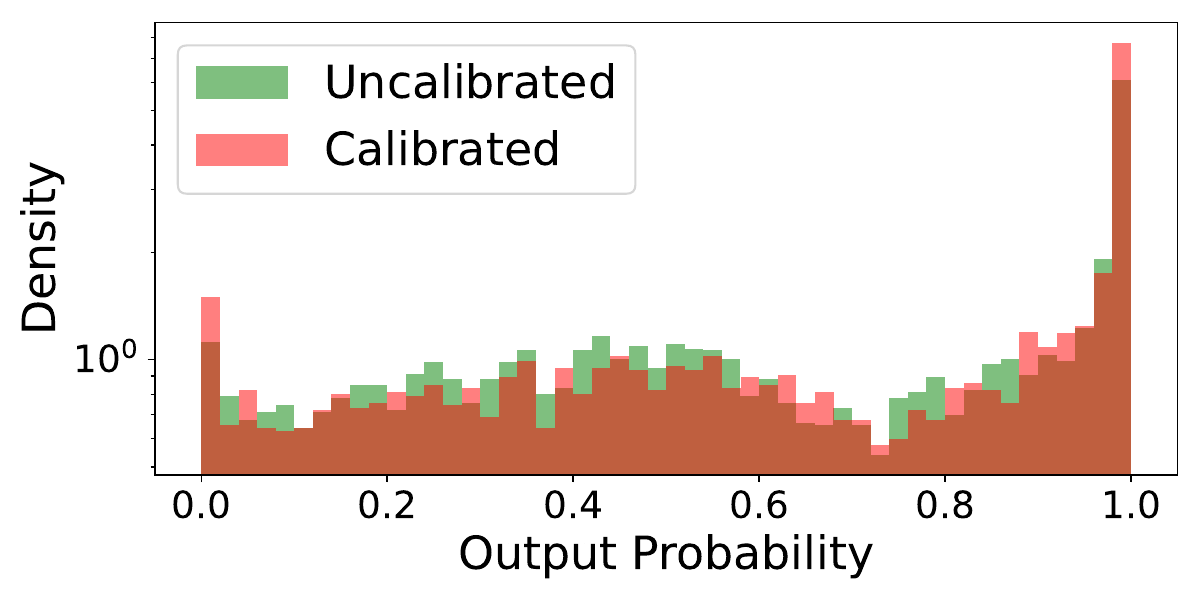}
    \caption{[$-0.15, 0$] high var}
\end{subfigure}
\hfill
\begin{subfigure}[b]{0.245\textwidth}
    \centering
    \includegraphics[width=\linewidth]{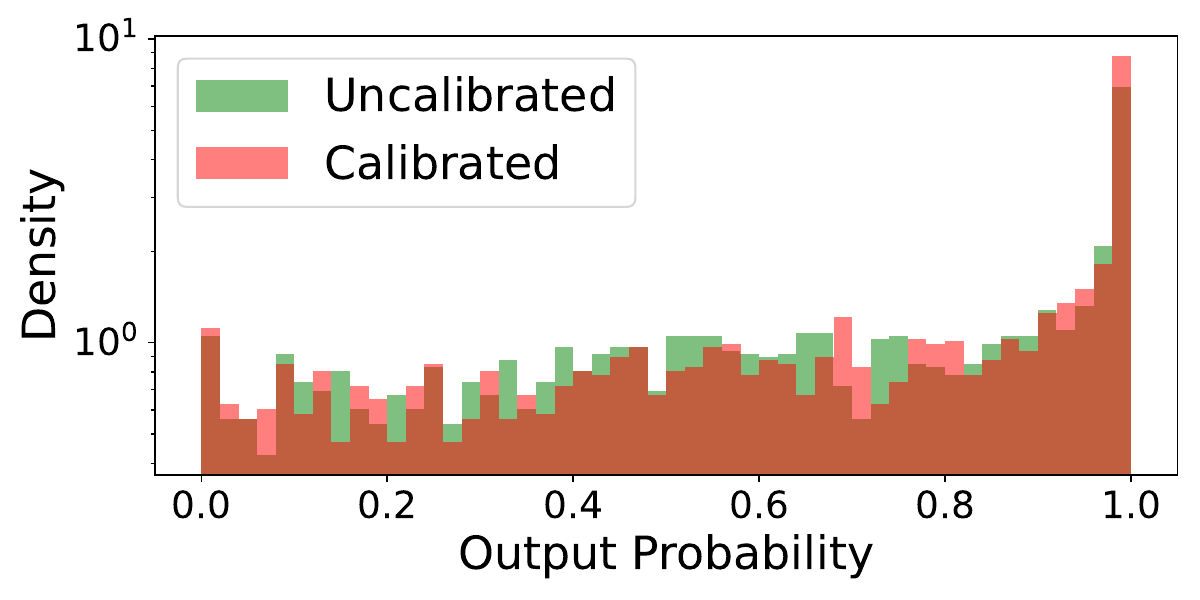}
    \caption{[$0, 0.15$] low var}
\end{subfigure}
\hfill
\begin{subfigure}[b]{0.245\textwidth}
    \centering
    \includegraphics[width=\linewidth]{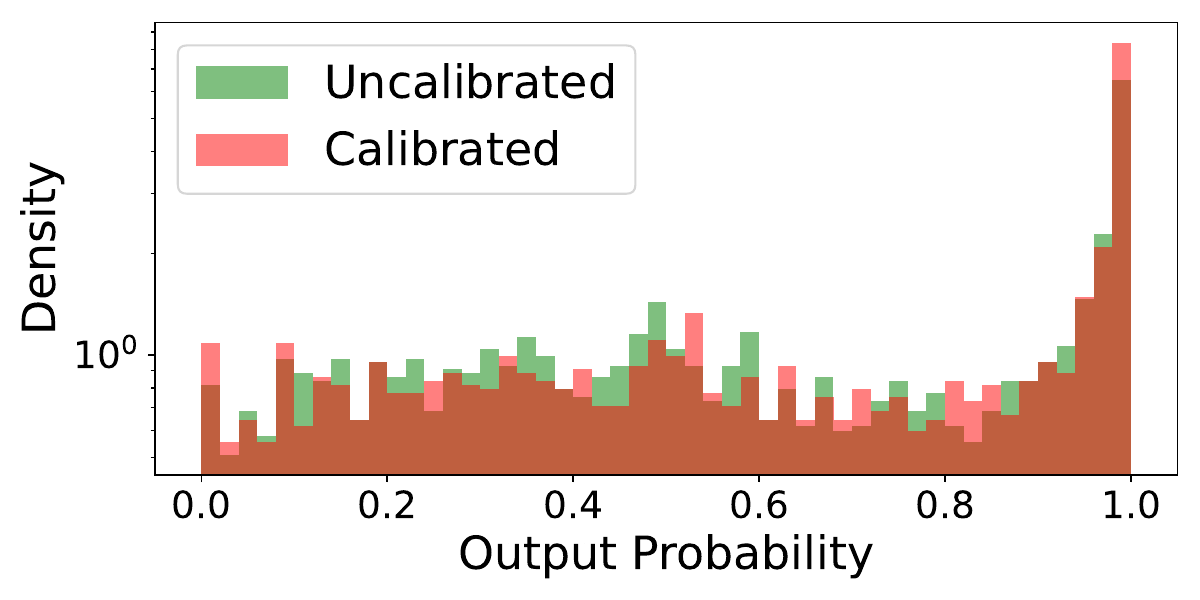}
    \caption{[$0, 0.15$] high var}
\end{subfigure}

\begin{subfigure}[b]{0.245\textwidth}
    \centering
    \includegraphics[width=\linewidth]{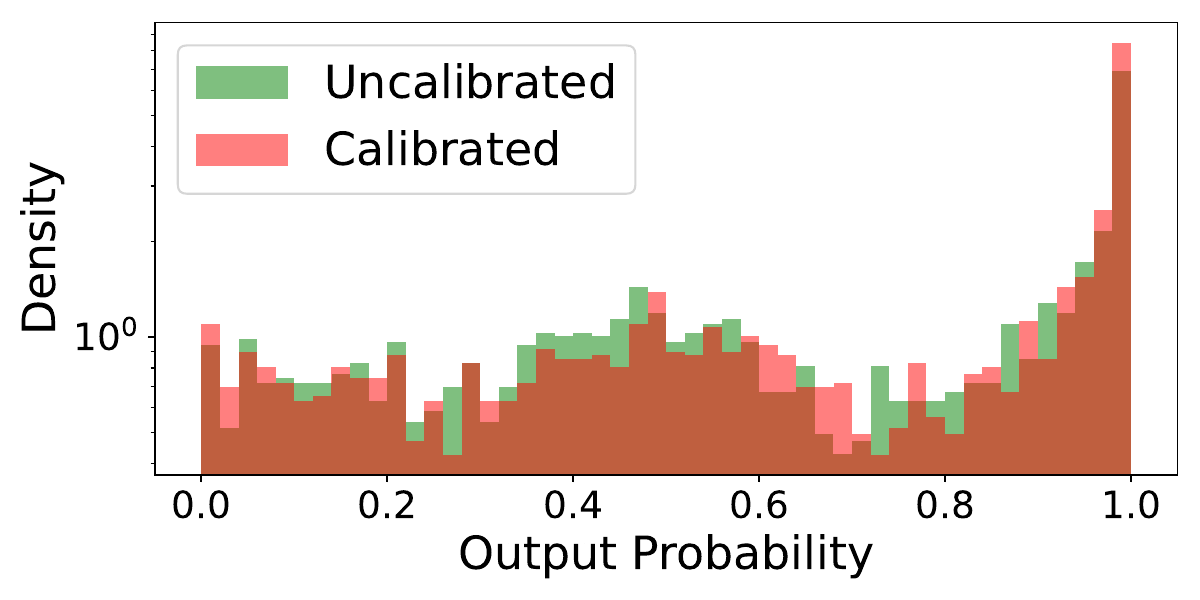}
    \caption{[0.15, 0.4] low var}
\end{subfigure}
\hfill
\begin{subfigure}[b]{0.245\textwidth}
    \centering
    \includegraphics[width=\linewidth]{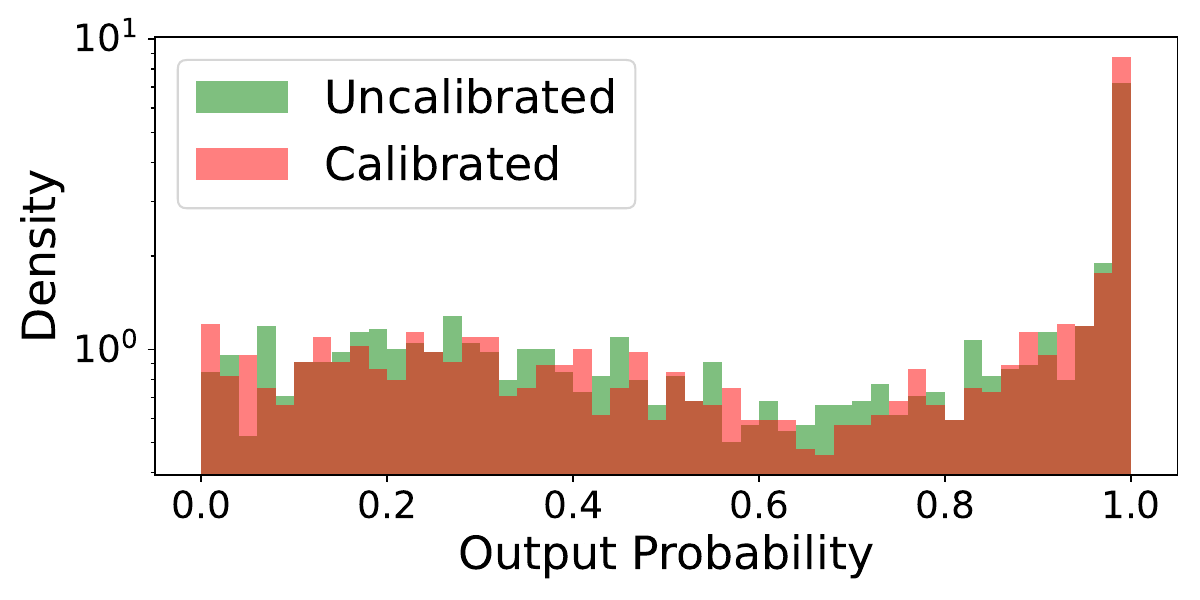}
    \caption{[0.15, 0.4] high var}
\end{subfigure}
\hfill
\begin{subfigure}[b]{0.245\textwidth}
    \centering
    \includegraphics[width=\linewidth]{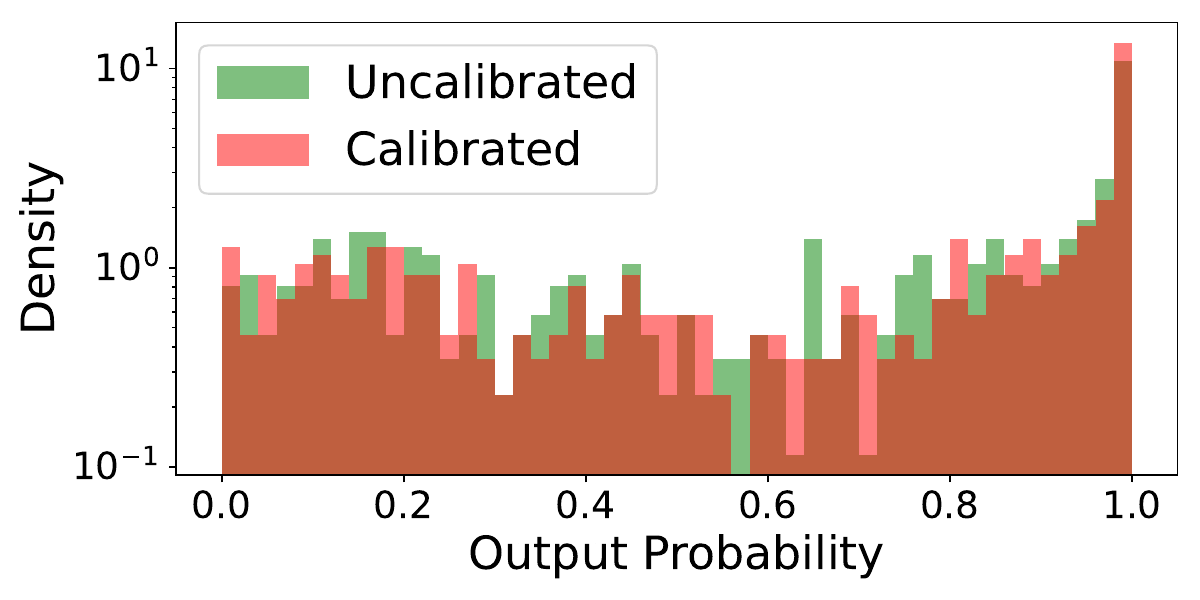}
    \caption{[0.4, $\infty$] low var}
\end{subfigure}
\hfill
\begin{subfigure}[b]{0.245\textwidth}
    \centering
    \includegraphics[width=\linewidth]{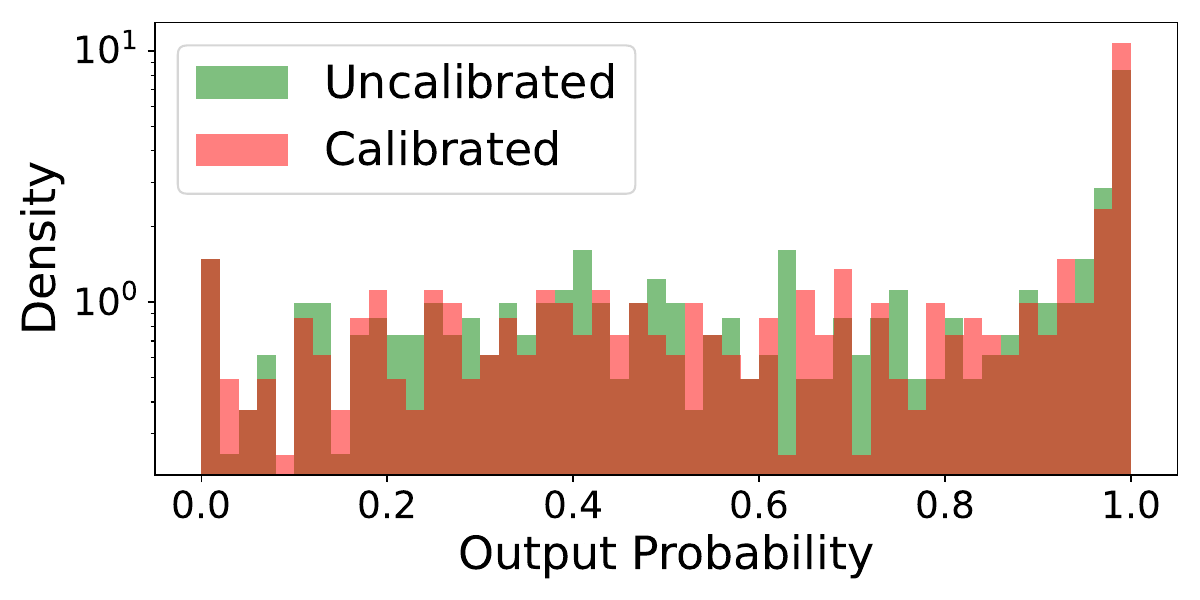}
    \caption{[0.4, $\infty$] high var}
\end{subfigure}
\caption{Comparison of output probability between calibrated and uncalibrated $\hat{\pi_{\mathcal{D}}}$. The plot shows a histogram of output probability and the number of counts in the dataset with a logarithm scale on the y-axis of the 12 stratified patient groups, where 'low var' and 'high var' mean low and high NEWS2 score variations, respectively.}
\end{figure}

%% file: appendix_sections/model_calibration.tex
\section{Model Calibration}
\label{sec:app-model calibration}
We implemented temperature scaling \cite{guo2017calibration} to calibrate the behavior model. Temperature scaling is a post-processing technique that adjusts the model's predicted probabilities to better match the observed probabilities, thereby (potentially) improving the model's calibration. We present the implementation details, OPE results, and a visual comparison between the calibrated and uncalibrated models.

\subsection{A brief Introduction to Temperature Scaling}
Formally, let \(f(x)\) represent the logits output by a neural network for a given input \(x\), and let \(P(y|x)\) denote the predicted probability distribution over classes \(y\), obtained by applying the softmax function. Temperature scaling introduces a temperature parameter \(T > 0\) to adjust this distribution as follows:
\begin{equation}
P_T(y|x) = \text{softmax}\left(\frac{f(x)}{T}\right)
\end{equation}

Here, the softmax function is defined as \(\text{softmax}(z_i) = \frac{e^{z_i}}{\sum_j e^{z_j}}\) for logits \(z_i\), where the summation in the denominator spans all class logits for the instance. The temperature \(T\) serves to "soften" (\(T > 1\)) or "sharpen" (\(T < 1\)) the probability distribution, with \(T = 1\) leaving the original predictions unchanged.

The optimal value of \(T\) is typically determined through a calibration process on the validation dataset, aiming to minimize a calibration-specific loss function. Here, we choose the Negative Log Likelihood (NLL). This optimization can be succinctly expressed as:
\begin{equation}
T^* = \arg\min_T \, L_{\text{calib}}(P_T(y|x), y_{\text{true}})    
\end{equation}

where \(y_{\text{true}}\) represents the true class labels in the validation dataset. We search hyperparameters to determine the best-calibrated model (See Table \ref{table:hp}).
\begin{table}[h]
\centering
\begin{tabular}{|c|c|}
\hline
learning rate& [0.1, 0.05, 0.02, 
                            0.01, 0.001, 5e-4, 
                            1e-4, 5e-5, 1e-5]\\
\hline
batch size& [512]\\
\hline
\end{tabular}
\caption{Hyperparameter search for temperature scaling calibration on the behavior policy.}
\label{table:hp}
\end{table}

\subsection{Comparing Calibrated and Uncalibrated Behavior Models on OPE}
To further investigate the impact of model calibration on OPE, we ran OPE for all naive baselines using the calibrated behavior policy. The results are presented in Tables \ref{table: calibrated-Outcome-all}, \ref{table: calibrated-SOFA-all} and \ref{table: calibrated-NEWS2-all}.  Since we do not have access to the ground truth reward estimates for the naive baselines, we use the criterion 'higher than $G_\cD$' as a sanity check: \textbf{If a naive baseline can surpass the performance of clinicians according to the OPE results, it suggests that the OPE method may not be reliable.} This is because we expect clinicians who have extensive domain knowledge to generally outperform naive baselines. The table results show that some naive baselines can still surpass clinical experts, regardless of the reward design. This result again supports our position of reevaluating DTR and indicates that model calibration may not be universally helpful in DTR.

\begin{table}[ht]
  \centering
  \begin{minipage}{0.40\textwidth}
    \centering
    \tiny
        \begin{tabular}{cccccc}
        \hline
        Policy Name & $\text{WIS}$ & $\text{WIS}_{b}$ & $\text{WIS}_{t}$ & $\text{WIS}_{bt}$ & DR \\
        \hline
        alt & \textbf{84.29} & \textbf{79.54} & \textbf{84.29} & \textbf{72.97} & -0.68 \\
        max & -90.47 & -74.67 & -90.47 & -77.09 & -0.38 \\
        min & \textbf{84.29} & \textbf{84.29} & \textbf{84.29} & \textbf{84.24} & -0.91 \\
        random & \textbf{86.0} & \textbf{85.48} & \textbf{84.3} & \textbf{82.87} & -0.58 \\
        weight & \textbf{86.87} & \textbf{63.98} & \textbf{86.87} & 57.87 & -0.5 \\
        \hline
        $G_{\cD}$ & \multicolumn{5}{c}{61.54}\\
        \hline
        \end{tabular}
    \caption{Outcome reward}
    \label{table: calibrated-Outcome-all}
  \end{minipage}\hfill
  \begin{minipage}{0.29\textwidth}
    \centering
    \tiny
        \begin{tabular}{ccccc}
        \hline
        $\text{WIS}$ & $\text{WIS}_{b}$ & $\text{WIS}_{t}$ & $\text{WIS}_{bt}$ & DR \\
        \hline
        \textbf{13.34} & \textbf{12.28} & \textbf{13.34} & \textbf{11.42} & -0.54 \\
        -9.44 & -10.03 & -9.44 & -9.53 & -0.3 \\
        \textbf{13.34} & \textbf{12.86} & \textbf{13.34} & \textbf{12.93} & -0.73 \\
        \textbf{14.33} & \textbf{13.35} & \textbf{10.26} & \textbf{11.86} & -0.41 \\
        \textbf{16.43} & \textbf{11.24} & \textbf{16.43} & \textbf{9.96} & -0.38 \\
        \hline
        \multicolumn{5}{c}{9.47}\\
        \hline
        \end{tabular}
    \caption{SOFA reward}
    \label{table: calibrated-SOFA-all}
  \end{minipage}\hfill
  \begin{minipage}{0.29\textwidth}
    \centering
    \tiny
        \begin{tabular}{ccccc}
        \hline
        $\text{WIS}$ & $\text{WIS}_{b}$ & $\text{WIS}_{t}$ & $\text{WIS}_{bt}$ & DR \\
        \hline
        -4.57 & -5.63 & -4.57 & -5.43 & \textbf{-0.56} \\
        -4.6 & -4.85 & -4.6 & -4.63 & \textbf{-0.21} \\
        -4.57 & -5.59 & -4.57 & -5.56 & \textbf{-1.49} \\
        -4.47 & -4.77 & -5.82 & -5.09 & \textbf{-0.39} \\
        \textbf{-3.78} & -4.53 & \textbf{-3.78} & -4.61 & \textbf{-0.35} \\
        \hline
        \multicolumn{5}{c}{-4.39}\\
        \hline
        \end{tabular}
    \caption{NEWS2 reward}
    \label{table: calibrated-NEWS2-all}
  \end{minipage}
\end{table}

In conclusion, our analysis demonstrates that temperature scaling may not be sufficient to mitigate the challenges associated with OPE in the presence of small probabilities. Researchers should be aware of the limitations and consider exploring alternative OPE methods or implementing other calibration methods to obtain more reliable policy evaluations.

%% file: appendix_sections/6full_result.tex
\newpage
\section{Full Result}

\label{sec:app-full result}
\subsection{Full Result on Test set}

\input{rl_tables/all/MIMIC3SepsisOutcomeEnv-all_train-all-table}

\input{rl_tables/all/MIMIC3SepsisSOFAEnv-all_train-all-table}

\newpage
\subsection{Full Result on Stratified Subsets}
\input{rl_tables/sub/MIMIC3SepsisOutcomeEnv-all_train-sub-test_rate_--0.4_low_std-table}

\input{rl_tables/sub/MIMIC3SepsisOutcomeEnv-all_train-sub-test_rate_--0.4_high_std-table}
\input{rl_tables/sub/MIMIC3SepsisOutcomeEnv-all_train-sub-test_rate_-0.4--0.15_low_std-table}
\input{rl_tables/sub/MIMIC3SepsisOutcomeEnv-all_train-sub-test_rate_-0.4--0.15_high_std-table}
\input{rl_tables/sub/MIMIC3SepsisOutcomeEnv-all_train-sub-test_rate_-0.15-0_low_std-table}
\input{rl_tables/sub/MIMIC3SepsisOutcomeEnv-all_train-sub-test_rate_-0.15-0_high_std-table}
\input{rl_tables/sub/MIMIC3SepsisOutcomeEnv-all_train-sub-test_rate_0-0.15_low_std-table}
\input{rl_tables/sub/MIMIC3SepsisOutcomeEnv-all_train-sub-test_rate_0-0.15_high_std-table}
\input{rl_tables/sub/MIMIC3SepsisOutcomeEnv-all_train-sub-test_rate_0.15-0.4_low_std-table}
\input{rl_tables/sub/MIMIC3SepsisOutcomeEnv-all_train-sub-test_rate_0.15-0.4_high_std-table}
\input{rl_tables/sub/MIMIC3SepsisOutcomeEnv-all_train-sub-test_rate_0.4-_low_std-table}
\input{rl_tables/sub/MIMIC3SepsisOutcomeEnv-all_train-sub-test_rate_0.4-_high_std-table}

\input{rl_tables/sub/MIMIC3SepsisSOFAEnv-all_train-sub-test_rate_--0.4_low_std-table}
\input{rl_tables/sub/MIMIC3SepsisSOFAEnv-all_train-sub-test_rate_--0.4_high_std-table}
\input{rl_tables/sub/MIMIC3SepsisSOFAEnv-all_train-sub-test_rate_-0.4--0.15_low_std-table}
\input{rl_tables/sub/MIMIC3SepsisSOFAEnv-all_train-sub-test_rate_-0.4--0.15_high_std-table}
\input{rl_tables/sub/MIMIC3SepsisSOFAEnv-all_train-sub-test_rate_-0.15-0_low_std-table}
\input{rl_tables/sub/MIMIC3SepsisSOFAEnv-all_train-sub-test_rate_-0.15-0_high_std-table}
\input{rl_tables/sub/MIMIC3SepsisSOFAEnv-all_train-sub-test_rate_0-0.15_low_std-table}
\input{rl_tables/sub/MIMIC3SepsisSOFAEnv-all_train-sub-test_rate_0-0.15_high_std-table}
\input{rl_tables/sub/MIMIC3SepsisSOFAEnv-all_train-sub-test_rate_0.15-0.4_low_std-table}
\input{rl_tables/sub/MIMIC3SepsisSOFAEnv-all_train-sub-test_rate_0.15-0.4_high_std-table}
\input{rl_tables/sub/MIMIC3SepsisSOFAEnv-all_train-sub-test_rate_0.4-_low_std-table}
\input{rl_tables/sub/MIMIC3SepsisSOFAEnv-all_train-sub-test_rate_0.4-_high_std-table}

\input{rl_tables/sub/MIMIC3SepsisNEWS2Env-all_train-sub-test_rate_--0.4_low_std-table}
\input{rl_tables/sub/MIMIC3SepsisNEWS2Env-all_train-sub-test_rate_-0.4--0.15_low_std-table}
\input{rl_tables/sub/MIMIC3SepsisNEWS2Env-all_train-sub-test_rate_-0.4--0.15_high_std-table}
\input{rl_tables/sub/MIMIC3SepsisNEWS2Env-all_train-sub-test_rate_-0.15-0_low_std-table}
\input{rl_tables/sub/MIMIC3SepsisNEWS2Env-all_train-sub-test_rate_-0.15-0_high_std-table}
\input{rl_tables/sub/MIMIC3SepsisNEWS2Env-all_train-sub-test_rate_0-0.15_low_std-table}
\input{rl_tables/sub/MIMIC3SepsisNEWS2Env-all_train-sub-test_rate_0-0.15_high_std-table}
\input{rl_tables/sub/MIMIC3SepsisNEWS2Env-all_train-sub-test_rate_0.15-0.4_low_std-table}
\input{rl_tables/sub/MIMIC3SepsisNEWS2Env-all_train-sub-test_rate_0.15-0.4_high_std-table}
\input{rl_tables/sub/MIMIC3SepsisNEWS2Env-all_train-sub-test_rate_0.4-_low_std-table}
\input{rl_tables/sub/MIMIC3SepsisNEWS2Env-all_train-sub-test_rate_0.4-_high_std-table}

%% file: rl_tables/all/MIMIC3SepsisOutcomeEnv-all_train-all-table.tex
\begin{table}[h]
\centering
\tiny
\begin{tabular}{lcccccccccc}
\toprule
metric & alt & max & min & random & weight & SL & DQN & CQL & IQL & BCQ \\
\midrule
$\text{RMSE}_{\text{IV}}$ & ${763.89}$ & ${861.51}$ & ${645.83}$ & ${671.39}$ & ${645.83}$ & \textcolor{red}{$556.48 \pm 6.23$} & ${624.06 \pm 17.85}$ & ${616.71 \pm 17.94}$ & \textcolor{blue}{$578.13 \pm 6.14$} & ${620.74 \pm 2.12}$ \\
$\text{RMSE}_{\text{vaso}}$ & ${0.67}$ & ${0.89}$ & \textcolor{blue}{$0.32$} & ${0.5}$ & ${0.59}$ & \textcolor{red}{$0.31$} & \textcolor{blue}{$0.32$} & \textcolor{red}{$0.31$} & \textcolor{red}{$0.31$} & \textcolor{red}{$0.31$} \\
$\text{WIS}$ & ${84.29}$ & ${-89.05}$ & ${84.29}$ & ${86.8}$ & ${86.87}$ & ${86.66 \pm 0.2}$ & \textcolor{blue}{$87.62 \pm 3.34$} & \textcolor{red}{$89.37 \pm 2.95$} & ${84.29}$ & ${84.29}$ \\
$\text{WIS}_{b}$ & ${81.21}$ & ${-75.94}$ & ${84.05}$ & ${84.35}$ & ${54.6}$ & ${85.05 \pm 0.95}$ & \textcolor{blue}{$87.52 \pm 1.33$} & \textcolor{red}{$88.26 \pm 1.82$} & ${77.57 \pm 6.87}$ & ${84.52 \pm 0.17}$ \\
$\text{WIS}_{t}$ & ${84.29}$ & ${-89.05}$ & ${84.29}$ & ${86.64}$ & ${86.87}$ & ${86.3 \pm 0.52}$ & \textcolor{blue}{$87.62 \pm 3.34$} & \textcolor{red}{$89.37 \pm 2.95$} & ${84.29}$ & ${84.29}$ \\
$\text{WIS}_{bt}$ & ${79.85}$ & ${-79.75}$ & ${84.23}$ & ${80.84}$ & ${66.22}$ & ${84.74 \pm 0.99}$ & \textcolor{blue}{$87.6 \pm 1.33$} & \textcolor{red}{$88.28 \pm 1.74$} & ${75.85 \pm 4.38}$ & ${84.51 \pm 0.16}$ \\
DR & ${-0.66}$ & \textcolor{blue}{$-0.38$} & ${-0.95}$ & ${-0.58}$ & ${-0.49}$ & ${-0.54 \pm 0.01}$ & \textcolor{red}{$-0.35 \pm 0.05$} & ${-0.51 \pm 0.08}$ & ${-0.54 \pm 0.03}$ & ${-0.95}$ \\
\bottomrule

\end{tabular}
\caption{Outcome all}
\label{table: Outcome-all}
\end{table}

%% file: rl_tables/all/MIMIC3SepsisSOFAEnv-all_train-all-table.tex
\begin{table}[h]
\centering
\tiny
\begin{tabular}{lcccccccccc}
\toprule
metric & alt & max & min & random & weight & SL & DQN & CQL & IQL & BCQ \\
\midrule
$\text{RMSE}_{\text{IV}}$ & ${763.89}$ & ${861.51}$ & ${645.83}$ & ${671.39}$ & ${645.83}$ & \textcolor{red}{$536.36 \pm 13.56$} & ${628.24 \pm 16.8}$ & ${582.33 \pm 10.28}$ & \textcolor{blue}{$578.13 \pm 6.14$} & ${627.97 \pm 1.67}$ \\
$\text{RMSE}_{\text{vaso}}$ & ${0.67}$ & ${0.89}$ & ${0.32}$ & ${0.5}$ & ${0.59}$ & \textcolor{red}{$0.3$} & ${0.32}$ & ${0.32}$ & \textcolor{blue}{$0.31$} & \textcolor{blue}{$0.31$} \\
$\text{WIS}$ & ${13.34}$ & ${-12.25}$ & ${13.34}$ & \textcolor{blue}{$16.26$} & \textcolor{red}{$16.43$} & ${15.49 \pm 0.91}$ & ${12.41 \pm 0.72}$ & ${13.84 \pm 1.79}$ & ${10.24}$ & ${13.34}$ \\
$\text{WIS}_{b}$ & ${12.49}$ & ${-11.3}$ & ${12.86}$ & \textcolor{red}{$14.19$} & ${9.52}$ & \textcolor{blue}{$13.78 \pm 0.66$} & ${12.48 \pm 0.78}$ & ${11.85 \pm 3.78}$ & ${11.08 \pm 0.62}$ & ${12.97 \pm 0.12}$ \\
$\text{WIS}_{t}$ & ${13.34}$ & ${-12.25}$ & ${13.34}$ & \textcolor{blue}{$15.86$} & \textcolor{red}{$16.43$} & ${14.36 \pm 1.57}$ & ${12.41 \pm 0.72}$ & ${13.84 \pm 1.79}$ & ${10.24}$ & ${13.34}$ \\
$\text{WIS}_{bt}$ & ${12.19}$ & ${-11.25}$ & ${12.92}$ & \textcolor{blue}{$13.34$} & ${11.49}$ & \textcolor{red}{$13.44 \pm 0.87$} & ${12.27 \pm 1.23}$ & ${11.76 \pm 3.49}$ & ${10.73 \pm 0.44}$ & ${12.91 \pm 0.03}$ \\
DR & ${-0.5}$ & \textcolor{blue}{$-0.27$} & ${-0.76}$ & ${-0.38}$ & ${-0.37}$ & ${-0.35 \pm 0.01}$ & \textcolor{red}{$-0.25 \pm 0.06$} & ${-0.37 \pm 0.1}$ & ${-0.36 \pm 0.02}$ & ${-0.76}$ \\
\bottomrule

\end{tabular}
\caption{SOFA all}
\label{table: SOFA-all}
\end{table}

%% file: rl_tables/sub/MIMIC3SepsisOutcomeEnv-all_train-sub-test_rate_--0.4_low_std-table.tex
\begin{table}[h]
\centering
\tablestyle{4pt}{1.0}
\tiny
\begin{tabular}{lcccccccccc}
\toprule
metric & alt & max & min & random & weight & imitation & dqn & cql & iql & bcq \\
\midrule
$\text{RMSE}_{\text{IV}}$ & ${880.57}$ & ${929.48}$ & ${825.16}$ & ${821.09}$ & ${828.29}$ & \textcolor{red}{$682.74 \pm 8.09$} & ${802.31 \pm 20.11}$ & ${793.05 \pm 13.33}$ & \textcolor{blue}{$719.46 \pm 13.2$} & ${795.1 \pm 5.0}$ \\
$\text{RMSE}_{\text{vaso}}$ & ${0.71}$ & ${0.87}$ & ${0.2}$ & ${0.44}$ & ${0.55}$ & \textcolor{blue}{$0.19 \pm 0.01$} & ${0.2}$ & ${0.2}$ & \textcolor{blue}{$0.19 \pm 0.01$} & \textcolor{red}{$0.18$} \\
$\text{WIS}$ & \textcolor{red}{$93.3$} & ${86.87}$ & ${90.77}$ & ${86.87}$ & ${86.87}$ & ${89.34 \pm 3.42}$ & ${92.01 \pm 1.46}$ & \textcolor{blue}{$92.72 \pm 3.19$} & ${91.95 \pm 1.68}$ & ${91.74 \pm 0.89}$ \\
$\text{WIS}_{b}$ & \textcolor{red}{$92.12$} & ${76.88}$ & ${91.06}$ & ${62.66}$ & ${86.34}$ & ${88.92 \pm 2.47}$ & ${91.35 \pm 0.68}$ & ${91.05 \pm 0.68}$ & ${90.11 \pm 1.23}$ & \textcolor{blue}{$91.57 \pm 0.73$} \\
$\text{WIS}_{t}$ & \textcolor{red}{$93.3$} & ${86.87}$ & ${90.77}$ & ${86.87}$ & ${86.87}$ & ${89.34 \pm 3.42}$ & ${92.01 \pm 1.46}$ & \textcolor{blue}{$92.72 \pm 3.19$} & ${91.95 \pm 1.68}$ & ${91.74 \pm 0.89}$ \\
$\text{WIS}_{bt}$ & \textcolor{red}{$92.32$} & ${74.96}$ & ${90.89}$ & ${72.06}$ & ${87.75}$ & ${89.43 \pm 2.0}$ & ${91.56 \pm 0.76}$ & ${90.97 \pm 1.06}$ & ${90.54 \pm 0.78}$ & \textcolor{blue}{$91.59 \pm 0.63$} \\
DR & ${-0.22}$ & ${-0.43}$ & ${-0.73}$ & ${-0.22}$ & ${-0.22}$ & \textcolor{blue}{$-0.13 \pm 0.12$} & ${-0.22}$ & ${-0.3 \pm 0.1}$ & \textcolor{red}{$-0.07 \pm 0.1$} & ${-0.55}$ \\
\bottomrule

\end{tabular}
\caption{Outcome sub rate  $\in[-\infty, -0.4]$ low std}
\label{table: Outcome reward-subgroup-test_rate_--0.4_low_std}
\end{table}

%% file: rl_tables/sub/MIMIC3SepsisOutcomeEnv-all_train-sub-test_rate_--0.4_high_std-table.tex
\begin{table}[h]
\centering
\tablestyle{4pt}{1.0}
\tiny

\begin{tabular}{lcccccccccc}
\toprule
metric & alt & max & min & random & weight & imitation & dqn & cql & iql & bcq \\
\midrule
$\text{RMSE}_{\text{IV}}$ & ${788.91}$ & ${880.8}$ & ${756.41}$ & ${774.78}$ & ${749.58}$ & \textcolor{red}{$637.96 \pm 8.04$} & ${729.93 \pm 14.25}$ & ${700.73 \pm 25.47}$ & \textcolor{blue}{$648.47 \pm 8.58$} & ${719.53 \pm 5.46}$ \\
$\text{RMSE}_{\text{vaso}}$ & ${0.54}$ & ${0.85}$ & ${0.27}$ & ${0.47}$ & ${0.56}$ & \textcolor{red}{$0.24 \pm 0.02$} & ${0.27}$ & ${0.27}$ & ${0.27 \pm 0.01}$ & \textcolor{blue}{$0.26$} \\
$\text{WIS}$ & ${23.66}$ & ${26.75}$ & ${-8.29}$ & ${88.59}$ & ${-94.02}$ & ${88.64}$ & ${87.46 \pm 13.66}$ & \textcolor{red}{$94.02 \pm 1.51$} & \textcolor{blue}{$89.36 \pm 1.52$} & ${29.48 \pm 34.13}$ \\
$\text{WIS}_{b}$ & ${14.77}$ & ${24.94}$ & ${-0.78}$ & ${44.85}$ & ${-39.81}$ & \textcolor{blue}{$90.85 \pm 0.16$} & ${83.92 \pm 14.47}$ & \textcolor{red}{$92.61 \pm 0.85$} & ${90.81 \pm 0.78}$ & ${33.48 \pm 22.12}$ \\
$\text{WIS}_{t}$ & ${23.66}$ & ${26.75}$ & ${-8.29}$ & ${88.59}$ & ${-94.02}$ & ${88.64}$ & ${87.46 \pm 13.66}$ & \textcolor{red}{$94.02 \pm 1.51$} & \textcolor{blue}{$89.36 \pm 1.52$} & ${29.48 \pm 34.13}$ \\
$\text{WIS}_{bt}$ & ${26.13}$ & ${23.87}$ & ${-0.81}$ & ${50.99}$ & ${-13.95}$ & \textcolor{blue}{$90.43 \pm 0.42$} & ${84.03 \pm 14.76}$ & \textcolor{red}{$93.13 \pm 1.1$} & ${90.28 \pm 0.75}$ & ${30.69 \pm 20.3}$ \\
DR & ${-0.12}$ & \textcolor{red}{$0.0$} & ${-0.3}$ & ${-0.27}$ & ${-0.4}$ & ${-0.06 \pm 0.08}$ & \textcolor{blue}{$-0.03 \pm 0.07$} & \textcolor{red}{$0.0$} & ${-0.11 \pm 0.02}$ & ${-0.12}$ \\
\bottomrule

\end{tabular}
\caption{Outcome sub rate  $\in[-\infty, -0.4]$ high std}
\label{table: Outcome reward-subgroup-test_rate_--0.4_high_std}
\end{table}

%% file: rl_tables/sub/MIMIC3SepsisOutcomeEnv-all_train-sub-test_rate_-0.4--0.15_low_std-table.tex
\begin{table}[h]
\centering
\tablestyle{4pt}{1.0}
\tiny
\begin{tabular}{lcccccccccc}
\toprule
metric & alt & max & min & random & weight & imitation & dqn & cql & iql & bcq \\
\midrule
$\text{RMSE}_{\text{IV}}$ & ${732.68}$ & ${849.88}$ & ${620.73}$ & ${649.77}$ & ${620.52}$ & \textcolor{red}{$531.04 \pm 6.78$} & ${599.25 \pm 17.6}$ & ${584.99 \pm 13.05}$ & \textcolor{blue}{$557.24 \pm 7.32$} & ${599.44 \pm 2.38}$ \\
$\text{RMSE}_{\text{vaso}}$ & ${0.6}$ & ${0.87}$ & \textcolor{blue}{$0.2$} & ${0.44}$ & ${0.55}$ & \textcolor{red}{$0.19 \pm 0.01$} & \textcolor{blue}{$0.2$} & \textcolor{red}{$0.19$} & \textcolor{blue}{$0.2 \pm 0.01$} & \textcolor{red}{$0.19$} \\
$\text{WIS}$ & ${84.29}$ & ${-87.75}$ & ${84.29}$ & ${-94.09}$ & ${-73.02}$ & ${-83.9 \pm 7.91}$ & \textcolor{blue}{$84.64 \pm 0.75$} & \textcolor{red}{$89.57 \pm 3.58$} & ${61.66 \pm 32.53}$ & ${84.52 \pm 0.21}$ \\
$\text{WIS}_{b}$ & \textcolor{blue}{$84.51$} & ${-54.45}$ & ${84.36}$ & ${-33.01}$ & ${-43.75}$ & ${-24.56 \pm 13.79}$ & ${84.4 \pm 0.08}$ & ${84.5 \pm 0.29}$ & ${64.9 \pm 25.17}$ & \textcolor{red}{$84.76 \pm 0.45$} \\
$\text{WIS}_{t}$ & ${84.29}$ & ${-87.75}$ & ${84.29}$ & ${-94.09}$ & ${-73.02}$ & ${-83.91 \pm 7.92}$ & \textcolor{blue}{$84.64 \pm 0.75$} & \textcolor{red}{$89.57 \pm 3.58$} & ${61.62 \pm 32.57}$ & ${84.52 \pm 0.21}$ \\
$\text{WIS}_{bt}$ & ${84.41}$ & ${-76.1}$ & ${84.3}$ & ${-15.19}$ & ${-38.84}$ & ${-20.09 \pm 12.61}$ & ${84.39 \pm 0.11}$ & \textcolor{blue}{$84.44 \pm 0.22$} & ${63.09 \pm 28.61}$ & \textcolor{red}{$84.78 \pm 0.39$} \\
DR & ${-0.33}$ & \textcolor{red}{$-0.04$} & ${-0.67}$ & ${-0.18}$ & ${-0.13}$ & ${-0.07 \pm 0.02}$ & \textcolor{blue}{$-0.06 \pm 0.04$} & ${-0.13 \pm 0.07}$ & ${-0.08 \pm 0.02}$ & ${-0.67}$ \\
\bottomrule

\end{tabular}
\caption{Outcome sub rate  $\in[-0.4, -0.15]$ low std}
\label{table: Outcome reward-subgroup-test_rate_-0.4--0.15_low_std}
\end{table}

%% file: rl_tables/sub/MIMIC3SepsisOutcomeEnv-all_train-sub-test_rate_-0.4--0.15_high_std-table.tex
\begin{table}[h]
\centering
\tablestyle{4pt}{1.0}
\tiny
\begin{tabular}{lcccccccccc}
\toprule
metric & alt & max & min & random & weight & imitation & dqn & cql & iql & bcq \\
\midrule
$\text{RMSE}_{\text{IV}}$ & ${820.25}$ & ${887.4}$ & ${770.94}$ & ${771.37}$ & ${767.21}$ & \textcolor{red}{$661.62 \pm 8.98$} & ${745.14 \pm 22.99}$ & ${709.98 \pm 19.82}$ & \textcolor{blue}{$675.19 \pm 6.55$} & ${738.97 \pm 2.38}$ \\
$\text{RMSE}_{\text{vaso}}$ & ${0.82}$ & ${1.03}$ & \textcolor{blue}{$0.65$} & ${0.74}$ & ${0.81}$ & \textcolor{red}{$0.64 \pm 0.01$} & \textcolor{blue}{$0.65$} & \textcolor{red}{$0.64$} & \textcolor{red}{$0.64 \pm 0.01$} & \textcolor{red}{$0.64$} \\
$\text{WIS}$ & ${84.29}$ & ${84.29}$ & ${84.29}$ & ${88.63}$ & ${85.02}$ & ${88.62 \pm 0.01}$ & \textcolor{blue}{$89.24 \pm 3.67$} & \textcolor{red}{$91.67 \pm 0.59$} & ${87.94 \pm 2.25}$ & ${84.29}$ \\
$\text{WIS}_{b}$ & ${84.04}$ & ${69.52}$ & ${84.84}$ & ${86.01}$ & ${86.88}$ & ${88.25 \pm 1.19}$ & \textcolor{blue}{$89.04 \pm 3.6$} & \textcolor{red}{$89.94 \pm 0.39$} & ${88.07 \pm 1.34}$ & ${85.4 \pm 0.59}$ \\
$\text{WIS}_{t}$ & ${84.29}$ & ${84.29}$ & ${84.29}$ & ${88.63}$ & ${85.02}$ & ${88.62 \pm 0.01}$ & \textcolor{blue}{$89.24 \pm 3.67$} & \textcolor{red}{$91.67 \pm 0.59$} & ${87.94 \pm 2.25}$ & ${84.29}$ \\
$\text{WIS}_{bt}$ & ${83.92}$ & ${67.52}$ & ${81.18}$ & ${88.22}$ & ${86.39}$ & ${88.04 \pm 0.99}$ & \textcolor{blue}{$88.85 \pm 3.57$} & \textcolor{red}{$89.59 \pm 0.37$} & ${87.66 \pm 1.73}$ & ${84.86 \pm 0.73}$ \\
DR & ${-0.55}$ & ${-0.45}$ & ${-0.67}$ & ${-0.28}$ & \textcolor{blue}{$-0.26$} & ${-0.27 \pm 0.01}$ & \textcolor{red}{$-0.22 \pm 0.08$} & \textcolor{blue}{$-0.26 \pm 0.12$} & ${-0.33 \pm 0.06}$ & ${-0.67}$ \\
\bottomrule

\end{tabular}
\caption{Outcome sub rate  $\in[-0.4, -0.15]$ high std}
\label{table: Outcome reward-subgroup-test_rate_-0.4--0.15_high_std}
\end{table}

%% file: rl_tables/sub/MIMIC3SepsisOutcomeEnv-all_train-sub-test_rate_-0.15-0_low_std-table.tex
\begin{table}[h]
\centering
\tablestyle{4pt}{1.0}
\tiny
\begin{tabular}{lcccccccccc}
\toprule
metric & alt & max & min & random & weight & imitation & dqn & cql & iql & bcq \\
\midrule
$\text{RMSE}_{\text{IV}}$ & ${688.63}$ & ${833.0}$ & ${457.4}$ & ${532.53}$ & ${486.77}$ & \textcolor{red}{$408.54 \pm 9.26$} & ${443.51 \pm 7.56}$ & ${442.82 \pm 2.72}$ & \textcolor{blue}{$428.3 \pm 8.78$} & ${438.72 \pm 1.4}$ \\
$\text{RMSE}_{\text{vaso}}$ & ${0.67}$ & ${0.89}$ & \textcolor{blue}{$0.26$} & ${0.48}$ & ${0.58}$ & \textcolor{red}{$0.25$} & \textcolor{blue}{$0.26$} & \textcolor{blue}{$0.26$} & \textcolor{blue}{$0.26$} & \textcolor{red}{$0.25 \pm 0.01$} \\
$\text{WIS}$ & ${84.29}$ & ${-90.44}$ & ${84.29}$ & ${-84.32}$ & ${-84.21}$ & ${84.3}$ & \textcolor{red}{$92.01 \pm 1.19$} & \textcolor{blue}{$90.24 \pm 2.64$} & ${84.29 \pm 0.05}$ & ${84.29}$ \\
$\text{WIS}_{b}$ & ${34.74}$ & ${-35.59}$ & ${71.32}$ & ${-60.3}$ & ${-62.77}$ & ${84.21 \pm 0.37}$ & \textcolor{red}{$90.89 \pm 0.74$} & \textcolor{blue}{$86.12 \pm 7.18$} & ${83.42 \pm 2.45}$ & ${81.81 \pm 4.11}$ \\
$\text{WIS}_{t}$ & ${84.29}$ & ${-90.44}$ & ${84.29}$ & ${-84.32}$ & ${-84.21}$ & ${84.32 \pm 0.05}$ & \textcolor{red}{$92.01 \pm 1.19$} & \textcolor{blue}{$90.24 \pm 2.64$} & ${83.84 \pm 1.0}$ & ${84.29}$ \\
$\text{WIS}_{bt}$ & ${40.39}$ & ${-37.48}$ & ${64.83}$ & ${-68.6}$ & ${-62.39}$ & ${84.22 \pm 0.78}$ & \textcolor{red}{$91.06 \pm 0.79$} & \textcolor{blue}{$85.71 \pm 8.78$} & ${82.2 \pm 5.04}$ & ${81.1 \pm 6.22}$ \\
DR & ${-0.51}$ & \textcolor{red}{$-0.15$} & ${-0.97}$ & ${-0.3}$ & \textcolor{blue}{$-0.18$} & ${-0.28 \pm 0.03}$ & ${-0.23 \pm 0.12}$ & ${-0.36 \pm 0.09}$ & ${-0.21 \pm 0.05}$ & ${-0.97}$ \\
\bottomrule

\end{tabular}
\caption{Outcome sub rate  $\in[-0.15, 0]$ low std}
\label{table: Outcome reward-subgroup-test_rate_-0.15-0_low_std}
\end{table}

%% file: rl_tables/sub/MIMIC3SepsisOutcomeEnv-all_train-sub-test_rate_-0.15-0_high_std-table.tex
\begin{table}[h]
\centering
\tablestyle{4pt}{1.0}
\tiny
\begin{tabular}{lcccccccccc}
\toprule
metric & alt & max & min & random & weight & imitation & dqn & cql & iql & bcq \\
\midrule
$\text{RMSE}_{\text{IV}}$ & ${765.09}$ & ${867.74}$ & ${669.04}$ & ${696.63}$ & ${662.95}$ & \textcolor{red}{$571.17 \pm 6.95$} & ${646.22 \pm 19.11}$ & ${636.61 \pm 18.79}$ & \textcolor{blue}{$597.7 \pm 6.59$} & ${642.96 \pm 2.08}$ \\
$\text{RMSE}_{\text{vaso}}$ & ${0.66}$ & ${0.87}$ & ${0.28}$ & ${0.48}$ & ${0.57}$ & \textcolor{red}{$0.26 \pm 0.01$} & ${0.28}$ & \textcolor{blue}{$0.27 \pm 0.01$} & \textcolor{blue}{$0.27 \pm 0.01$} & \textcolor{blue}{$0.27$} \\
$\text{WIS}$ & ${84.29}$ & ${-86.87}$ & ${84.29}$ & ${84.29}$ & ${84.2}$ & ${84.29}$ & \textcolor{blue}{$85.88 \pm 3.52$} & \textcolor{red}{$89.31 \pm 3.79$} & ${84.29}$ & ${84.29}$ \\
$\text{WIS}_{b}$ & ${51.74}$ & ${-50.46}$ & ${83.73}$ & ${68.07}$ & ${68.21}$ & ${66.15 \pm 4.7}$ & \textcolor{blue}{$85.84 \pm 0.92$} & \textcolor{red}{$87.91 \pm 3.06$} & ${48.93 \pm 8.66}$ & ${84.55 \pm 0.33}$ \\
$\text{WIS}_{t}$ & ${84.29}$ & ${-86.87}$ & ${84.29}$ & ${84.29}$ & ${84.2}$ & ${84.29}$ & \textcolor{blue}{$85.88 \pm 3.52$} & \textcolor{red}{$89.31 \pm 3.79$} & ${84.29}$ & ${84.29}$ \\
$\text{WIS}_{bt}$ & ${52.42}$ & ${-43.31}$ & ${83.83}$ & ${63.38}$ & ${63.12}$ & ${65.74 \pm 8.39}$ & \textcolor{blue}{$85.6 \pm 0.65$} & \textcolor{red}{$86.94 \pm 1.69$} & ${47.17 \pm 9.39}$ & ${84.48 \pm 0.16}$ \\
DR & ${-0.7}$ & ${-0.36}$ & ${-0.85}$ & ${-0.46}$ & \textcolor{red}{$-0.3$} & ${-0.46 \pm 0.02}$ & \textcolor{blue}{$-0.32 \pm 0.11$} & ${-0.4 \pm 0.11}$ & ${-0.52 \pm 0.05}$ & ${-0.85}$ \\
\bottomrule

\end{tabular}
\caption{Outcome sub rate  $\in[-0.15, 0]$ high std}
\label{table: Outcome reward-subgroup-test_rate_-0.15-0_high_std}
\end{table}

%% file: rl_tables/sub/MIMIC3SepsisOutcomeEnv-all_train-sub-test_rate_0-0.15_low_std-table.tex
\begin{table}[h]
\centering
\tablestyle{4pt}{1.0}
\tiny
\begin{tabular}{lcccccccccc}
\toprule
metric & alt & max & min & random & weight & imitation & dqn & cql & iql & bcq \\
\midrule
$\text{RMSE}_{\text{IV}}$ & ${756.44}$ & ${875.08}$ & ${576.23}$ & ${624.74}$ & ${597.18}$ & \textcolor{red}{$541.54 \pm 8.98$} & ${561.71 \pm 10.18}$ & ${562.25 \pm 10.25}$ & \textcolor{blue}{$555.25 \pm 12.44$} & ${559.56 \pm 1.79}$ \\
$\text{RMSE}_{\text{vaso}}$ & ${0.6}$ & ${0.87}$ & \textcolor{blue}{$0.13$} & ${0.44}$ & ${0.54}$ & \textcolor{blue}{$0.13$} & \textcolor{blue}{$0.13$} & \textcolor{blue}{$0.13$} & ${0.15 \pm 0.01}$ & \textcolor{red}{$0.12$} \\
$\text{WIS}$ & \textcolor{red}{$91.35$} & ${-68.28}$ & \textcolor{red}{$91.35$} & ${84.29}$ & ${84.29}$ & \textcolor{blue}{$87.58 \pm 1.96$} & \textcolor{red}{$91.35$} & \textcolor{red}{$91.35$} & ${86.2 \pm 4.71}$ & \textcolor{red}{$91.35$} \\
$\text{WIS}_{b}$ & ${86.25}$ & ${-10.25}$ & ${80.9}$ & ${75.39}$ & ${75.64}$ & ${86.06 \pm 1.04}$ & \textcolor{red}{$90.0 \pm 1.06$} & \textcolor{blue}{$89.88 \pm 0.77$} & ${85.69 \pm 0.54}$ & ${88.8 \pm 1.21}$ \\
$\text{WIS}_{t}$ & \textcolor{red}{$91.35$} & ${-68.28}$ & \textcolor{red}{$91.35$} & ${84.29}$ & ${84.29}$ & \textcolor{blue}{$87.64 \pm 1.95$} & \textcolor{red}{$91.35$} & \textcolor{red}{$91.35$} & ${86.07 \pm 4.83}$ & \textcolor{red}{$91.35$} \\
$\text{WIS}_{bt}$ & ${84.24}$ & ${-11.44}$ & ${80.24}$ & ${72.66}$ & ${74.64}$ & ${86.5 \pm 0.97}$ & \textcolor{red}{$90.14 \pm 1.0$} & \textcolor{blue}{$89.85 \pm 0.55$} & ${85.86 \pm 0.94}$ & ${88.38 \pm 1.6}$ \\
DR & ${-0.51}$ & \textcolor{red}{$-0.21$} & ${-0.99}$ & ${-0.85}$ & ${-0.85}$ & \textcolor{blue}{$-0.31 \pm 0.06$} & ${-0.41 \pm 0.11}$ & ${-0.7 \pm 0.26}$ & \textcolor{blue}{$-0.31 \pm 0.09$} & ${-0.87 \pm 0.23}$ \\
\bottomrule

\end{tabular}
\caption{Outcome sub rate $\in[0, 0.15]$ low std}
\label{table: Outcome reward-subgroup-test_rate_0-0.15_low_std}
\end{table}

%% file: rl_tables/sub/MIMIC3SepsisOutcomeEnv-all_train-sub-test_rate_0-0.15_high_std-table.tex
\begin{table}[h]
\centering
\tablestyle{4pt}{1.0}
\tiny
\begin{tabular}{lcccccccccc}
\toprule
metric & alt & max & min & random & weight & imitation & dqn & cql & iql & bcq \\
\midrule
$\text{RMSE}_{\text{IV}}$ & ${760.98}$ & ${834.61}$ & ${628.74}$ & ${637.52}$ & ${628.39}$ & \textcolor{red}{$530.49 \pm 8.58$} & ${605.18 \pm 20.02}$ & ${587.38 \pm 22.11}$ & \textcolor{blue}{$559.66 \pm 7.59$} & ${601.25 \pm 2.88}$ \\
$\text{RMSE}_{\text{vaso}}$ & ${0.7}$ & ${0.88}$ & \textcolor{blue}{$0.3$} & ${0.49}$ & ${0.58}$ & \textcolor{red}{$0.29 \pm 0.01$} & \textcolor{blue}{$0.3$} & \textcolor{blue}{$0.3$} & \textcolor{red}{$0.29 \pm 0.01$} & \textcolor{red}{$0.29$} \\
$\text{WIS}$ & ${84.29}$ & ${-84.26}$ & ${84.29}$ & ${84.29}$ & ${84.29}$ & ${84.29 \pm 0.01}$ & \textcolor{red}{$89.92 \pm 3.21$} & \textcolor{blue}{$86.19 \pm 3.46$} & ${83.45 \pm 1.89}$ & ${84.72 \pm 0.39}$ \\
$\text{WIS}_{b}$ & ${84.6}$ & ${-47.35}$ & ${84.58}$ & ${47.19}$ & ${79.95}$ & ${77.89 \pm 6.35}$ & \textcolor{red}{$89.05 \pm 3.03$} & \textcolor{blue}{$86.03 \pm 1.7$} & ${72.98 \pm 16.41}$ & ${84.95 \pm 0.36}$ \\
$\text{WIS}_{t}$ & ${84.29}$ & ${-84.26}$ & ${84.29}$ & ${84.29}$ & ${84.29}$ & ${84.29}$ & \textcolor{red}{$89.92 \pm 3.21$} & \textcolor{blue}{$86.19 \pm 3.46$} & ${83.07 \pm 2.74}$ & ${84.72 \pm 0.39}$ \\
$\text{WIS}_{bt}$ & ${84.68}$ & ${-58.99}$ & ${84.57}$ & ${52.31}$ & ${73.77}$ & ${80.23 \pm 2.9}$ & \textcolor{red}{$89.87 \pm 2.85$} & \textcolor{blue}{$85.97 \pm 1.38$} & ${72.25 \pm 17.75}$ & ${84.93 \pm 0.35}$ \\
DR & ${-0.71}$ & \textcolor{red}{$-0.37$} & ${-1.02}$ & ${-0.67}$ & \textcolor{blue}{$-0.55$} & ${-0.69 \pm 0.03}$ & ${-0.56 \pm 0.15}$ & ${-0.73 \pm 0.15}$ & ${-0.7 \pm 0.08}$ & ${-0.92 \pm 0.1}$ \\
\bottomrule

\end{tabular}
\caption{Outcome sub rate $\in[0, 0.15]$ high std}
\label{table: Outcome reward-subgroup-test_rate_0-0.15_high_std}
\end{table}

%% file: rl_tables/sub/MIMIC3SepsisOutcomeEnv-all_train-sub-test_rate_0.15-0.4_low_std-table.tex
\begin{table}[h]
\centering
\tablestyle{4pt}{1.0}
\tiny
\begin{tabular}{lcccccccccc}
\toprule
metric & alt & max & min & random & weight & imitation & dqn & cql & iql & bcq \\
\midrule
$\text{RMSE}_{\text{IV}}$ & ${736.69}$ & ${853.28}$ & ${581.25}$ & ${616.67}$ & ${592.53}$ & \textcolor{red}{$497.49 \pm 11.07$} & ${560.83 \pm 15.47}$ & ${561.57 \pm 14.77}$ & \textcolor{blue}{$519.75 \pm 18.49$} & ${559.81 \pm 2.82}$ \\
$\text{RMSE}_{\text{vaso}}$ & ${0.6}$ & ${0.87}$ & \textcolor{blue}{$0.24$} & ${0.46}$ & ${0.57}$ & \textcolor{red}{$0.23 \pm 0.01$} & \textcolor{blue}{$0.24$} & \textcolor{blue}{$0.24$} & \textcolor{red}{$0.23 \pm 0.01$} & \textcolor{red}{$0.23$} \\
$\text{WIS}$ & ${-92.24}$ & ${-92.27}$ & ${87.46}$ & ${-88.64}$ & ${-84.29}$ & ${84.32 \pm 0.15}$ & ${84.32 \pm 0.06}$ & \textcolor{red}{$88.32 \pm 3.11$} & ${84.72 \pm 0.99}$ & \textcolor{blue}{$87.48 \pm 0.08$} \\
$\text{WIS}_{b}$ & ${-21.8}$ & ${-28.01}$ & ${88.38}$ & ${-79.87}$ & ${-80.21}$ & ${84.5 \pm 0.38}$ & ${85.71 \pm 0.7}$ & \textcolor{blue}{$88.67 \pm 1.14$} & ${83.33 \pm 2.44}$ & \textcolor{red}{$88.81 \pm 0.31$} \\
$\text{WIS}_{t}$ & ${-92.24}$ & ${-92.27}$ & ${87.46}$ & ${-88.64}$ & ${-84.29}$ & ${84.32 \pm 0.15}$ & ${84.32 \pm 0.06}$ & \textcolor{red}{$88.32 \pm 3.11$} & ${84.72 \pm 0.99}$ & \textcolor{blue}{$87.48 \pm 0.08$} \\
$\text{WIS}_{bt}$ & ${-14.44}$ & ${-33.15}$ & ${88.22}$ & ${-81.18}$ & ${-76.28}$ & ${84.24 \pm 0.66}$ & ${85.53 \pm 0.54}$ & \textcolor{blue}{$88.52 \pm 1.53$} & ${82.99 \pm 2.66}$ & \textcolor{red}{$88.79 \pm 0.49$} \\
DR & ${-0.85}$ & \textcolor{blue}{$-0.65$} & ${-1.15}$ & ${-0.99}$ & ${-0.93}$ & ${-0.99 \pm 0.01}$ & \textcolor{red}{$-0.56 \pm 0.06$} & ${-0.8 \pm 0.19}$ & ${-1.06 \pm 0.07}$ & ${-1.09 \pm 0.07}$ \\
\bottomrule

\end{tabular}
\caption{Outcome sub rate $\in[0.15, 0.4]$ low std}
\label{table: Outcome reward-subgroup-test_rate_0.15-0.4_low_std}
\end{table}

%% file: rl_tables/sub/MIMIC3SepsisOutcomeEnv-all_train-sub-test_rate_0.15-0.4_high_std-table.tex
\begin{table}[h]
\centering
\tablestyle{4pt}{1.0}
\tiny
\begin{tabular}{lcccccccccc}
\toprule
metric & alt & max & min & random & weight & imitation & dqn & cql & iql & bcq \\
\midrule
$\text{RMSE}_{\text{IV}}$ & ${782.93}$ & ${832.02}$ & ${699.16}$ & ${694.47}$ & ${686.84}$ & \textcolor{red}{$594.95 \pm 8.31$} & ${668.62 \pm 28.5}$ & ${653.72 \pm 18.89}$ & \textcolor{blue}{$624.3 \pm 16.96$} & ${672.05 \pm 2.09}$ \\
$\text{RMSE}_{\text{vaso}}$ & ${0.62}$ & ${0.85}$ & ${0.25}$ & ${0.45}$ & ${0.54}$ & \textcolor{blue}{$0.24 \pm 0.01$} & ${0.25}$ & ${0.25}$ & \textcolor{blue}{$0.24 \pm 0.01$} & \textcolor{red}{$0.23$} \\
$\text{WIS}$ & \textcolor{red}{$93.07$} & ${-84.29}$ & ${40.12}$ & ${-36.02}$ & ${-84.51}$ & ${84.2 \pm 0.09}$ & ${89.58 \pm 3.95}$ & \textcolor{blue}{$92.2 \pm 1.69$} & ${82.16 \pm 4.71}$ & ${40.17 \pm 0.03}$ \\
$\text{WIS}_{b}$ & ${66.26}$ & ${-32.94}$ & ${35.96}$ & ${-22.56}$ & ${-73.83}$ & ${79.76 \pm 2.53}$ & \textcolor{blue}{$89.41 \pm 2.37$} & \textcolor{red}{$90.54 \pm 0.5$} & ${74.85 \pm 9.66}$ & ${37.09 \pm 5.96}$ \\
$\text{WIS}_{t}$ & \textcolor{red}{$93.07$} & ${-84.29}$ & ${40.12}$ & ${-36.02}$ & ${-84.51}$ & ${84.2 \pm 0.09}$ & ${89.58 \pm 3.95}$ & \textcolor{blue}{$92.2 \pm 1.69$} & ${82.16 \pm 4.71}$ & ${40.17 \pm 0.03}$ \\
$\text{WIS}_{bt}$ & ${59.8}$ & ${-29.34}$ & ${36.47}$ & ${-24.04}$ & ${-76.8}$ & ${79.8 \pm 2.69}$ & \textcolor{blue}{$89.84 \pm 2.51$} & \textcolor{red}{$90.52 \pm 0.72$} & ${74.11 \pm 9.32}$ & ${37.15 \pm 2.23}$ \\
DR & ${-0.98}$ & \textcolor{blue}{$-0.46$} & ${-1.12}$ & ${-0.83}$ & ${-0.8}$ & ${-0.7 \pm 0.05}$ & \textcolor{red}{$-0.29 \pm 0.07$} & \textcolor{blue}{$-0.46 \pm 0.14$} & ${-0.83 \pm 0.09}$ & ${-0.76 \pm 0.13}$ \\
\bottomrule

\end{tabular}
\caption{Outcome sub rate $\in[0.15, 0.4]$ high std}
\label{table: Outcome reward-subgroup-test_rate_0.15-0.4_high_std}
\end{table}

%% file: rl_tables/sub/MIMIC3SepsisOutcomeEnv-all_train-sub-test_rate_0.4-_low_std-table.tex
\begin{table}[h]
\centering
\tablestyle{4pt}{1.0}
\tiny
\begin{tabular}{lcccccccccc}
\toprule
metric & alt & max & min & random & weight & imitation & dqn & cql & iql & bcq \\
\midrule
$\text{RMSE}_{\text{IV}}$ & ${828.61}$ & ${828.61}$ & ${626.63}$ & ${642.26}$ & ${637.93}$ & \textcolor{blue}{$553.92 \pm 13.64$} & ${596.82 \pm 30.51}$ & ${595.55 \pm 32.69}$ & \textcolor{red}{$537.16 \pm 6.65$} & ${596.26 \pm 4.61}$ \\
$\text{RMSE}_{\text{vaso}}$ & ${0.87}$ & ${0.87}$ & ${0.4}$ & ${0.53}$ & ${0.62}$ & \textcolor{red}{$0.29 \pm 0.03$} & ${0.38 \pm 0.01}$ & ${0.38 \pm 0.01}$ & \textcolor{blue}{$0.33 \pm 0.02$} & ${0.36 \pm 0.04}$ \\
$\text{WIS}$ & \textcolor{red}{$94.8$} & ${-89.68}$ & ${93.72}$ & ${-92.49}$ & ${-90.44}$ & ${-21.63 \pm 25.99}$ & \textcolor{blue}{$94.25 \pm 1.16$} & ${93.72}$ & ${-53.63 \pm 20.54}$ & ${93.72}$ \\
$\text{WIS}_{b}$ & ${71.67}$ & ${-90.58}$ & ${84.53}$ & ${-91.33}$ & ${-91.33}$ & ${-12.25 \pm 18.16}$ & ${71.45 \pm 13.1}$ & \textcolor{blue}{$89.53 \pm 2.77$} & ${-25.64 \pm 7.78}$ & \textcolor{red}{$89.86 \pm 1.19$} \\
$\text{WIS}_{t}$ & \textcolor{red}{$94.8$} & ${-89.68}$ & ${93.72}$ & ${-92.49}$ & ${-90.44}$ & ${-21.63 \pm 25.99}$ & \textcolor{blue}{$94.25 \pm 1.16$} & ${93.72}$ & ${-53.63 \pm 20.54}$ & ${93.72}$ \\
$\text{WIS}_{bt}$ & ${71.72}$ & ${-90.15}$ & ${90.31}$ & ${-90.64}$ & ${-91.15}$ & ${-10.64 \pm 13.05}$ & ${73.56 \pm 16.71}$ & \textcolor{red}{$91.28 \pm 1.72$} & ${-22.53 \pm 19.15}$ & \textcolor{blue}{$90.54 \pm 3.49$} \\
DR & ${-1.92}$ & ${-1.61}$ & ${-2.05}$ & ${-2.49}$ & ${-1.78}$ & ${-2.28 \pm 0.04}$ & \textcolor{red}{$-1.22 \pm 0.08$} & \textcolor{blue}{$-1.53 \pm 0.08$} & ${-2.21 \pm 0.09}$ & ${-1.74 \pm 0.19}$ \\
\bottomrule

\end{tabular}
\caption{Outcome sub rate $\in[0.4, \infty]$ low std}
\label{table: Outcome reward-subgroup-test_rate_0.4-_low_std}
\end{table}

%% file: rl_tables/sub/MIMIC3SepsisOutcomeEnv-all_train-sub-test_rate_0.4-_high_std-table.tex
\begin{table}[h]
\centering
\tablestyle{4pt}{1.0}
\tiny
\begin{tabular}{lcccccccccc}
\toprule
metric & alt & max & min & random & weight & imitation & dqn & cql & iql & bcq \\
\midrule
$\text{RMSE}_{\text{IV}}$ & ${1131.35}$ & ${1108.49}$ & ${1131.35}$ & ${1090.9}$ & ${1100.26}$ & \textcolor{blue}{$952.83 \pm 9.14$} & ${1106.01 \pm 22.44}$ & ${1032.13 \pm 14.06}$ & \textcolor{red}{$950.08 \pm 7.98$} & ${1085.45 \pm 4.58}$ \\
$\text{RMSE}_{\text{vaso}}$ & ${0.21}$ & ${0.86}$ & ${0.21}$ & ${0.45}$ & ${0.54}$ & ${0.21}$ & ${0.21}$ & \textcolor{blue}{$0.2 \pm 0.01$} & ${0.21 \pm 0.01}$ & \textcolor{red}{$0.18 \pm 0.01$} \\
$\text{WIS}$ & ${90.0}$ & ${3.99}$ & ${89.54}$ & ${-68.14}$ & ${-83.06}$ & ${92.27 \pm 0.01}$ & ${90.65 \pm 2.48}$ & \textcolor{red}{$95.1$} & \textcolor{blue}{$92.31 \pm 0.29$} & ${90.27 \pm 0.91}$ \\
$\text{WIS}_{b}$ & ${81.32}$ & ${-0.46}$ & ${70.64}$ & ${-20.11}$ & ${-50.82}$ & ${91.54 \pm 0.71}$ & ${80.52 \pm 7.96}$ & \textcolor{blue}{$91.62 \pm 2.23$} & \textcolor{red}{$91.83 \pm 0.43$} & ${88.1 \pm 1.72}$ \\
$\text{WIS}_{t}$ & ${90.0}$ & ${3.94}$ & ${89.54}$ & ${-85.19}$ & ${-83.06}$ & ${92.07 \pm 0.41}$ & ${90.65 \pm 2.48}$ & \textcolor{red}{$95.1$} & \textcolor{blue}{$92.63 \pm 1.01$} & ${90.27 \pm 0.91}$ \\
$\text{WIS}_{bt}$ & ${72.37}$ & ${-3.14}$ & ${83.27}$ & ${-47.0}$ & ${-68.58}$ & ${90.96 \pm 1.05}$ & ${74.47 \pm 11.5}$ & \textcolor{red}{$92.32 \pm 1.75$} & \textcolor{blue}{$91.54 \pm 0.91$} & ${85.09 \pm 2.4}$ \\
DR & ${-2.66}$ & ${-2.05}$ & ${-2.82}$ & ${-2.05}$ & \textcolor{blue}{$-1.99$} & ${-2.05}$ & \textcolor{red}{$-1.93 \pm 0.17$} & ${-2.25 \pm 0.4}$ & ${-2.25 \pm 0.25}$ & ${-2.65 \pm 0.15}$ \\
\bottomrule

\end{tabular}
\caption{Outcome sub rate $\in[0.4, \infty]$ high std}
\label{table: Outcome reward-subgroup-test_rate_0.4-_high_std}
\end{table}

%% file: rl_tables/sub/MIMIC3SepsisSOFAEnv-all_train-sub-test_rate_--0.4_low_std-table.tex
\begin{table}[h]
\centering
\tablestyle{4pt}{1.0}
\tiny
\begin{tabular}{lcccccccccc}
\toprule
metric & alt & max & min & random & weight & imitation & dqn & cql & iql & bcq \\
\midrule
$\text{RMSE}_{\text{IV}}$ & ${880.57}$ & ${929.48}$ & ${825.16}$ & ${821.09}$ & ${828.29}$ & \textcolor{red}{$656.94 \pm 10.21$} & ${804.13 \pm 10.32}$ & ${722.12 \pm 41.74}$ & \textcolor{blue}{$720.14 \pm 13.87$} & ${803.12 \pm 7.26}$ \\
$\text{RMSE}_{\text{vaso}}$ & ${0.71}$ & ${0.87}$ & ${0.2}$ & ${0.44}$ & ${0.55}$ & \textcolor{red}{$0.18 \pm 0.01$} & ${0.21}$ & ${0.2}$ & \textcolor{blue}{$0.19 \pm 0.01$} & \textcolor{blue}{$0.19 \pm 0.01$} \\
$\text{WIS}$ & ${15.08}$ & \textcolor{red}{$16.43$} & ${14.47}$ & \textcolor{red}{$16.43$} & \textcolor{red}{$16.43$} & \textcolor{red}{$16.43$} & \textcolor{red}{$16.43$} & \textcolor{blue}{$16.19 \pm 0.5$} & \textcolor{red}{$16.43$} & ${14.59 \pm 0.15}$ \\
$\text{WIS}_{b}$ & ${14.88}$ & ${12.73}$ & ${14.52}$ & ${11.31}$ & \textcolor{red}{$16.4$} & \textcolor{blue}{$15.21 \pm 0.24$} & ${15.01 \pm 0.49}$ & ${14.87 \pm 0.44}$ & ${14.97 \pm 0.53}$ & ${14.65 \pm 0.06}$ \\
$\text{WIS}_{t}$ & ${15.08}$ & \textcolor{red}{$16.43$} & ${14.47}$ & \textcolor{red}{$16.43$} & \textcolor{red}{$16.43$} & \textcolor{red}{$16.43$} & \textcolor{red}{$16.43$} & \textcolor{blue}{$16.09 \pm 0.51$} & \textcolor{red}{$16.43$} & ${14.59 \pm 0.15}$ \\
$\text{WIS}_{bt}$ & ${14.82}$ & ${12.12}$ & ${14.63}$ & ${13.15}$ & \textcolor{red}{$16.75$} & ${15.07 \pm 0.17}$ & \textcolor{blue}{$15.1 \pm 0.42$} & ${14.91 \pm 0.42}$ & ${14.91 \pm 0.14}$ & ${14.66 \pm 0.03}$ \\
DR & ${-0.22}$ & ${-0.22}$ & ${-0.32}$ & ${-0.22}$ & ${-0.22}$ & \textcolor{red}{$-0.01 \pm 0.02$} & ${-0.22}$ & ${-0.28 \pm 0.09}$ & \textcolor{blue}{$-0.02 \pm 0.03$} & ${-0.3 \pm 0.14}$ \\
\bottomrule

\end{tabular}
\caption{SOFA sub rate  $\in[-\infty, -0.4]$ low std}
\label{table: SOFA reward-subgroup-test_rate_--0.4_low_std}
\end{table}

%% file: rl_tables/sub/MIMIC3SepsisSOFAEnv-all_train-sub-test_rate_--0.4_high_std-table.tex
\begin{table}[h]
\centering
\tablestyle{4pt}{1.0}
\tiny
\begin{tabular}{lcccccccccc}
\toprule
metric & alt & max & min & random & weight & imitation & dqn & cql & iql & bcq \\
\midrule
$\text{RMSE}_{\text{IV}}$ & ${788.91}$ & ${880.8}$ & ${756.41}$ & ${774.78}$ & ${749.58}$ & \textcolor{red}{$608.58 \pm 16.25$} & ${734.77 \pm 21.49}$ & ${661.93 \pm 14.83}$ & \textcolor{blue}{$648.47 \pm 8.58$} & ${729.71 \pm 3.49}$ \\
$\text{RMSE}_{\text{vaso}}$ & ${0.54}$ & ${0.85}$ & ${0.27}$ & ${0.47}$ & ${0.56}$ & \textcolor{red}{$0.24 \pm 0.02$} & ${0.27 \pm 0.01}$ & \textcolor{blue}{$0.26$} & ${0.27 \pm 0.01}$ & \textcolor{blue}{$0.26 \pm 0.01$} \\
$\text{WIS}$ & ${1.45}$ & ${4.98}$ & ${-4.14}$ & ${14.24}$ & ${-12.51}$ & ${14.24}$ & \textcolor{red}{$14.59 \pm 0.87$} & \textcolor{blue}{$14.52 \pm 0.81$} & ${14.24 \pm 0.01}$ & ${0.69 \pm 6.55}$ \\
$\text{WIS}_{b}$ & ${0.09}$ & ${4.73}$ & ${-2.51}$ & ${7.71}$ & ${-4.69}$ & ${12.38 \pm 1.57}$ & \textcolor{blue}{$13.64 \pm 0.28$} & \textcolor{red}{$14.08 \pm 0.69$} & ${13.4 \pm 0.67}$ & ${2.25 \pm 4.11}$ \\
$\text{WIS}_{t}$ & ${1.45}$ & ${4.98}$ & ${-4.14}$ & ${14.24}$ & ${-12.51}$ & ${14.24}$ & \textcolor{red}{$14.59 \pm 0.87$} & \textcolor{blue}{$14.52 \pm 0.81$} & ${14.24 \pm 0.01}$ & ${0.69 \pm 6.55}$ \\
$\text{WIS}_{bt}$ & ${2.14}$ & ${4.55}$ & ${-2.64}$ & ${8.67}$ & ${-0.84}$ & ${12.62 \pm 2.77}$ & ${13.59 \pm 0.31}$ & \textcolor{red}{$14.1 \pm 0.39$} & \textcolor{blue}{$13.67 \pm 0.43$} & ${2.27 \pm 2.96}$ \\
DR & ${-0.09}$ & \textcolor{red}{$0.0$} & ${-0.35}$ & \textcolor{red}{$0.0$} & ${-0.18}$ & \textcolor{red}{$-0.0 \pm 0.01$} & ${-0.04 \pm 0.08}$ & \textcolor{red}{$0.0$} & \textcolor{blue}{$-0.01 \pm 0.02$} & ${-0.21 \pm 0.08}$ \\
\bottomrule

\end{tabular}
\caption{SOFA sub rate  $\in[-\infty, -0.4]$ high std}
\label{table: SOFA reward-subgroup-test_rate_--0.4_high_std}
\end{table}

%% file: rl_tables/sub/MIMIC3SepsisSOFAEnv-all_train-sub-test_rate_-0.4--0.15_low_std-table.tex
\begin{table}[h]
\centering
\tablestyle{4pt}{1.0}
\tiny
\begin{tabular}{lcccccccccc}
\toprule
metric & alt & max & min & random & weight & imitation & dqn & cql & iql & bcq \\
\midrule
$\text{RMSE}_{\text{IV}}$ & ${732.68}$ & ${849.88}$ & ${620.73}$ & ${649.77}$ & ${620.52}$ & \textcolor{red}{$513.03 \pm 17.23$} & ${604.3 \pm 14.72}$ & \textcolor{blue}{$556.43 \pm 16.69$} & ${556.7 \pm 6.58}$ & ${605.62 \pm 1.0}$ \\
$\text{RMSE}_{\text{vaso}}$ & ${0.6}$ & ${0.87}$ & ${0.2}$ & ${0.44}$ & ${0.55}$ & \textcolor{red}{$0.18 \pm 0.01$} & ${0.2}$ & ${0.2}$ & ${0.2 \pm 0.01}$ & \textcolor{blue}{$0.19$} \\
$\text{WIS}$ & ${13.56}$ & ${-10.69}$ & ${13.55}$ & ${-15.24}$ & ${-12.41}$ & ${-6.77 \pm 7.5}$ & ${13.1 \pm 0.27}$ & \textcolor{red}{$13.84 \pm 0.06$} & ${8.73 \pm 5.08}$ & \textcolor{blue}{$13.63 \pm 0.18$} \\
$\text{WIS}_{b}$ & \textcolor{blue}{$13.57$} & ${-7.14}$ & ${13.56}$ & ${-4.9}$ & ${-8.29}$ & ${-2.04 \pm 4.06}$ & ${13.28 \pm 0.15}$ & \textcolor{red}{$13.87 \pm 0.11$} & ${9.39 \pm 4.08}$ & \textcolor{blue}{$13.57 \pm 0.14$} \\
$\text{WIS}_{t}$ & ${13.56}$ & ${-10.69}$ & ${13.55}$ & ${-15.24}$ & ${-12.41}$ & ${-6.77 \pm 7.5}$ & ${13.1 \pm 0.27}$ & \textcolor{red}{$13.84 \pm 0.06$} & ${8.73 \pm 5.08}$ & \textcolor{blue}{$13.63 \pm 0.18$} \\
$\text{WIS}_{bt}$ & ${13.49}$ & ${-9.88}$ & ${13.52}$ & ${-1.86}$ & ${-7.61}$ & ${-1.62 \pm 3.65}$ & ${13.24 \pm 0.14}$ & \textcolor{red}{$13.85 \pm 0.09$} & ${9.11 \pm 4.52}$ & \textcolor{blue}{$13.55 \pm 0.12$} \\
DR & ${-0.27}$ & ${-0.07}$ & ${-0.7}$ & ${-0.07}$ & ${-0.1}$ & \textcolor{red}{$-0.04 \pm 0.02$} & ${-0.09 \pm 0.04}$ & ${-0.18 \pm 0.1}$ & \textcolor{blue}{$-0.05 \pm 0.03$} & ${-0.69 \pm 0.02}$ \\
\bottomrule

\end{tabular}
\caption{SOFA sub rate  $\in[-0.4, -0.15]$ low std}
\label{table: SOFA reward-subgroup-test_rate_-0.4--0.15_low_std}
\end{table}

%% file: rl_tables/sub/MIMIC3SepsisSOFAEnv-all_train-sub-test_rate_-0.4--0.15_high_std-table.tex
\begin{table}[h]
\centering
\tablestyle{4pt}{1.0}
\tiny
\begin{tabular}{lcccccccccc}
\toprule
metric & alt & max & min & random & weight & imitation & dqn & cql & iql & bcq \\
\midrule
$\text{RMSE}_{\text{IV}}$ & ${820.25}$ & ${887.4}$ & ${770.94}$ & ${771.37}$ & ${767.21}$ & \textcolor{red}{$638.76 \pm 14.54$} & ${746.47 \pm 26.15}$ & ${685.1 \pm 16.71}$ & \textcolor{blue}{$675.19 \pm 6.55$} & ${747.52 \pm 2.79}$ \\
$\text{RMSE}_{\text{vaso}}$ & ${0.82}$ & ${1.03}$ & ${0.65}$ & ${0.74}$ & ${0.81}$ & \textcolor{red}{$0.63$} & \textcolor{blue}{$0.64$} & ${0.65}$ & \textcolor{blue}{$0.64 \pm 0.01$} & \textcolor{blue}{$0.64$} \\
$\text{WIS}$ & ${12.0}$ & ${11.61}$ & ${12.0}$ & \textcolor{red}{$17.13$} & ${12.88}$ & \textcolor{blue}{$17.09 \pm 0.05$} & ${14.89 \pm 1.23}$ & ${12.46 \pm 1.55}$ & ${14.58 \pm 2.22}$ & ${12.0}$ \\
$\text{WIS}_{b}$ & ${11.46}$ & ${10.32}$ & ${11.0}$ & \textcolor{blue}{$15.43$} & ${13.01}$ & \textcolor{red}{$15.87 \pm 0.29$} & ${14.64 \pm 1.03}$ & ${12.22 \pm 1.99}$ & ${14.23 \pm 1.4}$ & ${11.36 \pm 0.25}$ \\
$\text{WIS}_{t}$ & ${12.0}$ & ${11.61}$ & ${12.0}$ & \textcolor{red}{$17.13$} & ${12.88}$ & \textcolor{blue}{$17.09 \pm 0.05$} & ${14.89 \pm 1.23}$ & ${12.46 \pm 1.55}$ & ${14.58 \pm 2.22}$ & ${12.0}$ \\
$\text{WIS}_{bt}$ & ${11.38}$ & ${9.96}$ & ${10.53}$ & \textcolor{red}{$15.78$} & ${12.97}$ & \textcolor{blue}{$15.74 \pm 0.26$} & ${14.68 \pm 0.9}$ & ${12.17 \pm 1.96}$ & ${14.15 \pm 1.45}$ & ${11.35 \pm 0.28}$ \\
DR & ${-0.51}$ & ${-0.39}$ & ${-0.68}$ & ${-0.31}$ & ${-0.27}$ & ${-0.3 \pm 0.01}$ & \textcolor{red}{$-0.17 \pm 0.06$} & \textcolor{blue}{$-0.21 \pm 0.1$} & ${-0.3 \pm 0.02}$ & ${-0.68}$ \\
\bottomrule

\end{tabular}
\caption{SOFA sub rate  $\in[-0.4, -0.15]$ high std}
\label{table: SOFA reward-subgroup-test_rate_-0.4--0.15_high_std}
\end{table}

%% file: rl_tables/sub/MIMIC3SepsisSOFAEnv-all_train-sub-test_rate_-0.15-0_low_std-table.tex
\begin{table}[h]
\centering
\tablestyle{4pt}{1.0}
\tiny
\begin{tabular}{lcccccccccc}
\toprule
metric & alt & max & min & random & weight & imitation & dqn & cql & iql & bcq \\
\midrule
$\text{RMSE}_{\text{IV}}$ & ${688.63}$ & ${833.0}$ & ${457.4}$ & ${532.53}$ & ${486.77}$ & \textcolor{red}{$390.13 \pm 15.74$} & ${451.62 \pm 3.96}$ & \textcolor{blue}{$416.64 \pm 11.45$} & ${428.5 \pm 8.67}$ & ${445.05 \pm 1.22}$ \\
$\text{RMSE}_{\text{vaso}}$ & ${0.67}$ & ${0.89}$ & \textcolor{blue}{$0.26$} & ${0.48}$ & ${0.58}$ & \textcolor{red}{$0.25$} & ${0.27}$ & \textcolor{blue}{$0.26$} & \textcolor{blue}{$0.26$} & \textcolor{blue}{$0.26$} \\
$\text{WIS}$ & \textcolor{blue}{$13.34$} & ${-9.01}$ & \textcolor{blue}{$13.34$} & ${-14.1}$ & ${-14.12}$ & ${9.74 \pm 0.45}$ & ${12.86 \pm 0.97}$ & \textcolor{red}{$13.87 \pm 1.76$} & ${11.65 \pm 1.61}$ & \textcolor{blue}{$13.34$} \\
$\text{WIS}_{b}$ & ${7.17}$ & ${-2.14}$ & ${11.83}$ & ${-9.45}$ & ${-10.28}$ & ${10.83 \pm 0.49}$ & ${12.21 \pm 0.65}$ & \textcolor{red}{$12.72 \pm 2.69$} & ${11.65 \pm 1.21}$ & \textcolor{blue}{$12.43 \pm 1.26$} \\
$\text{WIS}_{t}$ & \textcolor{blue}{$13.34$} & ${-9.01}$ & \textcolor{blue}{$13.34$} & ${-14.1}$ & ${-14.12}$ & ${11.08 \pm 1.27}$ & ${12.86 \pm 0.97}$ & \textcolor{red}{$13.87 \pm 1.76$} & ${11.9 \pm 1.79}$ & \textcolor{blue}{$13.34$} \\
$\text{WIS}_{bt}$ & ${7.84}$ & ${-2.32}$ & ${10.89}$ & ${-10.12}$ & ${-10.21}$ & ${11.53 \pm 0.44}$ & ${12.24 \pm 0.63}$ & \textcolor{red}{$12.57 \pm 3.02$} & ${11.92 \pm 1.45}$ & \textcolor{blue}{$12.5 \pm 1.33$} \\
DR & ${-0.46}$ & \textcolor{blue}{$-0.14$} & ${-0.72}$ & ${-0.17}$ & ${-0.15}$ & \textcolor{blue}{$-0.14 \pm 0.02$} & \textcolor{red}{$-0.13 \pm 0.04$} & ${-0.35 \pm 0.19}$ & ${-0.15 \pm 0.03}$ & ${-0.72}$ \\
\bottomrule

\end{tabular}
\caption{SOFA sub rate  $\in[-0.15, 0]$ low std}
\label{table: SOFA reward-subgroup-test_rate_-0.15-0_low_std}
\end{table}

%% file: rl_tables/sub/MIMIC3SepsisSOFAEnv-all_train-sub-test_rate_-0.15-0_high_std-table.tex
\begin{table}[h]
\centering
\tablestyle{4pt}{1.0}
\tiny
\begin{tabular}{lcccccccccc}
\toprule
metric & alt & max & min & random & weight & imitation & dqn & cql & iql & bcq \\
\midrule
$\text{RMSE}_{\text{IV}}$ & ${765.09}$ & ${867.74}$ & ${669.04}$ & ${696.63}$ & ${662.95}$ & \textcolor{red}{$551.05 \pm 9.71$} & ${650.28 \pm 19.61}$ & ${605.84 \pm 15.26}$ & \textcolor{blue}{$597.7 \pm 6.59$} & ${652.23 \pm 1.63}$ \\
$\text{RMSE}_{\text{vaso}}$ & ${0.66}$ & ${0.87}$ & ${0.28}$ & ${0.48}$ & ${0.57}$ & \textcolor{red}{$0.25$} & ${0.28 \pm 0.01}$ & ${0.28}$ & \textcolor{blue}{$0.27 \pm 0.01$} & \textcolor{blue}{$0.27$} \\
$\text{WIS}$ & ${12.06}$ & ${-21.32}$ & ${12.06}$ & ${10.24}$ & \textcolor{red}{$13.4$} & ${10.24}$ & ${12.71 \pm 1.65}$ & \textcolor{blue}{$13.03 \pm 0.7$} & ${10.24}$ & ${12.06}$ \\
$\text{WIS}_{b}$ & ${5.79}$ & ${-13.27}$ & ${11.83}$ & ${8.56}$ & ${10.92}$ & ${7.87 \pm 0.48}$ & \textcolor{blue}{$12.58 \pm 0.82$} & \textcolor{red}{$12.66 \pm 0.81$} & ${3.81 \pm 1.28}$ & ${11.89 \pm 0.19}$ \\
$\text{WIS}_{t}$ & ${12.06}$ & ${-21.32}$ & ${12.06}$ & ${10.24}$ & \textcolor{red}{$13.4$} & ${10.24}$ & ${12.71 \pm 1.65}$ & \textcolor{blue}{$13.03 \pm 0.7$} & ${10.24}$ & ${12.06}$ \\
$\text{WIS}_{bt}$ & ${5.98}$ & ${-11.51}$ & ${11.98}$ & ${8.09}$ & ${10.1}$ & ${8.19 \pm 1.45}$ & \textcolor{blue}{$12.63 \pm 1.02$} & \textcolor{red}{$12.75 \pm 0.95$} & ${3.4 \pm 1.44}$ & ${11.99 \pm 0.09}$ \\
DR & ${-0.4}$ & \textcolor{blue}{$-0.19$} & ${-0.61}$ & ${-0.33}$ & ${-0.31}$ & ${-0.34 \pm 0.02}$ & \textcolor{red}{$-0.11 \pm 0.03$} & ${-0.29 \pm 0.17}$ & ${-0.38 \pm 0.04}$ & ${-0.58 \pm 0.02}$ \\
\bottomrule

\end{tabular}
\caption{SOFA sub rate  $\in[-0.15, 0]$ high std}
\label{table: SOFA reward-subgroup-test_rate_-0.15-0_high_std}
\end{table}

%% file: rl_tables/sub/MIMIC3SepsisSOFAEnv-all_train-sub-test_rate_0-0.15_low_std-table.tex
\begin{table}[h]
\centering
\tablestyle{4pt}{1.0}
\tiny
\begin{tabular}{lcccccccccc}
\toprule
metric & alt & max & min & random & weight & imitation & dqn & cql & iql & bcq \\
\midrule
$\text{RMSE}_{\text{IV}}$ & ${756.44}$ & ${875.08}$ & ${576.23}$ & ${624.74}$ & ${597.18}$ & \textcolor{red}{$524.0 \pm 20.68$} & ${567.79 \pm 7.77}$ & \textcolor{blue}{$549.11 \pm 17.14$} & ${555.3 \pm 4.66}$ & ${563.58 \pm 0.81}$ \\
$\text{RMSE}_{\text{vaso}}$ & ${0.6}$ & ${0.87}$ & \textcolor{blue}{$0.13$} & ${0.44}$ & ${0.54}$ & \textcolor{blue}{$0.13$} & \textcolor{blue}{$0.13 \pm 0.01$} & \textcolor{blue}{$0.13$} & ${0.15 \pm 0.01}$ & \textcolor{red}{$0.12$} \\
$\text{WIS}$ & ${12.98}$ & ${-7.35}$ & ${12.98}$ & ${12.14}$ & ${12.14}$ & ${12.49 \pm 0.47}$ & \textcolor{blue}{$13.51 \pm 1.17$} & \textcolor{red}{$13.61 \pm 1.22$} & ${12.93 \pm 0.84}$ & ${12.98}$ \\
$\text{WIS}_{b}$ & \textcolor{red}{$12.67$} & ${0.77}$ & ${11.6}$ & ${11.92}$ & ${11.57}$ & ${12.06 \pm 0.51}$ & ${12.29 \pm 0.57}$ & ${12.2 \pm 0.25}$ & \textcolor{blue}{$12.57 \pm 0.75$} & ${12.09 \pm 0.36}$ \\
$\text{WIS}_{t}$ & ${12.98}$ & ${-7.35}$ & ${12.98}$ & ${12.14}$ & ${12.14}$ & ${12.58 \pm 0.51}$ & \textcolor{blue}{$13.51 \pm 1.17$} & \textcolor{red}{$13.61 \pm 1.22$} & ${12.95 \pm 0.81}$ & ${12.98}$ \\
$\text{WIS}_{bt}$ & \textcolor{blue}{$12.39$} & ${0.55}$ & ${11.54}$ & ${11.45}$ & ${11.51}$ & ${12.24 \pm 0.53}$ & ${12.29 \pm 0.54}$ & ${12.3 \pm 0.43}$ & \textcolor{red}{$12.6 \pm 0.75$} & ${12.11 \pm 0.33}$ \\
DR & ${-0.47}$ & \textcolor{red}{$-0.19$} & ${-0.86}$ & ${-0.64}$ & ${-0.68}$ & ${-0.35 \pm 0.05}$ & \textcolor{blue}{$-0.26 \pm 0.16$} & ${-0.43 \pm 0.12}$ & ${-0.37 \pm 0.04}$ & ${-0.78 \pm 0.1}$ \\
\bottomrule

\end{tabular}
\caption{SOFA sub rate $\in[0, 0.15]$ low std}
\label{table: SOFA reward-subgroup-test_rate_0-0.15_low_std}
\end{table}

%% file: rl_tables/sub/MIMIC3SepsisSOFAEnv-all_train-sub-test_rate_0-0.15_high_std-table.tex
\begin{table}[h]
\centering
\tablestyle{4pt}{1.0}
\tiny
\begin{tabular}{lcccccccccc}
\toprule
metric & alt & max & min & random & weight & imitation & dqn & cql & iql & bcq \\
\midrule
$\text{RMSE}_{\text{IV}}$ & ${760.98}$ & ${834.61}$ & ${628.74}$ & ${637.52}$ & ${628.39}$ & \textcolor{red}{$512.98 \pm 13.95$} & ${608.05 \pm 19.71}$ & \textcolor{blue}{$559.49 \pm 24.66$} & ${559.66 \pm 7.59}$ & ${609.53 \pm 2.49}$ \\
$\text{RMSE}_{\text{vaso}}$ & ${0.7}$ & ${0.88}$ & ${0.3}$ & ${0.49}$ & ${0.58}$ & \textcolor{red}{$0.28 \pm 0.01$} & ${0.3}$ & ${0.3}$ & \textcolor{blue}{$0.29 \pm 0.01$} & \textcolor{blue}{$0.29$} \\
$\text{WIS}$ & ${9.59}$ & ${-14.02}$ & ${9.59}$ & ${11.62}$ & \textcolor{red}{$13.11$} & ${11.62}$ & ${11.46 \pm 1.78}$ & \textcolor{blue}{$12.59 \pm 0.87$} & ${11.6 \pm 0.06}$ & ${10.01 \pm 0.93}$ \\
$\text{WIS}_{b}$ & ${10.52}$ & ${-8.8}$ & ${10.56}$ & ${6.07}$ & \textcolor{blue}{$12.14$} & ${9.9 \pm 0.98}$ & ${12.07 \pm 1.31}$ & \textcolor{red}{$12.33 \pm 0.64$} & ${9.34 \pm 0.92}$ & ${10.58 \pm 0.44}$ \\
$\text{WIS}_{t}$ & ${9.59}$ & ${-14.02}$ & ${9.59}$ & ${11.62}$ & \textcolor{red}{$13.12$} & ${11.62}$ & ${11.46 \pm 1.77}$ & \textcolor{blue}{$12.38 \pm 0.52$} & ${11.57 \pm 0.12}$ & ${10.01 \pm 0.93}$ \\
$\text{WIS}_{bt}$ & ${10.7}$ & ${-10.69}$ & ${10.54}$ & ${7.03}$ & ${11.17}$ & ${10.31 \pm 0.33}$ & \textcolor{blue}{$11.89 \pm 1.49$} & \textcolor{red}{$12.13 \pm 0.35$} & ${9.1 \pm 1.35}$ & ${10.56 \pm 0.45}$ \\
DR & ${-0.45}$ & \textcolor{red}{$-0.33$} & ${-0.82}$ & ${-0.42}$ & ${-0.42}$ & \textcolor{blue}{$-0.36 \pm 0.03$} & ${-0.49 \pm 0.2}$ & ${-0.5 \pm 0.24}$ & ${-0.37 \pm 0.07}$ & ${-0.82}$ \\
\bottomrule

\end{tabular}
\caption{SOFA sub rate $\in[0, 0.15]$ high std}
\label{table: SOFA reward-subgroup-test_rate_0-0.15_high_std}
\end{table}

%% file: rl_tables/sub/MIMIC3SepsisSOFAEnv-all_train-sub-test_rate_0.15-0.4_low_std-table.tex
\begin{table}[h]
\centering
\tablestyle{4pt}{1.0}
\tiny
\begin{tabular}{lcccccccccc}
\toprule
metric & alt & max & min & random & weight & imitation & dqn & cql & iql & bcq \\
\midrule
$\text{RMSE}_{\text{IV}}$ & ${736.69}$ & ${853.28}$ & ${581.25}$ & ${616.67}$ & ${592.53}$ & \textcolor{red}{$480.48 \pm 15.3$} & ${568.16 \pm 11.04}$ & ${520.1 \pm 15.18}$ & \textcolor{blue}{$519.96 \pm 17.85$} & ${566.24 \pm 1.92}$ \\
$\text{RMSE}_{\text{vaso}}$ & ${0.6}$ & ${0.87}$ & ${0.24}$ & ${0.46}$ & ${0.57}$ & \textcolor{red}{$0.22 \pm 0.01$} & ${0.24}$ & ${0.24}$ & \textcolor{blue}{$0.23 \pm 0.01$} & \textcolor{blue}{$0.23$} \\
$\text{WIS}$ & ${-16.39}$ & ${-16.4}$ & ${14.4}$ & ${-17.91}$ & ${-14.87}$ & ${11.49 \pm 0.63}$ & \textcolor{blue}{$15.15 \pm 0.28$} & \textcolor{red}{$15.27 \pm 0.02$} & ${12.65 \pm 0.85}$ & ${14.41}$ \\
$\text{WIS}_{b}$ & ${-5.02}$ & ${-5.66}$ & ${14.2}$ & ${-15.01}$ & ${-14.46}$ & ${11.46 \pm 0.81}$ & \textcolor{blue}{$14.68 \pm 0.33$} & \textcolor{red}{$14.76 \pm 0.04$} & ${11.51 \pm 1.8}$ & ${13.99 \pm 0.06}$ \\
$\text{WIS}_{t}$ & ${-16.39}$ & ${-16.4}$ & ${14.4}$ & ${-17.91}$ & ${-14.87}$ & ${11.49 \pm 0.63}$ & \textcolor{blue}{$15.15 \pm 0.28$} & \textcolor{red}{$15.27 \pm 0.02$} & ${12.65 \pm 0.85}$ & ${14.41}$ \\
$\text{WIS}_{bt}$ & ${-3.9}$ & ${-6.52}$ & ${14.14}$ & ${-15.0}$ & ${-13.38}$ & ${11.45 \pm 0.83}$ & \textcolor{blue}{$14.65 \pm 0.36$} & \textcolor{red}{$14.76 \pm 0.14$} & ${11.62 \pm 1.7}$ & ${14.01 \pm 0.12}$ \\
DR & ${-0.49}$ & \textcolor{blue}{$-0.32$} & ${-0.65}$ & ${-0.61}$ & ${-0.46}$ & ${-0.47 \pm 0.01}$ & \textcolor{red}{$-0.26 \pm 0.02$} & ${-0.37 \pm 0.15}$ & ${-0.5 \pm 0.04}$ & ${-0.65}$ \\
\bottomrule

\end{tabular}
\caption{SOFA sub rate $\in[0.15, 0.4]$ low std}
\label{table: SOFA reward-subgroup-test_rate_0.15-0.4_low_std}
\end{table}

%% file: rl_tables/sub/MIMIC3SepsisSOFAEnv-all_train-sub-test_rate_0.15-0.4_high_std-table.tex
\begin{table}[h]
\centering
\tablestyle{4pt}{1.0}
\tiny
\begin{tabular}{lcccccccccc}
\toprule
metric & alt & max & min & random & weight & imitation & dqn & cql & iql & bcq \\
\midrule
$\text{RMSE}_{\text{IV}}$ & ${782.93}$ & ${832.02}$ & ${699.16}$ & ${694.47}$ & ${686.84}$ & \textcolor{red}{$573.43 \pm 13.11$} & ${670.46 \pm 30.69}$ & ${635.01 \pm 10.67}$ & \textcolor{blue}{$623.91 \pm 15.84$} & ${680.34 \pm 2.92}$ \\
$\text{RMSE}_{\text{vaso}}$ & ${0.62}$ & ${0.85}$ & ${0.25}$ & ${0.45}$ & ${0.54}$ & \textcolor{red}{$0.22 \pm 0.01$} & ${0.25}$ & ${0.24}$ & ${0.24 \pm 0.01}$ & \textcolor{blue}{$0.23$} \\
$\text{WIS}$ & \textcolor{red}{$15.85$} & ${-13.08}$ & ${7.84}$ & ${-5.98}$ & ${-13.33}$ & \textcolor{blue}{$15.43 \pm 0.55$} & ${13.43 \pm 1.56}$ & ${15.28 \pm 1.25}$ & ${14.15 \pm 1.2}$ & ${7.84}$ \\
$\text{WIS}_{b}$ & ${11.61}$ & ${-6.02}$ & ${7.13}$ & ${-4.04}$ & ${-12.1}$ & \textcolor{blue}{$13.74 \pm 1.13$} & ${13.43 \pm 0.87}$ & \textcolor{red}{$14.53 \pm 0.75$} & ${11.91 \pm 1.81}$ & ${7.5 \pm 0.79}$ \\
$\text{WIS}_{t}$ & \textcolor{red}{$15.85$} & ${-13.08}$ & ${7.84}$ & ${-5.98}$ & ${-13.33}$ & \textcolor{blue}{$15.43 \pm 0.55$} & ${13.43 \pm 1.56}$ & ${15.28 \pm 1.25}$ & ${14.15 \pm 1.2}$ & ${7.84}$ \\
$\text{WIS}_{bt}$ & ${10.6}$ & ${-5.53}$ & ${7.35}$ & ${-4.23}$ & ${-12.44}$ & \textcolor{blue}{$13.61 \pm 1.0$} & ${13.5 \pm 1.05}$ & \textcolor{red}{$14.48 \pm 0.78$} & ${11.67 \pm 1.9}$ & ${7.42 \pm 0.21}$ \\
DR & ${-0.94}$ & \textcolor{blue}{$-0.4$} & ${-0.94}$ & ${-0.53}$ & ${-0.56}$ & ${-0.51 \pm 0.01}$ & \textcolor{red}{$-0.3 \pm 0.12$} & ${-0.53 \pm 0.14}$ & ${-0.52 \pm 0.04}$ & ${-0.85 \pm 0.09}$ \\
\bottomrule

\end{tabular}
\caption{SOFA sub rate $\in[0.15, 0.4]$ high std}
\label{table: SOFA reward-subgroup-test_rate_0.15-0.4_high_std}
\end{table}

%% file: rl_tables/sub/MIMIC3SepsisSOFAEnv-all_train-sub-test_rate_0.4-_low_std-table.tex
\begin{table}[h]
\centering
\tablestyle{4pt}{1.0}
\tiny
\begin{tabular}{lcccccccccc}
\toprule
metric & alt & max & min & random & weight & imitation & dqn & cql & iql & bcq \\
\midrule
$\text{RMSE}_{\text{IV}}$ & ${828.61}$ & ${828.61}$ & ${626.63}$ & ${642.26}$ & ${637.93}$ & \textcolor{red}{$516.24 \pm 10.8$} & ${601.13 \pm 29.95}$ & ${563.64 \pm 20.54}$ & \textcolor{blue}{$537.16 \pm 6.65$} & ${605.02 \pm 7.26}$ \\
$\text{RMSE}_{\text{vaso}}$ & ${0.87}$ & ${0.87}$ & ${0.4}$ & ${0.53}$ & ${0.62}$ & \textcolor{red}{$0.27 \pm 0.02$} & ${0.39 \pm 0.02}$ & ${0.39 \pm 0.02}$ & \textcolor{blue}{$0.34 \pm 0.01$} & ${0.38 \pm 0.01}$ \\
$\text{WIS}$ & \textcolor{red}{$14.88$} & ${-15.5}$ & ${14.54}$ & ${-14.88}$ & ${-22.55}$ & ${-6.23 \pm 8.93}$ & ${14.16 \pm 0.21}$ & \textcolor{blue}{$14.62 \pm 0.38$} & ${-7.97 \pm 9.23}$ & ${14.54}$ \\
$\text{WIS}_{b}$ & ${10.74}$ & ${-15.99}$ & ${12.7}$ & ${-14.89}$ & ${-18.7}$ & ${-2.08 \pm 4.19}$ & ${9.72 \pm 3.4}$ & \textcolor{blue}{$13.47 \pm 0.54$} & ${-4.49 \pm 1.9}$ & \textcolor{red}{$13.54 \pm 0.3$} \\
$\text{WIS}_{t}$ & \textcolor{red}{$14.88$} & ${-15.5}$ & ${14.54}$ & ${-14.88}$ & ${-22.55}$ & ${-6.23 \pm 8.93}$ & ${14.16 \pm 0.21}$ & \textcolor{blue}{$14.62 \pm 0.38$} & ${-7.97 \pm 9.23}$ & ${14.54}$ \\
$\text{WIS}_{bt}$ & ${10.76}$ & ${-15.92}$ & ${13.69}$ & ${-15.33}$ & ${-19.14}$ & ${-1.93 \pm 4.64}$ & ${9.96 \pm 4.61}$ & \textcolor{blue}{$13.87 \pm 0.44$} & ${-3.58 \pm 3.76}$ & \textcolor{red}{$14.0 \pm 0.3$} \\
DR & ${-1.1}$ & \textcolor{red}{$-0.87$} & ${-1.47}$ & ${-1.45}$ & \textcolor{blue}{$-0.88$} & ${-1.35 \pm 0.06}$ & ${-1.3 \pm 0.02}$ & ${-1.05 \pm 0.02}$ & ${-1.64 \pm 0.09}$ & ${-1.44 \pm 0.06}$ \\
\bottomrule

\end{tabular}
\caption{SOFA sub rate $\in[0.4, \infty]$ low std}
\label{table: SOFA reward-subgroup-test_rate_0.4-_low_std}
\end{table}

%% file: rl_tables/sub/MIMIC3SepsisSOFAEnv-all_train-sub-test_rate_0.4-_high_std-table.tex
\begin{table}[h]
\centering
\tablestyle{4pt}{1.0}
\tiny
\begin{tabular}{lcccccccccc}
\toprule
metric & alt & max & min & random & weight & imitation & dqn & cql & iql & bcq \\
\midrule
$\text{RMSE}_{\text{IV}}$ & ${1131.35}$ & ${1108.49}$ & ${1131.35}$ & ${1090.9}$ & ${1100.26}$ & \textcolor{red}{$928.45 \pm 15.0$} & \makecell[c]{1087.4\\ $\pm$ 30.34} & ${996.8 \pm 34.24}$ & \textcolor{blue}{$950.08 \pm 7.98$} & ${1086.09 \pm 7.96}$ \\
$\text{RMSE}_{\text{vaso}}$ & ${0.21}$ & ${0.86}$ & ${0.21}$ & ${0.45}$ & ${0.54}$ & ${0.21 \pm 0.01}$ & ${0.21}$ & \textcolor{blue}{$0.2$} & ${0.21 \pm 0.01}$ & \textcolor{red}{$0.18 \pm 0.01$} \\
$\text{WIS}$ & ${12.2}$ & ${0.64}$ & ${11.9}$ & ${-9.46}$ & ${-13.74}$ & ${13.48 \pm 0.01}$ & \textcolor{red}{$15.43 \pm 0.15$} & \textcolor{blue}{$13.78 \pm 1.78$} & ${13.61 \pm 0.29}$ & ${11.99 \pm 0.21}$ \\
$\text{WIS}_{b}$ & ${11.49}$ & ${-0.25}$ & ${9.89}$ & ${-2.63}$ & ${-8.99}$ & \textcolor{blue}{$13.13 \pm 0.24$} & ${12.56 \pm 3.55}$ & ${11.1 \pm 2.11}$ & \textcolor{red}{$13.31 \pm 0.4$} & ${12.32 \pm 0.5}$ \\
$\text{WIS}_{t}$ & ${12.2}$ & ${0.63}$ & ${11.9}$ & ${-11.9}$ & ${-13.74}$ & ${13.39 \pm 0.16}$ & \textcolor{red}{$15.43 \pm 0.15$} & \textcolor{blue}{$13.78 \pm 1.78$} & ${13.71 \pm 0.63}$ & ${11.99 \pm 0.21}$ \\
$\text{WIS}_{bt}$ & ${9.75}$ & ${-0.63}$ & ${11.76}$ & ${-6.56}$ & ${-11.61}$ & \textcolor{blue}{$13.06 \pm 0.32$} & ${12.09 \pm 4.65}$ & ${10.44 \pm 2.47}$ & \textcolor{red}{$13.42 \pm 0.59$} & ${11.83 \pm 0.43}$ \\
DR & ${-2.05}$ & ${-1.69}$ & ${-2.99}$ & ${-1.65}$ & \textcolor{blue}{$-1.63$} & \textcolor{red}{$-1.6 \pm 0.02$} & ${-1.72 \pm 0.29}$ & ${-1.64 \pm 0.34}$ & ${-1.68 \pm 0.09}$ & ${-2.78 \pm 0.12}$ \\
\bottomrule

\end{tabular}
\caption{SOFA sub rate $\in[0.4, \infty]$ high std}
\label{table: SOFA reward-subgroup-test_rate_0.4-_high_std}
\end{table}

%% file: rl_tables/sub/MIMIC3SepsisNEWS2Env-all_train-sub-test_rate_--0.4_low_std-table.tex
\begin{table}[h]
\centering
\tablestyle{4pt}{1.0}
\tiny
\begin{tabular}{lcccccccccc}
\toprule
metric & alt & max & min & random & weight & imitation & dqn & cql & iql & bcq \\
\midrule
$\text{RMSE}_{\text{IV}}$ & ${880.57}$ & ${929.48}$ & ${825.16}$ & ${821.09}$ & ${828.29}$ & \textcolor{blue}{$683.96 \pm 3.08$} & ${811.24 \pm 26.66}$ & \textcolor{red}{$653.9 \pm 6.87$} & ${706.65 \pm 13.36}$ & ${798.7 \pm 10.64}$ \\
$\text{RMSE}_{\text{vaso}}$ & ${0.71}$ & ${0.87}$ & ${0.2}$ & ${0.44}$ & ${0.55}$ & ${0.2}$ & ${0.39 \pm 0.09}$ & \textcolor{red}{$0.18 \pm 0.01$} & \textcolor{blue}{$0.19 \pm 0.01$} & \textcolor{blue}{$0.19 \pm 0.01$} \\
$\text{WIS}$ & \textcolor{red}{$-1.78$} & ${-3.78}$ & ${-3.04}$ & ${-3.78}$ & ${-3.78}$ & ${-3.78}$ & ${-3.32 \pm 1.03}$ & \textcolor{blue}{$-2.12 \pm 0.94$} & ${-2.8 \pm 0.75}$ & ${-2.92 \pm 0.27}$ \\
$\text{WIS}_{b}$ & \textcolor{red}{$-2.25$} & ${-3.78}$ & ${-2.93}$ & ${-3.23}$ & ${-3.59}$ & ${-3.14 \pm 0.09}$ & ${-2.99 \pm 0.18}$ & \textcolor{blue}{$-2.46 \pm 0.56$} & ${-2.84 \pm 0.35}$ & ${-2.85 \pm 0.08}$ \\
$\text{WIS}_{t}$ & \textcolor{red}{$-1.78$} & ${-3.78}$ & ${-3.04}$ & ${-3.78}$ & ${-3.78}$ & ${-3.78}$ & ${-3.32 \pm 1.03}$ & \textcolor{blue}{$-2.12 \pm 0.94$} & ${-2.8 \pm 0.75}$ & ${-2.92 \pm 0.27}$ \\
$\text{WIS}_{bt}$ & \textcolor{red}{$-2.14$} & ${-3.87}$ & ${-2.99}$ & ${-3.35}$ & ${-3.61}$ & ${-3.18 \pm 0.08}$ & ${-3.04 \pm 0.56}$ & \textcolor{blue}{$-2.36 \pm 0.66$} & ${-2.83 \pm 0.37}$ & ${-2.93 \pm 0.27}$ \\
DR & ${-0.49}$ & ${-0.46}$ & ${-1.08}$ & \textcolor{blue}{$-0.35$} & \textcolor{red}{$-0.29$} & ${-0.37}$ & \textcolor{red}{$-0.29 \pm 0.02$} & ${-0.62 \pm 0.12}$ & ${-0.5 \pm 0.04}$ & ${-0.95 \pm 0.07}$ \\
\bottomrule

\end{tabular}
\caption{NEWS2 sub rate  $\in[-\infty, -0.4]$ low std}
\label{table: NEWS2 reward-subgroup-test_rate_--0.4_low_std}
\end{table}

%% file: rl_tables/sub/MIMIC3SepsisNEWS2Env-all_train-sub-test_rate_-0.4--0.15_low_std-table.tex
\begin{table}[h]
\centering
\tablestyle{4pt}{1.0}
\tiny
\begin{tabular}{lcccccccccc}
\toprule
metric & alt & max & min & random & weight & imitation & dqn & cql & iql & bcq \\
\midrule
$\text{RMSE}_{\text{IV}}$ & ${732.68}$ & ${849.88}$ & ${620.73}$ & ${649.77}$ & ${620.52}$ & \textcolor{blue}{$528.34 \pm 16.69$} & ${615.53 \pm 13.8}$ & \textcolor{red}{$511.36 \pm 7.79$} & ${559.08 \pm 5.67}$ & ${604.78 \pm 7.09}$ \\
$\text{RMSE}_{\text{vaso}}$ & ${0.6}$ & ${0.87}$ & ${0.2}$ & ${0.44}$ & ${0.55}$ & ${0.2}$ & ${0.37 \pm 0.09}$ & \textcolor{red}{$0.18 \pm 0.01$} & ${0.21 \pm 0.01}$ & \textcolor{blue}{$0.19$} \\
$\text{WIS}$ & ${-5.08}$ & ${-5.06}$ & ${-5.12}$ & ${-4.02}$ & ${-4.08}$ & \textcolor{blue}{$-3.92 \pm 0.07$} & \textcolor{red}{$-2.11 \pm 1.17$} & ${-3.96 \pm 1.46}$ & ${-4.12 \pm 0.18}$ & ${-4.28}$ \\
$\text{WIS}_{b}$ & ${-4.77}$ & ${-4.08}$ & ${-4.81}$ & ${-4.11}$ & ${-4.3}$ & ${-3.7 \pm 0.25}$ & \textcolor{red}{$-2.42 \pm 0.54$} & \textcolor{blue}{$-3.56 \pm 0.94$} & ${-4.09 \pm 0.48}$ & ${-4.37 \pm 0.05}$ \\
$\text{WIS}_{t}$ & ${-5.08}$ & ${-5.06}$ & ${-5.12}$ & ${-4.02}$ & ${-4.08}$ & \textcolor{blue}{$-3.92 \pm 0.07$} & \textcolor{red}{$-2.11 \pm 1.17$} & ${-3.96 \pm 1.46}$ & ${-4.12 \pm 0.19}$ & ${-4.28}$ \\
$\text{WIS}_{bt}$ & ${-4.77}$ & ${-4.61}$ & ${-4.87}$ & ${-4.02}$ & ${-4.37}$ & ${-3.82 \pm 0.06}$ & \textcolor{red}{$-2.5 \pm 0.54$} & \textcolor{blue}{$-3.55 \pm 1.04$} & ${-4.09 \pm 0.47}$ & ${-4.35 \pm 0.05}$ \\
DR & ${-0.52}$ & \textcolor{red}{$-0.05$} & ${-1.55}$ & ${-0.18}$ & ${-0.15}$ & ${-0.2}$ & \textcolor{blue}{$-0.12 \pm 0.04$} & ${-0.64 \pm 0.12}$ & ${-0.4 \pm 0.1}$ & ${-1.55}$ \\
\bottomrule

\end{tabular}
\caption{NEWS2 sub rate  $\in[-0.4, -0.15]$ low std}
\label{table: NEWS2 reward-subgroup-test_rate_-0.4--0.15_low_std}
\end{table}

%% file: rl_tables/sub/MIMIC3SepsisNEWS2Env-all_train-sub-test_rate_-0.4--0.15_high_std-table.tex
\begin{table}[h]
\centering
\tablestyle{4pt}{1.0}
\tiny
\begin{tabular}{lcccccccccc}
\toprule
metric & alt & max & min & random & weight & imitation & dqn & cql & iql & bcq \\
\midrule
$\text{RMSE}_{\text{IV}}$ & ${820.25}$ & ${887.4}$ & ${770.94}$ & ${771.37}$ & ${767.21}$ & \textcolor{blue}{$667.05 \pm 15.28$} & ${752.73 \pm 4.24}$ & \textcolor{red}{$639.54 \pm 5.5$} & ${668.83 \pm 7.44}$ & ${744.04 \pm 13.4}$ \\
$\text{RMSE}_{\text{vaso}}$ & ${0.82}$ & ${1.03}$ & ${0.65}$ & ${0.74}$ & ${0.81}$ & \textcolor{blue}{$0.64$} & ${0.71 \pm 0.04}$ & \textcolor{blue}{$0.64$} & \textcolor{red}{$0.63 \pm 0.01$} & \textcolor{blue}{$0.64$} \\
$\text{WIS}$ & ${-8.81}$ & \textcolor{blue}{$-3.48$} & ${-8.81}$ & ${-5.15}$ & ${-5.28}$ & ${-4.27 \pm 0.18}$ & \textcolor{red}{$-1.49 \pm 0.75$} & ${-4.43 \pm 0.73}$ & ${-4.06 \pm 0.32}$ & ${-8.81}$ \\
$\text{WIS}_{b}$ & ${-7.33}$ & ${-4.26}$ & ${-7.24}$ & ${-4.99}$ & ${-4.57}$ & ${-4.21 \pm 0.2}$ & \textcolor{red}{$-1.63 \pm 0.44$} & ${-4.47 \pm 0.56}$ & \textcolor{blue}{$-4.05 \pm 0.3$} & ${-7.2 \pm 0.41}$ \\
$\text{WIS}_{t}$ & ${-8.81}$ & \textcolor{blue}{$-3.48$} & ${-8.81}$ & ${-5.15}$ & ${-5.28}$ & ${-4.27 \pm 0.18}$ & \textcolor{red}{$-1.49 \pm 0.75$} & ${-4.43 \pm 0.73}$ & ${-4.06 \pm 0.32}$ & ${-8.81}$ \\
$\text{WIS}_{bt}$ & ${-7.12}$ & ${-4.22}$ & ${-7.45}$ & ${-5.01}$ & ${-4.81}$ & ${-4.2 \pm 0.22}$ & \textcolor{red}{$-1.61 \pm 0.4$} & ${-4.52 \pm 0.5}$ & \textcolor{blue}{$-4.07 \pm 0.3$} & ${-7.28 \pm 0.3}$ \\
DR & ${-0.64}$ & ${-0.38}$ & ${-1.46}$ & ${-0.34}$ & \textcolor{blue}{$-0.19$} & ${-0.36}$ & \textcolor{red}{$-0.09 \pm 0.02$} & ${-0.61 \pm 0.03}$ & ${-0.47 \pm 0.03}$ & ${-1.33 \pm 0.09}$ \\
\bottomrule

\end{tabular}
\caption{NEWS2 sub rate  $\in[-0.4, -0.15]$ high std}
\label{table: NEWS2 reward-subgroup-test_rate_-0.4--0.15_high_std}
\end{table}

%% file: rl_tables/sub/MIMIC3SepsisNEWS2Env-all_train-sub-test_rate_-0.15-0_low_std-table.tex
\begin{table}[h]
\centering
\tablestyle{4pt}{1.0}
\tiny
\begin{tabular}{lcccccccccc}
\toprule
metric & alt & max & min & random & weight & imitation & dqn & cql & iql & bcq \\
\midrule
$\text{RMSE}_{\text{IV}}$ & ${688.63}$ & ${833.0}$ & ${457.4}$ & ${532.53}$ & ${486.77}$ & \textcolor{blue}{$405.15 \pm 3.15$} & ${463.11 \pm 10.85}$ & \textcolor{red}{$398.49 \pm 8.64$} & ${444.65 \pm 15.04}$ & ${443.3 \pm 6.13}$ \\
$\text{RMSE}_{\text{vaso}}$ & ${0.67}$ & ${0.89}$ & \textcolor{blue}{$0.26$} & ${0.48}$ & ${0.58}$ & \textcolor{blue}{$0.26$} & ${0.41 \pm 0.09}$ & \textcolor{red}{$0.25 \pm 0.01$} & \textcolor{blue}{$0.26$} & \textcolor{red}{$0.25 \pm 0.01$} \\
$\text{WIS}$ & ${-4.56}$ & ${-4.6}$ & ${-4.56}$ & ${-5.59}$ & ${-5.59}$ & ${-5.12 \pm 1.03}$ & \textcolor{red}{$-2.31 \pm 1.41$} & \textcolor{blue}{$-3.61 \pm 1.39$} & ${-4.15 \pm 0.58}$ & ${-4.56}$ \\
$\text{WIS}_{b}$ & ${-4.65}$ & ${-4.28}$ & ${-4.69}$ & ${-5.11}$ & ${-5.31}$ & ${-4.33 \pm 0.81}$ & \textcolor{red}{$-2.57 \pm 1.1$} & \textcolor{blue}{$-3.68 \pm 0.8$} & ${-4.12 \pm 0.57}$ & ${-4.58 \pm 0.08}$ \\
$\text{WIS}_{t}$ & ${-4.56}$ & ${-4.6}$ & ${-4.56}$ & ${-5.59}$ & ${-5.59}$ & ${-3.65 \pm 0.37}$ & \textcolor{red}{$-2.31 \pm 1.41$} & \textcolor{blue}{$-3.61 \pm 1.39$} & ${-4.16 \pm 0.57}$ & ${-4.56}$ \\
$\text{WIS}_{bt}$ & ${-4.53}$ & ${-4.36}$ & ${-4.68}$ & ${-5.12}$ & ${-5.32}$ & \textcolor{blue}{$-3.56 \pm 0.22$} & \textcolor{red}{$-2.55 \pm 0.95$} & ${-3.62 \pm 0.82}$ & ${-4.13 \pm 0.57}$ & ${-4.57 \pm 0.11}$ \\
DR & ${-0.31}$ & \textcolor{blue}{$-0.1$} & ${-1.65}$ & ${-0.13}$ & \textcolor{red}{$-0.08$} & ${-0.15}$ & ${-0.11 \pm 0.05}$ & ${-0.77 \pm 0.1}$ & ${-0.33 \pm 0.08}$ & ${-1.61 \pm 0.03}$ \\
\bottomrule

\end{tabular}
\caption{NEWS2 sub rate  $\in[-0.15, 0]$ low std}
\label{table: NEWS2 reward-subgroup-test_rate_-0.15-0_low_std}
\end{table}

%% file: rl_tables/sub/MIMIC3SepsisNEWS2Env-all_train-sub-test_rate_-0.15-0_high_std-table.tex
\begin{table}[h]
\centering
\tablestyle{4pt}{1.0}
\tiny
%
\caption{NEWS2 sub rate  $\in[-0.15, 0]$ high std}
\label{table: NEWS2 reward-subgroup-test_rate_-0.15-0_high_std}
\end{table}

%% file: rl_tables/sub/MIMIC3SepsisNEWS2Env-all_train-sub-test_rate_0-0.15_low_std-table.tex
\begin{table}[h]
\centering
\tablestyle{4pt}{1.0}
\tiny
%
\caption{NEWS2 sub rate $\in[0, 0.15]$ low std}
\label{table: NEWS2 reward-subgroup-test_rate_0-0.15_low_std}
\end{table}

%% file: rl_tables/sub/MIMIC3SepsisNEWS2Env-all_train-sub-test_rate_0-0.15_high_std-table.tex
\begin{table}[h]
\centering
\tablestyle{4pt}{1.0}
\tiny
%
\caption{NEWS2 sub rate $\in[0, 0.15]$ high std}
\label{table: NEWS2 reward-subgroup-test_rate_0-0.15_high_std}
\end{table}

%% file: rl_tables/sub/MIMIC3SepsisNEWS2Env-all_train-sub-test_rate_0.15-0.4_low_std-table.tex
\begin{table}[h]
\centering
\tablestyle{4pt}{1.0}
\tiny
%
\caption{NEWS2 sub rate $\in[0.15, 0.4]$ low std}
\label{table: NEWS2 reward-subgroup-test_rate_0.15-0.4_low_std}
\end{table}

%% file: rl_tables/sub/MIMIC3SepsisNEWS2Env-all_train-sub-test_rate_0.15-0.4_high_std-table.tex
\begin{table}[h]
\centering
\tablestyle{4pt}{1.0}
\tiny
%
\caption{NEWS2 sub rate $\in[0.15, 0.4]$ high std}
\label{table: NEWS2 reward-subgroup-test_rate_0.15-0.4_high_std}
\end{table}

%% file: rl_tables/sub/MIMIC3SepsisNEWS2Env-all_train-sub-test_rate_0.4-_low_std-table.tex
\begin{table}[h]
\centering
\tablestyle{4pt}{1.0}
\tiny
%
\caption{NEWS2 sub rate $\in[0.4, \infty]$ low std}
\label{table: NEWS2 reward-subgroup-test_rate_0.4-_low_std}
\end{table}

%% file: rl_tables/sub/MIMIC3SepsisNEWS2Env-all_train-sub-test_rate_0.4-_high_std-table.tex
\begin{table}[h]
\centering
\tablestyle{4pt}{1.0}
\tiny
%
\caption{NEWS2 sub rate $\in[0.4, \infty]$ high std}
\label{table: NEWS2 reward-subgroup-test_rate_0.4-_high_std}
\end{table}

%% file: appendix_sections/ratio_plot.tex
\section{Importance Ratio Histogram of Naive Baselines}
\label{sec:app-ratio plot}
\begin{figure}[h]
\centering
\includegraphics[width=0.8\linewidth]{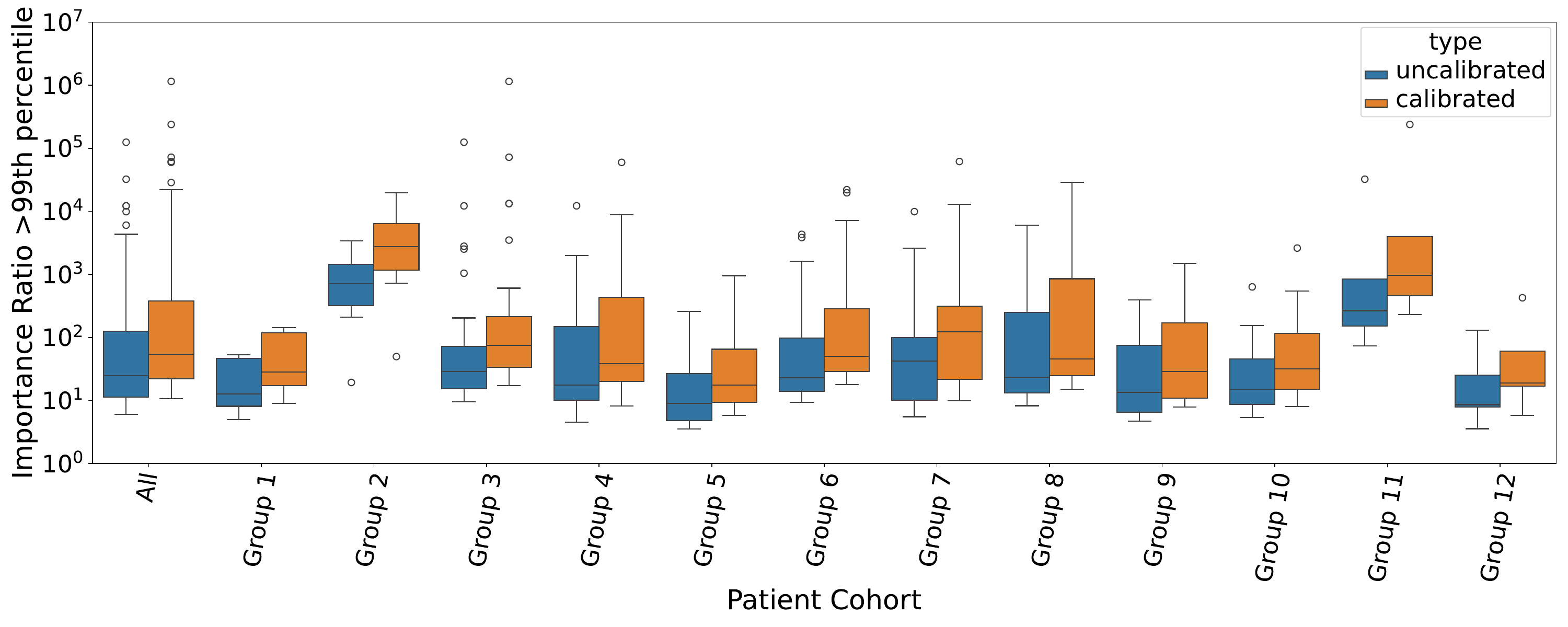}
\caption{\textbf{Importance ratio histogram of random policy $>$ 99th percentile on Outcome reward.}}
\label{fig: raw ratio Outcome random}
\end{figure}

\begin{figure}[h]
\centering
\includegraphics[width=0.8\linewidth]{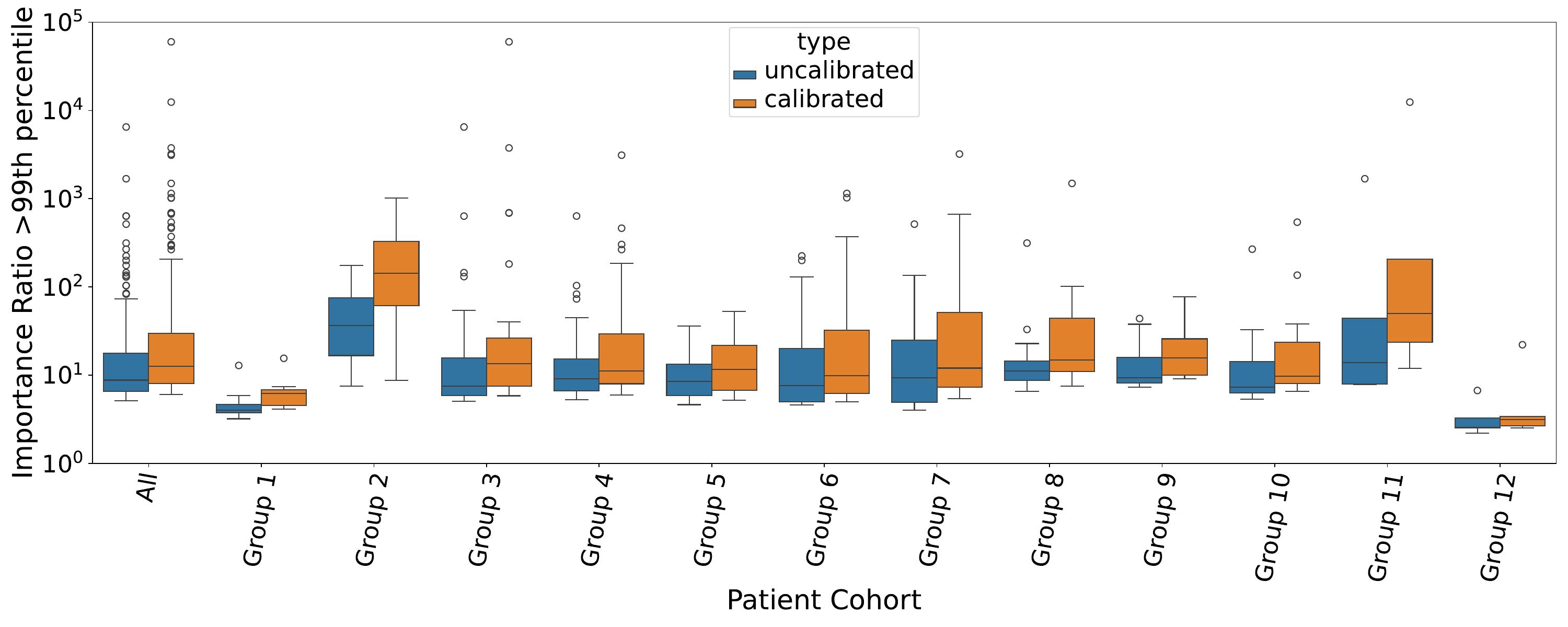}
\caption{\textbf{Importance ratio histogram of min policy $>$ 99th percentile on Outcome reward.}}
\label{fig: raw ratio Outcome min}
\end{figure}

\begin{figure}[h]
\centering
\includegraphics[width=0.8\linewidth]{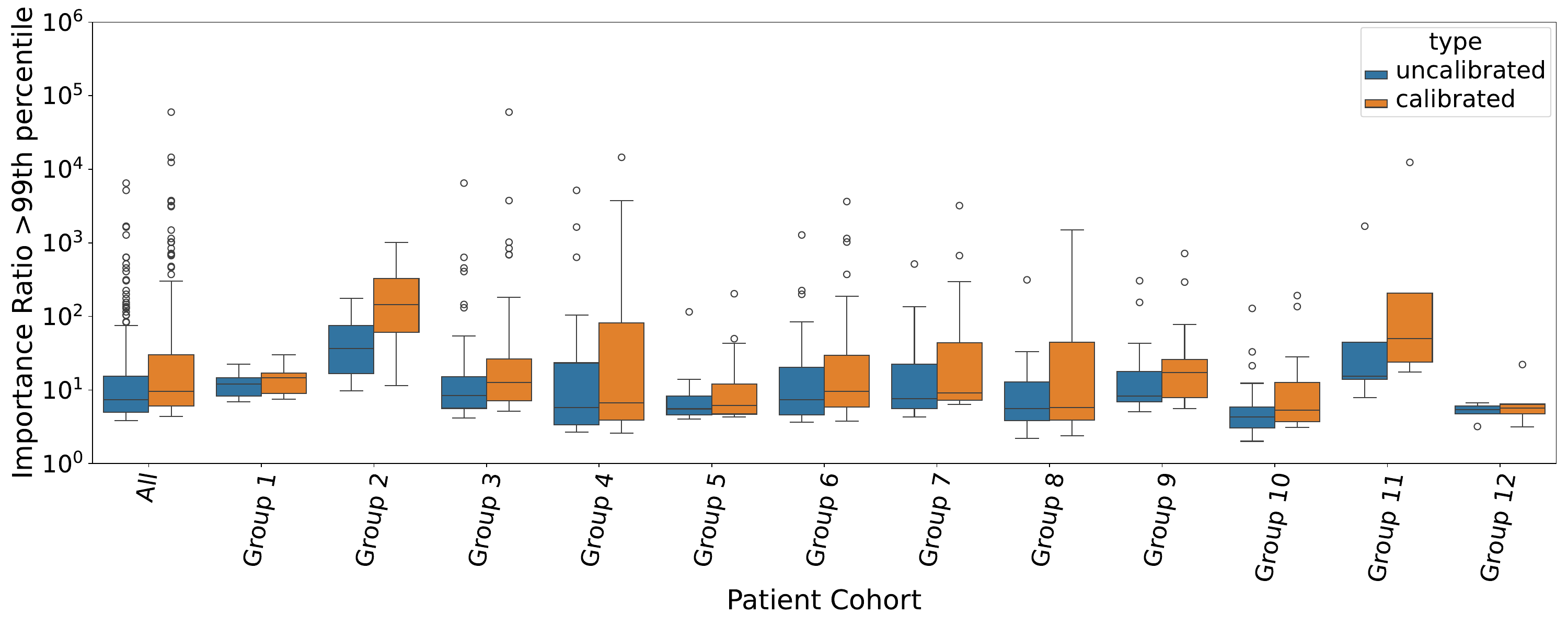}
\caption{\textbf{Importance ratio histogram of max policy $>$ 99th percentile on Outcome reward.}}
\label{fig: raw ratio Outcome max}
\end{figure}

\begin{figure}[h]
\centering
\includegraphics[width=0.8\linewidth]{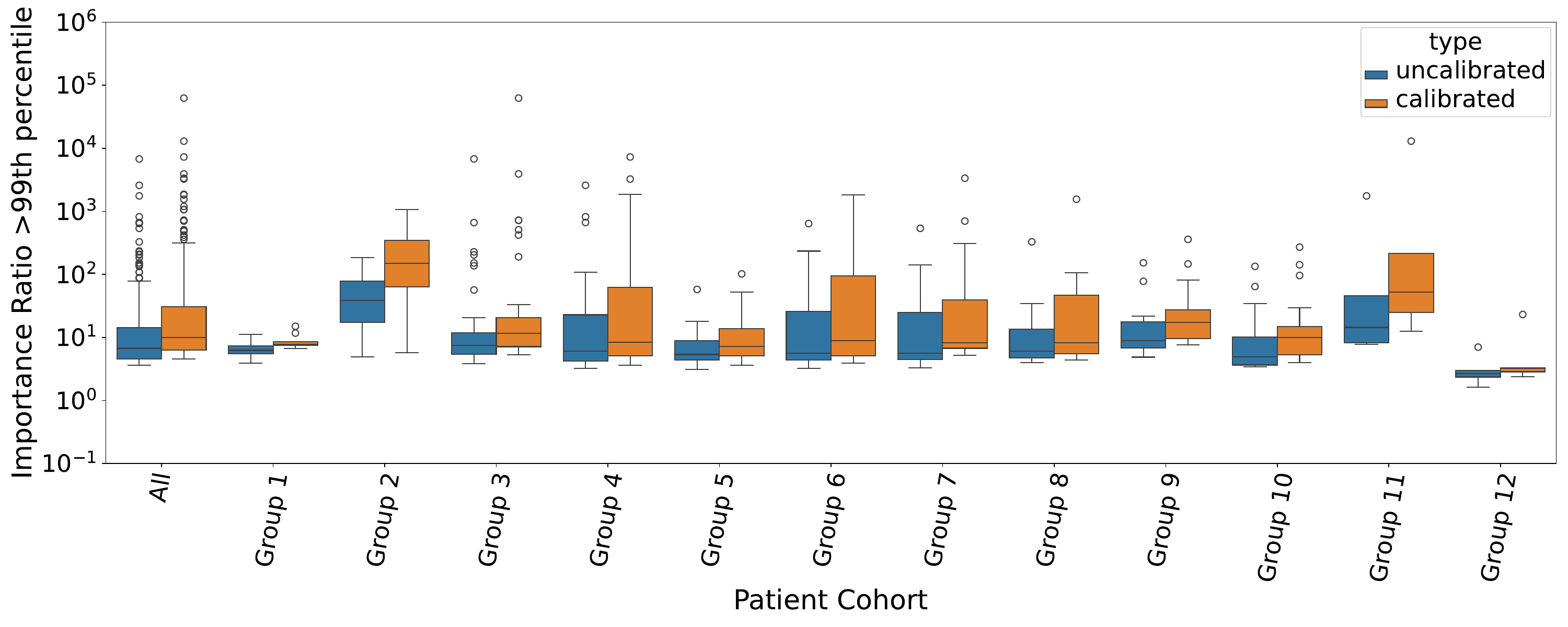}
\caption{\textbf{Importance ratio histogram of alt policy $>$ 99th percentile on Outcome reward.}}
\label{fig: raw ratio Outcome alt}
\end{figure}

\begin{figure}[h]
\centering
\includegraphics[width=0.8\linewidth]{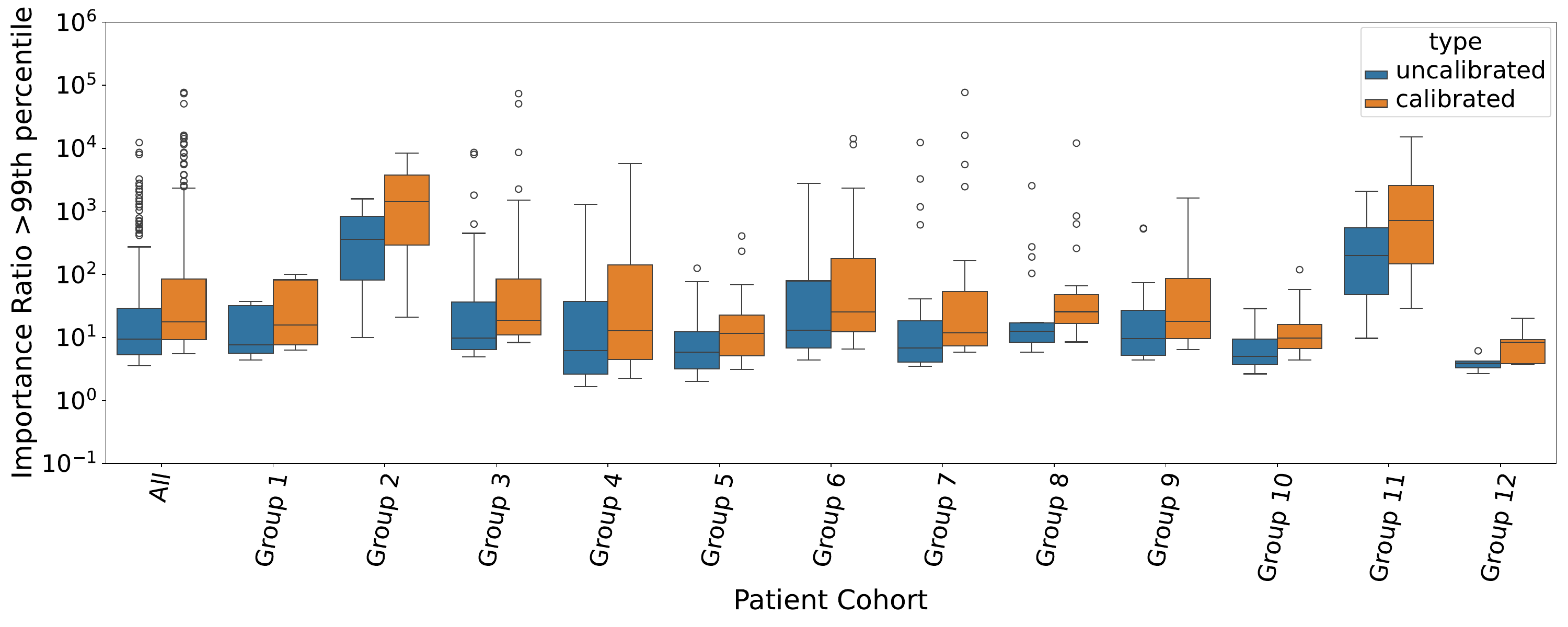}
\caption{\textbf{Importance ratio histogram of weight policy $>$ 99th percentile on Outcome reward.}}
\label{fig: raw ratio Outcome weight}
\end{figure}


\begin{figure}[h]
\centering
\includegraphics[width=0.8\linewidth]{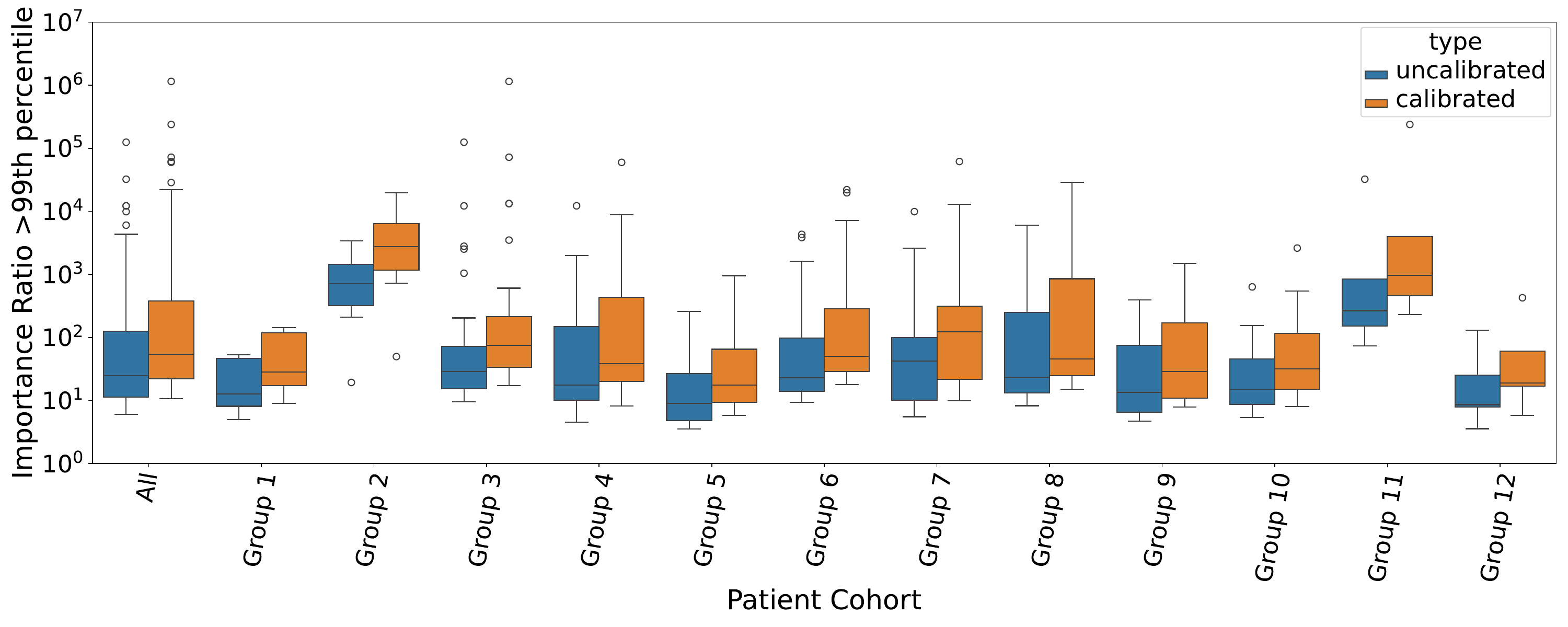}
\caption{\textbf{Importance ratio histogram of random policy $>$ 99th percentile on SOFA reward.}}
\label{fig: raw ratio SOFA random}
\end{figure}

\begin{figure}[h]
\centering
\includegraphics[width=0.8\linewidth]{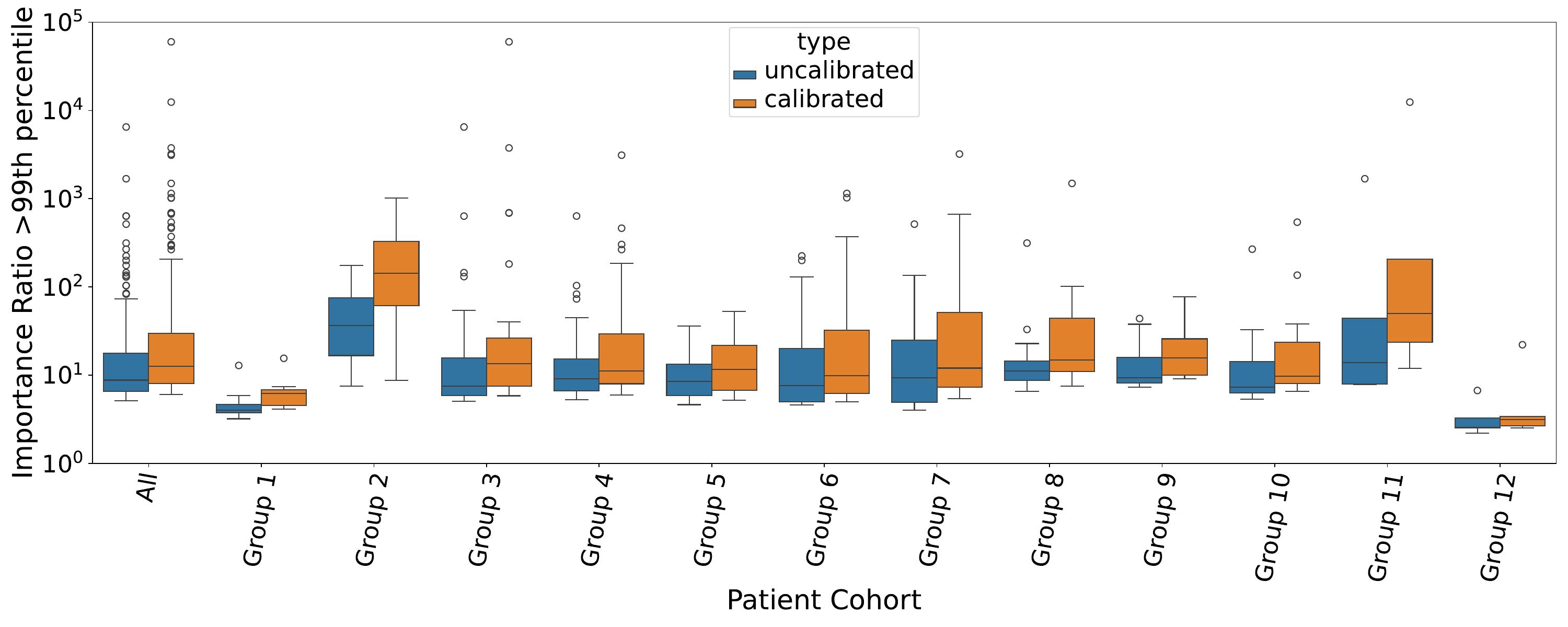}
\caption{\textbf{Importance ratio histogram of min policy $>$ 99th percentile on SOFA reward.}}
\label{fig: raw ratio SOFA min}
\end{figure}

\begin{figure}[h]
\centering
\includegraphics[width=0.8\linewidth]{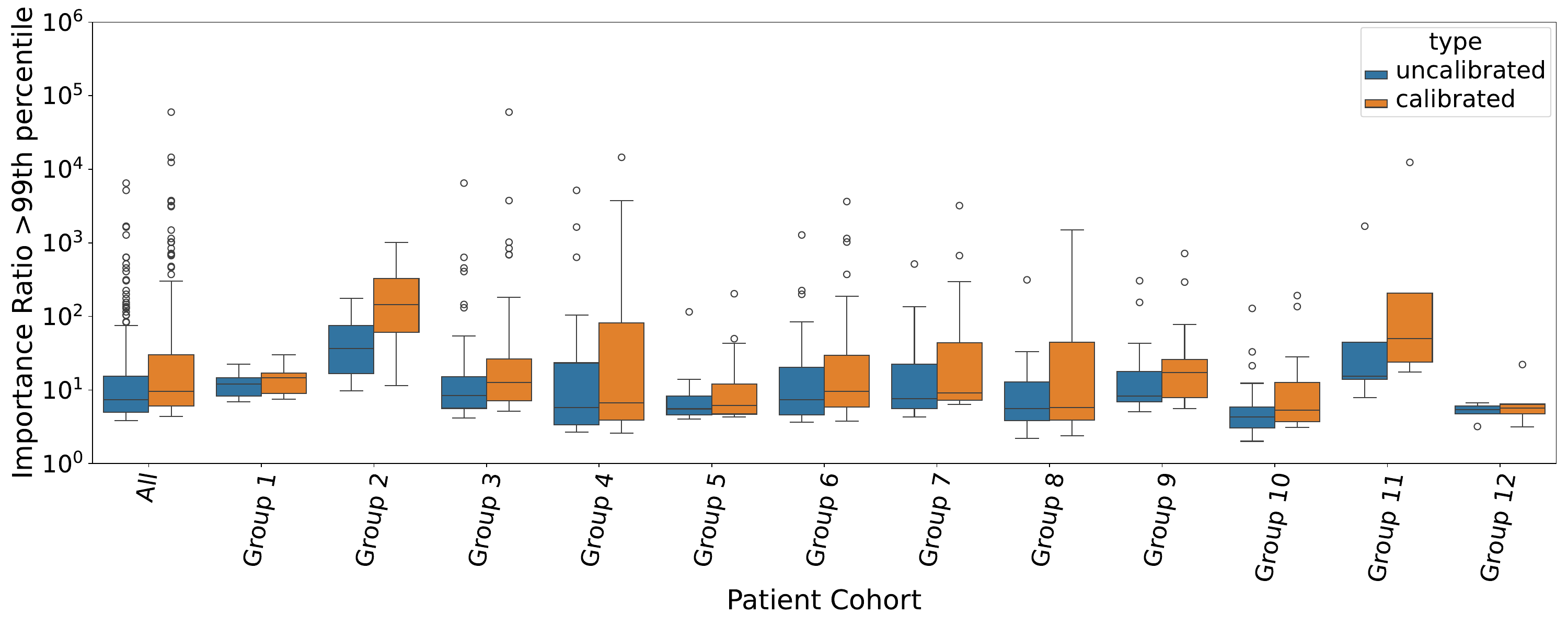}
\caption{\textbf{Importance ratio histogram of max policy $>$ 99th percentile on SOFA reward.}}
\label{fig: raw ratio SOFA max}
\end{figure}

\begin{figure}[h]
\centering
\includegraphics[width=0.8\linewidth]{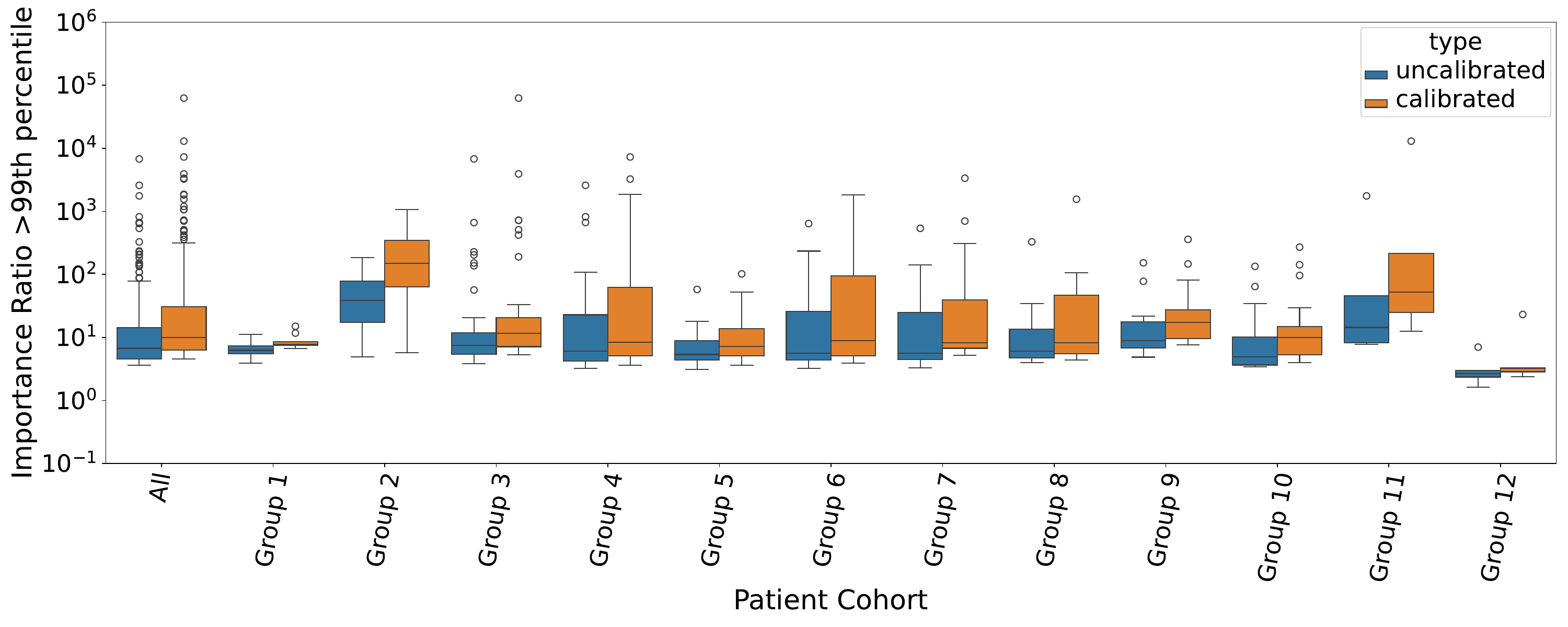}
\caption{\textbf{Importance ratio histogram of alt policy $>$ 99th percentile on SOFA reward.}}
\label{fig: raw ratio SOFA alt}
\end{figure}

\begin{figure}[h]
\centering
\includegraphics[width=0.8\linewidth]{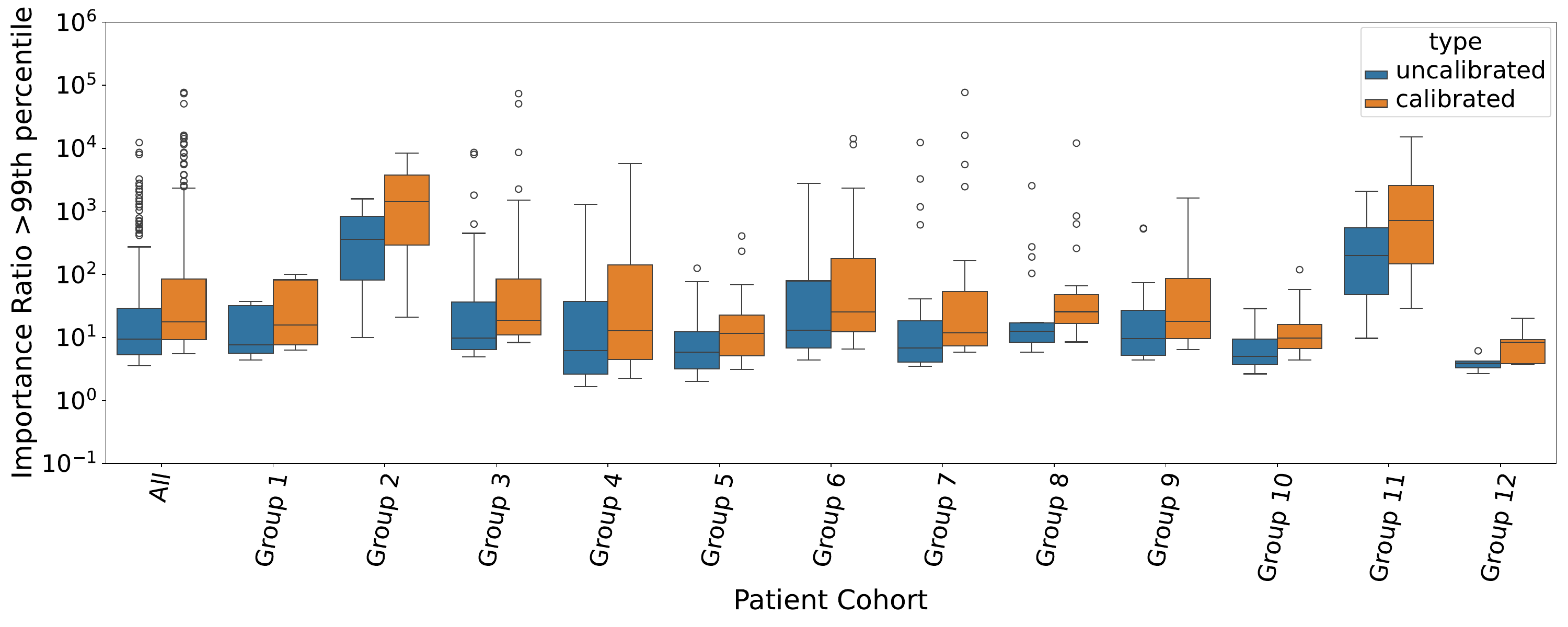}
\caption{\textbf{Importance ratio histogram of weight policy $>$ 99th percentile on SOFA reward.}}
\label{fig: raw ratio SOFA weight}
\end{figure}



\begin{figure}[h]
\centering
\includegraphics[width=0.8\linewidth]{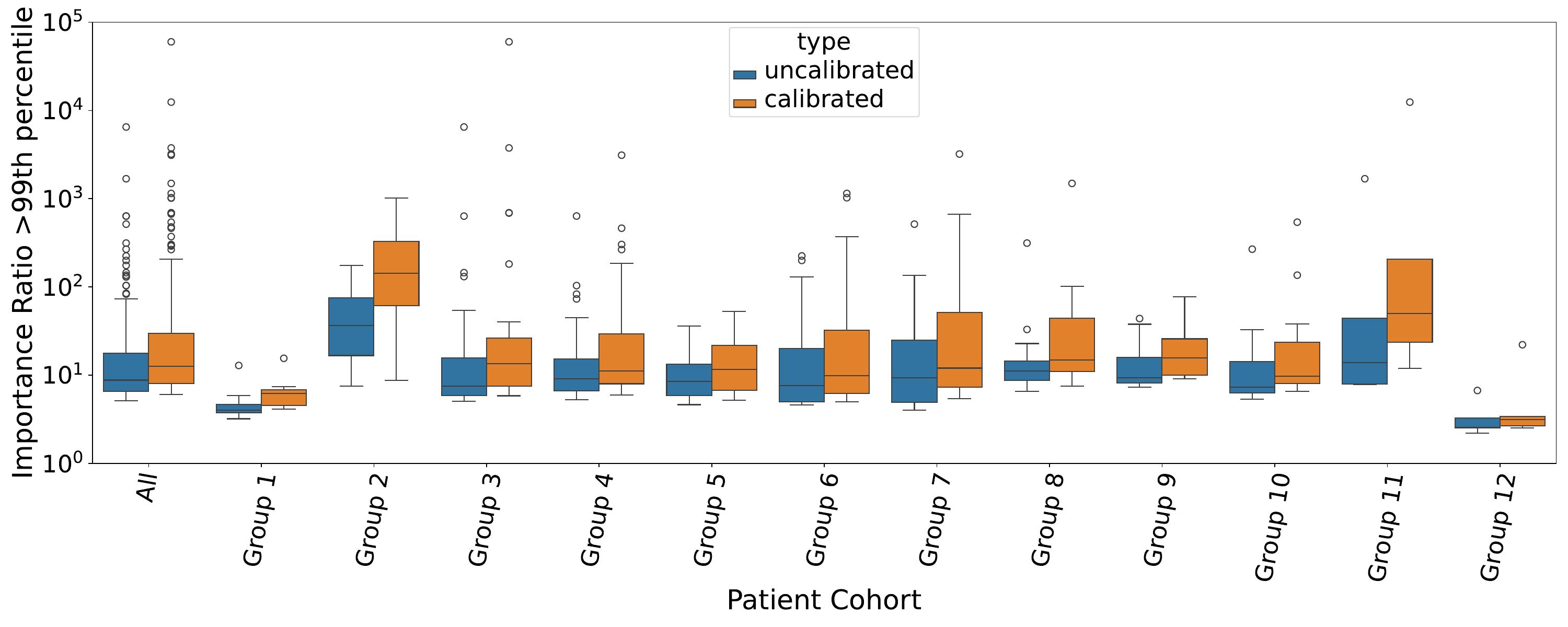}
\caption{\textbf{Importance ratio histogram of min policy $>$ 99th percentile on NEWS2 reward.}}
\label{fig: raw ratio NEWS2 min}
\end{figure}

\begin{figure}[h]
\centering
\includegraphics[width=0.8\linewidth]{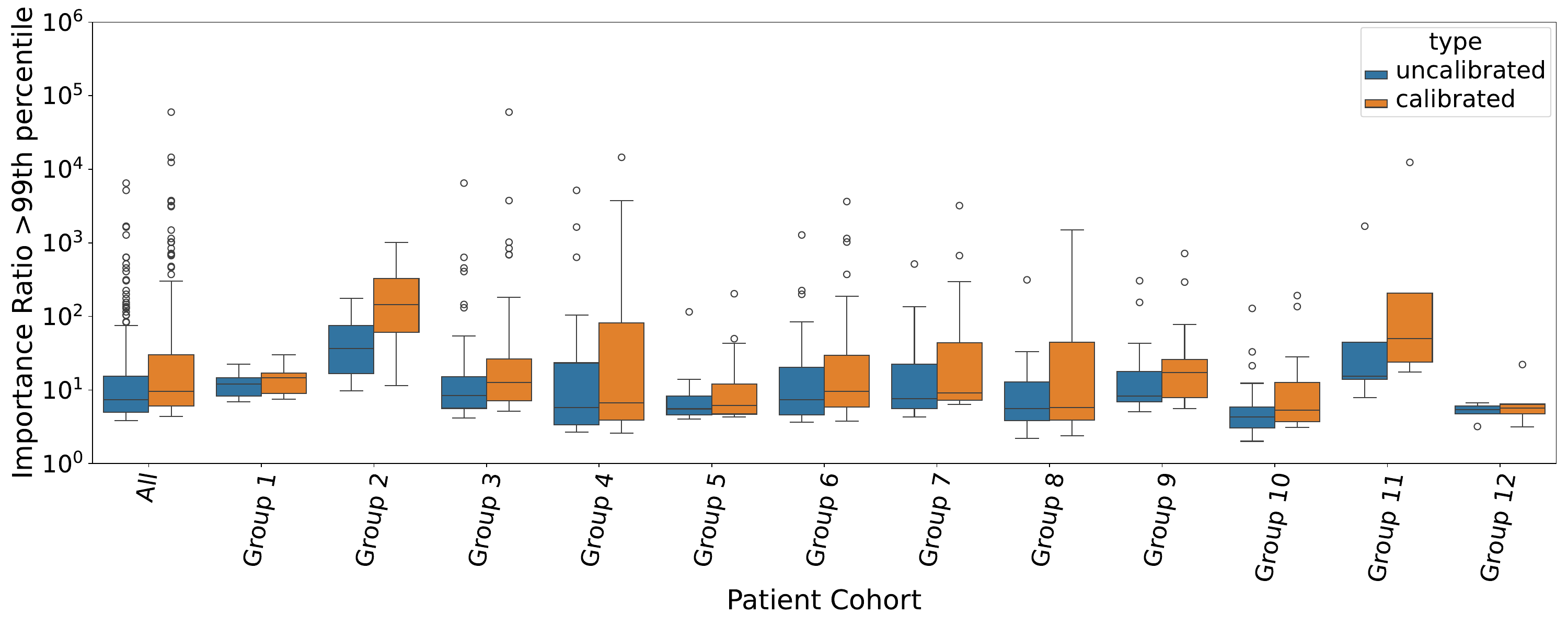}
\caption{\textbf{Importance ratio histogram of max policy $>$ 99th percentile on NEWS2 reward.}}
\label{fig: raw ratio NEWS2 max}
\end{figure}

\begin{figure}[h]
\centering
\includegraphics[width=0.8\linewidth]{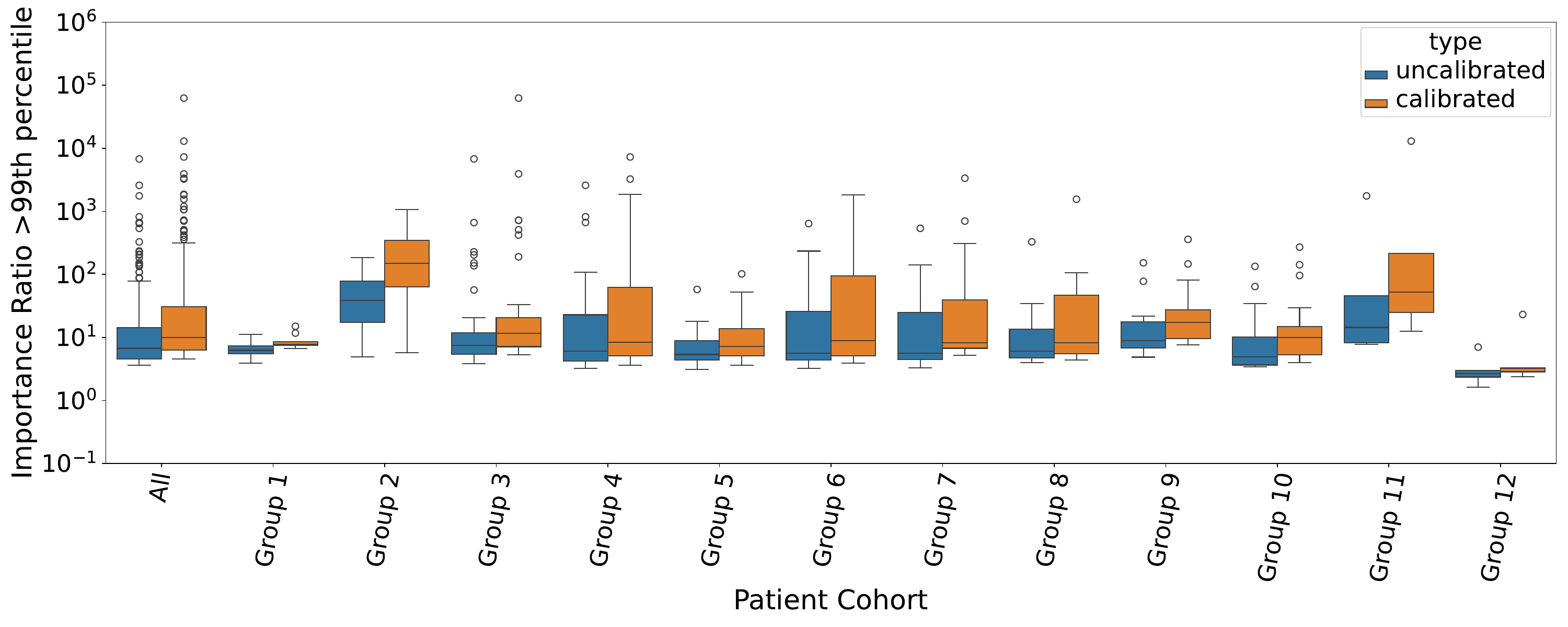}
\caption{\textbf{Importance ratio histogram of alt policy $>$ 99th percentile on NEWS2 reward.}}
\label{fig: raw ratio NEWS2 alt}
\end{figure}

\begin{figure}[h]
\centering
\includegraphics[width=0.8\linewidth]{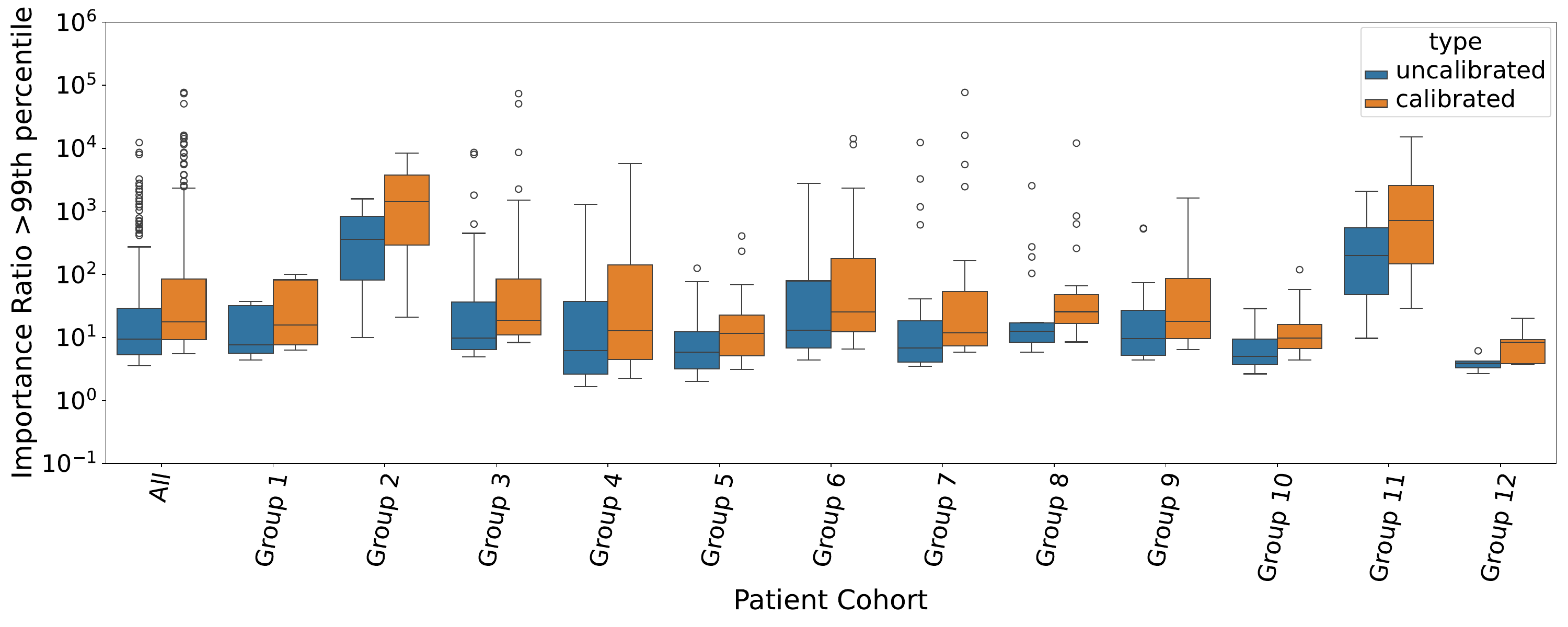}
\caption{\textbf{Importance ratio histogram of weight policy $>$ 99th percentile on NEWS2 reward.}}
\label{fig: raw ratio NEWS2 weight}
\end{figure}